\pdfoutput=1
\documentclass[11pt]{article}
\usepackage[preprint]{acl}
\usepackage{times}
\usepackage{latexsym}
\usepackage[T1]{fontenc}
\usepackage[utf8]{inputenc}
\usepackage{microtype}
\usepackage{graphicx}
\usepackage{booktabs}
\usepackage{array}
\usepackage{longtable}
\usepackage{amsmath,amssymb,amsthm}
\usepackage{multirow}
\usepackage{xcolor}
\usepackage{tikz}
\usetikzlibrary{positioning,arrows.meta,fit,decorations.pathreplacing,calc,backgrounds}
\usepackage{algorithm}
\usepackage[noend]{algpseudocode}
\usepackage{framed}
\usepackage{float}
\usepackage{latexml}
\newcommand{\ltlabel}[1]{\iflatexml\else\label{#1}\fi}
\newcommand{\ltlabelafter}[1]{\iflatexml\label{#1}\fi}

\newtheorem{proposition}{Proposition}
\newtheorem{lemma}{Lemma}
\newtheorem{corollary}{Corollary}
\theoremstyle{definition}

\newcommand{\papertitle}{From Seeing to Reporting: A Theory of Reliable\\Visual Set Generation in Vision-Language Models}
\newcommand{\cfp}{c_{\mathrm{FP}}}
\newcommand{\cfn}{c_{\mathrm{FN}}}
\newcommand{\SA}{\mathcal{S}_A}
\newcommand{\src}[1]{\mbox{\textsf{\footnotesize\color{black!60}[#1]}}}

\hypersetup{pdftitle={From Seeing to Reporting: A Theory of Reliable Visual Set Generation in Vision-Language Models},pdfauthor={Junzhe Shi, Yuan Gan, Shida Jiang},pdfsubject={Theory and inference-time procedures for reliable visual reports from vision-language models},pdfkeywords={vision-language models, VLMs, visual reporting, object counting, object detection, document layout analysis, screen parsing, GUI grounding, ScreenSpot-Pro, self-consistency, multi-model ensembles}}
\title{\papertitle}

\iflatexml\let\And\and\fi
\author{Junzhe Shi\thanks{Corresponding author: Junzhe Shi.} \\
  Systems Engineering \\
  University of California, Berkeley \\
  Berkeley, CA, USA \\
  \texttt{junzhe@berkeley.edu} \\\And
  Yuan Gan \\
  Department of AI \\
  Quotr AI \\
  Emeryville, CA, USA \\
  \texttt{yuan@quotr.ai} \\\And
  Shida Jiang \\
  Systems Engineering \\
  University of California, Berkeley \\
  Berkeley, CA, USA \\
  \texttt{shida\_jiang@berkeley.edu}}

\begin{document}
\maketitle
\begin{abstract}
Vision-language models (VLMs) are asked to report what they see: to count objects, box them, parse a screen or pick a click target (GUI grounding). We argue that reporting is part of the visual task, not merely its formatting. In a dense scene with 173 shapes, GPT-6.1 lists 173 records that match every shape, yet the counts it states in the same response sum to 188. \emph{From Seeing to Reporting} (FSR) is a theory of visual reporting and a set of inference-time procedures that improve the reports of frozen VLMs. It models reporting as a chain from responses to decoded records, constructed sets and task answers, linked by task-compatible matching. The chain gives exact relations for deleting records, averaging their coordinates and switching readouts, and it splits the error of one answer chosen among several calls into missing candidates, lost representatives and wrong choices. These relations guide procedures for counting, detection, document layout, screen parsing and GUI grounding that use the task specification and a shipped configuration, without requiring labels from the target task. On 50 dense synthetic scenes, a one-call reporting-and-decoding protocol cuts per-type count error by 85.2\% for GPT-6.1 and 98.8\% for Claude Sonnet 5.5. On 600 ScreenSpot-Pro items, a GUI-grounding tier that combines three models lowers the miss rate from 13.67\% for GPT-6.1's two-stage zoom to 9.33\%, against 12.67\% for three such zooms combined by a cluster vote at about the same cost. A plain vote over the same three models misses 12.33\%: the other models supply the candidates, and selecting among them turns those candidates into fewer misses. We release the protocols, the stored responses and the analysis code.
\end{abstract}

\section{Introduction}\label{sec:intro}

\begin{figure*}[tp]
\centering
\includegraphics[width=\textwidth,alt={Three panels. (a) Bar chart of five shape classes in one dense synthetic scene with 173 shapes: the list count of GPT-6.1 matches the truth for every class, while the counts it states in the same response sum to 188. (b) An Excel data-validation menu with the instruction only allow values from a list: a single click by Gemini 3.8 lands on the item Time, and a second call on a zoomed crop lands on the target item List. (c) Miss rate against cost per item on 600 ScreenSpot-Pro items: FSR's three-model GUI-grounding tier is lowest at 9.33 percent, against 13.67 percent for GPT-6.1 two-stage zoom, 12.67 percent for three such zooms, 12.33 percent for a plain cluster vote over the same three models and 11.50 percent for a crop-first baseline.}]{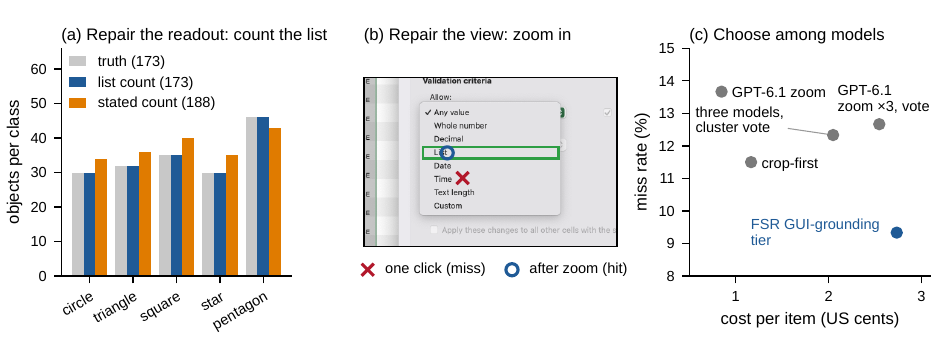}
\caption{Three repairs that change what a model delivers. (a) One response of GPT-6.1 to a dense synthetic scene (image 28 of SCP-dense, our dense synthetic counting scenes; 173 shapes of five classes): the \emph{list count}, the records of each class in the model's list counted by code, matches the truth, while the \emph{stated count} the model wrote in the same response sums to 188; each of the 173 records matches a different shape of the same class (Section~\ref{sec:res:readout}). (b) For the instruction ``only allow values from a list'' in an Excel menu, a single click by Gemini 3.8 lands on ``Time''; a second call on a zoomed crop (ReGround, \citealp{li2025screenspotpro}) lands inside the target ``List'' (green box). On 200 ScreenSpot-Pro items, two-stage zoom cuts Gemini 3.8's miss rate by 65\% (Section~\ref{sec:res:view}). (c) Miss rate against cost per item at list prices on 600 ScreenSpot-Pro items: GPT-6.1's two-stage zoom, three such zooms with a cluster vote, a crop-first baseline in the style of GUI-Lens \citep{fu2026guilens}, a cluster vote over three models' zoomed points, and FSR's GUI-grounding tier (Algorithm~\ref{alg:clicks}), which selects among the same points with a scorer and a comparison; every item is in the denominator. Paired differences with intervals are in Section~\ref{sec:res:choose}: the tier's difference from the crop-first baseline has an interval that reaches 0, and its differences from the other flows exclude 0.}
\label{fig:teaser}
\end{figure*}

Vision-language models are now asked to report a set of objects in an image: to count them, to box them, to parse every element of a screen, or to pick the one element a user means (GUI grounding: given a screenshot and an instruction, return the point to click). A typical answer has two parts. The first is a \emph{list}: one entry per object, each giving a class name and a location (a point or a box). We call each entry a \emph{record}. The second, when it is requested, is a \emph{stated count}: the numbers the model writes for how many objects of each class it found. Code can also count the records in the list; we call that the \emph{list count}. The two counts should agree. Often they do not.

Figure~\ref{fig:teaser}a shows one response of GPT-6.1 to a dense synthetic scene with 173 shapes of five classes. Its list has 173 records, and each record matches a different shape of the same class within 20 pixels, so the list is exact. In the same response, the model states counts that add up to 188. Across 50 such scenes, GPT-6.1 wrote an exact list in 20 scenes, and in 19 of them its stated counts were wrong. Sonnet 5.5 wrote an exact list in 39 scenes (after mapping its word ``diamond'' to the class ``square''), and its stated counts were wrong in all 39. The response contains a correct enumeration; the model's own summary of that enumeration is not faithful to it.

Visual set reports are becoming the interface between perception models and the systems that use them: inventory counts, document layout, screen parsing for computer-use agents, click targets. Practitioners compensate with folklore: ask again, ask the model to check itself, overlay a grid, tell it not to miss anything, vote across models. Some of these help, some do nothing, and some make things worse. Without a model of where reporting errors come from, it is hard to predict which. Our thesis is that \emph{reporting is part of the visual task, not merely its formatting}: what a model delivers depends on how it is asked to report, how the report is decoded and read, and how several reports are combined, and each of these steps can be analysed and improved.

\begin{figure*}[tp]
\centering
\iflatexml
\includegraphics[width=\textwidth,alt={Overview diagram of FSR in two aligned lanes with the same four steps: call, decode, construct, read out. Top lane, any task: image, task and budget; responses; records and stated counts; set; answer (count, set or click). Blue cards under each step give the repair: request and view (list-first request, regions or list per model, two-stage zoom); decode and recover (aliases mapped in code, degeneracy flags, one follow-up); construct (for set error on over-reporting tasks keep what all three calls report; delete if p1 is below one half at equal costs, averaging separate); read out (count the list, the stated count only if the list is flagged). Bottom lane, GUI grounding: screenshot and instruction; three zoomed points; candidates; cluster representatives; one point, or none, which counts as a miss. Cards: two-stage zoom by GPT-6.1, Sonnet 5.5 and Gemini 3.8, where F (no correct candidate) arises; register each point; cluster within delta and keep each cluster's best-scored point, where L (a correct candidate but no correct representative) arises; with one cluster return its point, with an alarm compare pairwise, where W (a correct representative not chosen) arises. A bracket marks the call step as seeing (coverage, F) and decoding, construction and readout as reporting (L, W). Bottom panels: Judge, where shipped defaults need no target labels and outputs alone cannot certify; Detect, where degeneracy flags and disagreement with a second model raise an alarm that triggers the comparison.}]{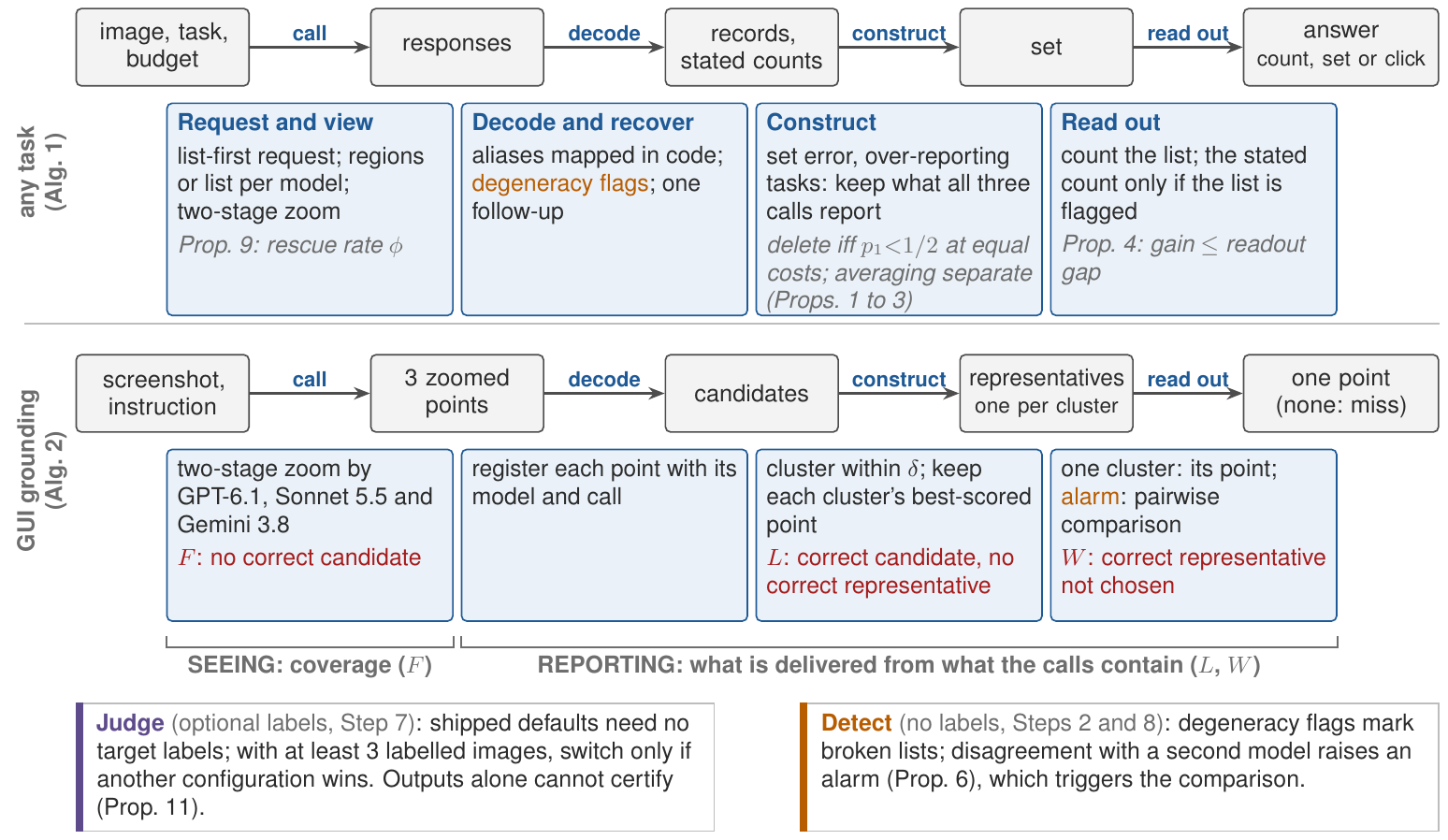}
\else
\resizebox{\textwidth}{!}{\providecommand\fsrref[1]{\ifdefined\hypersetup\ref*{#1}\else\ref{#1}\fi}%
\definecolor{fsrInk}{HTML}{2B2B2B}%
\definecolor{fsrLine}{HTML}{4D4D4D}%
\definecolor{fsrMuted}{HTML}{6E6E6E}%
\definecolor{fsrRule}{HTML}{BDBDBD}%
\definecolor{fsrObj}{HTML}{F3F3F3}%
\definecolor{fsrBlue}{HTML}{1F5A96}%
\definecolor{fsrBlueTint}{HTML}{EAF1F8}%
\definecolor{fsrJudge}{HTML}{5E4B8B}%
\definecolor{fsrDetect}{HTML}{B85C00}%
\definecolor{fsrFail}{HTML}{A61C1C}%
\def\fsrB{\fontsize{9pt}{10.8pt}\selectfont}\def\fsrS{\fontsize{8pt}{9.5pt}\selectfont}\def\fsrO{\fontsize{9pt}{10.5pt}\selectfont}%
\begin{tikzpicture}[
  font=\sffamily, text=fsrInk,
  obj/.style={draw=fsrLine, fill=fsrObj, line width=0.6pt, rounded corners=2pt, align=center, font=\sffamily\fsrO, minimum height=1.05cm, text width=2.1cm, inner sep=3.5pt},
  arr/.style={-{Stealth[length=2.2mm, width=1.7mm]}, line width=0.8pt, fsrLine},
  op/.style={font=\sffamily\fsrS\bfseries, text=fsrBlue, above=2pt, inner sep=0pt, align=center},
  card/.style={draw=fsrBlue, fill=fsrBlueTint, line width=0.6pt, rounded corners=2pt, align=left, font=\sffamily\fsrB, inner sep=4pt, anchor=north},
  fail/.style={font=\sffamily\fsrB, text=fsrFail, align=center, below=2pt, inner sep=0pt},
  lane/.style={font=\sffamily\fsrB\bfseries, text=fsrMuted, rotate=90, align=center},
  band/.style={line width=0.8pt, fsrMuted},
  blab/.style={font=\sffamily\fsrB\bfseries, text=fsrMuted, anchor=north, inner sep=2pt, below=1pt},
  side/.style={draw=fsrRule, fill=white, line width=0.6pt, align=left, font=\sffamily\fsrB, inner xsep=8pt, inner ysep=4pt}]
\node[obj] (x) at (0,0) {image, task,\\budget};
\node[obj] (h) at (4,0) {responses};
\node[obj] (z) at (8,0) {records,\\stated counts};
\node[obj] (s) at (12,0) {set};
\node[obj, text width=2.4cm] (y) at (16,0) {answer\\{\fsrS count, set or click}};
\draw[arr] (x) -- node[op]{call} (h);
\draw[arr] (h) -- node[op]{decode} (z);
\draw[arr] (z) -- node[op]{construct} (s);
\draw[arr] (s) -- node[op]{read out} (y);
\def\fsrcard#1#2#3{\parbox[t][2.6cm][t]{3.6cm}{\raggedright\hyphenpenalty=10000 {\bfseries\color{fsrBlue}#1}\\[2pt]#2\\[2pt]{\itshape\color{fsrMuted}#3}}}
\node[card] (c1) at (2,-0.75) {\fsrcard{Request and view}{list-first request; regions or list per model; two-stage zoom}{Prop.~\fsrref{prop:coverage}: rescue rate $\phi$}};
\node[card] (c2) at (6,-0.75) {\fsrcard{Decode and recover}{aliases mapped in code; \textcolor{fsrDetect}{degeneracy flags}; one follow-up}{}};
\node[card] (c3) at (10,-0.75) {\fsrcard{Construct}{set error, over-reporting tasks: keep what all three calls report}{delete iff $p_1{<}1/2$ at equal costs; averaging separate (Props.~\fsrref{prop:identity} to \fsrref{prop:geometry})}};
\node[card] (c4) at (14,-0.75) {\fsrcard{Read out}{count the list; the stated count only if the list is flagged}{Prop.~\fsrref{prop:readout}: gain $\le$ readout gap}};
\def\gy{-4.7}
\node[obj] (g0) at (0,\gy) {screenshot,\\instruction};
\node[obj] (g1) at (4,\gy) {3 zoomed\\points};
\node[obj] (g2) at (8,\gy) {candidates};
\node[obj] (g3) at (12,\gy) {representatives\\{\fsrS one per cluster}};
\node[obj, text width=2.4cm] (g4) at (16,\gy) {one point\\(none:\ miss)};
\draw[arr] (g0) -- node[op]{call} (g1);
\draw[arr] (g1) -- node[op]{decode} (g2);
\draw[arr] (g2) -- node[op]{construct} (g3);
\draw[arr] (g3) -- node[op]{read out} (g4);
\def\fsrgcard#1#2{\parbox[t][2.05cm][t]{3.6cm}{\raggedright\hyphenpenalty=10000 #1\\[2pt]{\color{fsrFail}#2}}}
\node[card] (e1) at (2,\gy-0.75) {\fsrgcard{two-stage zoom by GPT-6.1, Sonnet 5.5 and Gemini 3.8}{$F$: no correct candidate}};
\node[card] (e2) at (6,\gy-0.75) {\fsrgcard{register each point with its model and call}{}};
\node[card] (e3) at (10,\gy-0.75) {\fsrgcard{cluster within $\delta$; keep each cluster's best-scored point}{$L$: correct candidate, no correct representative}};
\node[card] (e4) at (14,\gy-0.75) {\fsrgcard{one cluster: its point; \textcolor{fsrDetect}{alarm}: pairwise comparison}{$W$: correct representative not chosen}};
\draw[fsrRule, line width=0.6pt] ($(x.west|-0,-3.75)+(-0.7,0)$) -- ($(y.east|-0,-3.75)$);
\node[lane] at ($(x.west)+(-0.45,-1.7)$) {any task\\(Alg.~\fsrref{alg:pipeline})};
\node[lane] at ($(g0.west)+(-0.45,-1.05)$) {GUI grounding\\(Alg.~\fsrref{alg:clicks})};
\coordinate (bandY) at (0,-8.15);
\draw[band] ($(c1.west|-bandY)+(0,0.15)$) -- ($(c1.west|-bandY)$) -- ($(c1.east|-bandY)$) -- ++(0,0.15);
\draw[band] ($(c2.west|-bandY)+(0,0.15)$) -- ($(c2.west|-bandY)$) -- ($(c4.east|-bandY)$) -- ++(0,0.15);
\node[blab] at ($(c1.west|-bandY)!0.5!(c1.east|-bandY)$) {SEEING: coverage ($F$)};
\node[blab] at ($(c2.west|-bandY)!0.5!(c4.east|-bandY)$) {REPORTING: what is delivered from what the calls contain ($L$, $W$)};
\node[side, anchor=north west] (j) at ($(x.west|-0,-8.9)$) {\parbox[t][1.45cm][t]{8.1cm}{\raggedright\hyphenpenalty=10000 {\bfseries\color{fsrJudge}Judge} {\color{fsrMuted}(optional labels, Step 7)}: shipped defaults need no target labels; with at least 3 labelled images, switch only if another configuration wins. Outputs alone cannot certify (Prop.~\fsrref{prop:nocert}).}};
\fill[fsrJudge] (j.north west) rectangle ($(j.south west)+(0.1,0)$);
\node[side, anchor=north east] (d) at ($(y.east|-0,-8.9)$) {\parbox[t][1.45cm][t]{8.1cm}{\raggedright\hyphenpenalty=10000 {\bfseries\color{fsrDetect}Detect} {\color{fsrMuted}(no labels, Steps 2 and 8)}: degeneracy flags mark broken lists; disagreement with a second model raises an alarm (Prop.~\fsrref{prop:alarm}), which triggers the comparison.}};
\fill[fsrDetect] (d.north west) rectangle ($(d.south west)+(0.1,0)$);
\end{tikzpicture}%
}
\fi
\caption{FSR for visual reporting with vision-language models. Top lane: the reporting chain of the default pipeline (Algorithm~\ref{alg:pipeline}), from the image, task and budget to the answer, with the repair at each step (blue cards) and the result that governs it. Bottom lane: the same four steps for GUI grounding with three models (Algorithm~\ref{alg:clicks}), and where each kind of miss arises: $F$, no correct candidate after the calls; $L$, a correct candidate but no correct representative after construction; $W$, a correct representative that readout does not choose. The miss rate is $F+L+W$ (Proposition~\ref{prop:three}). Calls determine coverage, what the responses contain; decoding, construction and readout determine what is delivered from it. $F$, $L$ and $W$ locate a miss; an operation can change more than one of them. Judge: choosing between configurations, for which the shipped defaults need no target labels. Detect: errors flagged without labels; in GUI grounding the alarm triggers the comparison. Every item is scored, and an item with no point counts as a miss.}
\label{fig:chain}
\end{figure*}

\paragraph{What we do.} We describe reporting as a chain with four objects (Figure~\ref{fig:chain}): a \emph{response}, the raw text of one call to a model; the \emph{records} decoded from it; a \emph{set} of records built from one or more responses, for example by keeping only records that two calls agree on; and the \emph{task answer} read from that set, such as a count per class or a click point. Three quantities connect the chain to every loss we study: the maximum number of records that can be paired one-to-one with true objects under the task's matching rule, the number of records that are scored, and the rule that reads the answer. They give exact relations for what each operation does. Deleting unconfirmed records lowers the error exactly when the deletion loses fewer than one match per two scored records removed (with equal costs of the two error types), whichever model or view supplied the confirmation. Averaging the coordinates of agreeing records is a separate effect that can help or hurt. The chain takes reports as they come: from one call on the whole image, or from several views produced by an existing view planner such as tiling or zoom, whose merging and cross-view confirmation are constructions like any other. Alongside these relations we state a behavioural model, kept separate from them: each object has a persistent tendency to be reported or missed, partly shared across models. This analysis of what a model reports complements analyses of what a single call can see \citep{shi2026zoomin}.

\paragraph{Seeing and reporting.} The title names the two things the chain separates. What a set of calls has seen is its \emph{coverage}: the correct records or candidates present anywhere in the responses, in any field, view or model. What is reported is a \emph{selection} from that coverage: which field is read, which records are kept, which point is output. The repairs in this paper add coverage, improve the selection, or, by averaging coordinates, move the records that are kept; on a fixed set of calls an operation helps exactly when, on what it changes, the errors it repairs outweigh the correct outputs it breaks, at the exchange rate the loss sets (Section~\ref{sec:theory}). Reading the list instead of the stated count, and deleting unconfirmed records, change only the selection. A smaller view or a second model adds coverage, and a selection must then use it: on 600 click items, three models' points with a plain vote have a miss rate not significantly different from that of three calls of one model, and choosing among the same points with a scorer and a comparison removes 18 more misses. Detecting and judging ask what the outputs alone reveal about a selection.

\paragraph{Three problems, kept apart.} Practice often treats three problems as one. \emph{Detecting} that a report is wrong can be done without labels, from disagreement between reports (Proposition~\ref{prop:alarm}). \emph{Judging} which of several reports or configurations is more accurate cannot in general be certified from outputs alone, because two scenes can produce the same outputs (Proposition~\ref{prop:nocert}); it needs labels, on the task itself or on related sources from which a configuration is carried over (Section~\ref{sec:res:transfer}). \emph{Repairing} a report works when the needed information is already available: in another field of the same response, in another view of the image, or in another model's report. A method that detects errors well need not judge well, and one that judges well need not repair. A model asked at run time to score or compare candidate answers, as in our GUI-grounding tier, is one more source of evidence for a repair, not a certificate. The paper is organized around this distinction.

\par\medskip\noindent\fbox{\parbox{\dimexpr\columnwidth-2\fboxsep-2\fboxrule\relax}{\small\textbf{Three answers.}
\emph{Do vision-language models report what they see?} Not reliably, and the gap is visible within one response: in dense scenes GPT-6.1 and Sonnet 5.5 often write an exact list while the counts they state in the same response are wrong.
\emph{Will better models make the repairs unnecessary?} Some failures are already absent on our tasks: on our synthetic counting scenes, the average coordinate offset is about 1\% of the matching tolerance, and Gemini 3.8's default reasoning setting removes most of its broken lists. The gap between list and stated count, and object-level errors that repeat across calls, remain in the newest models we tested.
\emph{More calls of the same model, or a different source of evidence?} For counts, clicks and error detection, a different source helped more: another field of the same response, the original pixels of a smaller crop, or a different model. Repeating the same call helps mainly by removing false records that do not repeat.
}}\par\medskip

\paragraph{What we find.} Measured against common remedies on the same stored calls (Section~\ref{sec:res:common}) and, for clicks, on 600 further items (Section~\ref{sec:res:choose}), the shipped defaults do better than these remedies in most comparisons, and the theory's quantities say where and why.
\begin{itemize}
\item \textbf{Repair the readout.} On dense synthetic scenes with five classes, counting the list instead of reading the stated count cuts per-type count error by 71\% to 84\% in the same responses of GPT-6.1 and Sonnet 5.5, and the full list-first protocol cuts it by 85.2\% and 98.8\% at the cost of one call, where self-consistency over three calls gains at most 23\%. On one-class photographs, and on aerial and document images where the lists themselves miss objects, the protocol's change is small and not significant for these two models; Gemini 3.8 needs its default setting, at which far fewer of its lists break (flagged lists 90\% to 12\%) (Sections~\ref{sec:res:readout} and \ref{sec:res:degen}).
\item \textbf{Repair the set.} When a partner call confirms the records of a base call, the change in set error splits exactly into deleting the records the partner does not report and averaging the coordinates of those it does. Deletion helps when the marginal matching rate $p_1$ of the deleted records is below $1/2$; the task family's default decides correctly in 85\% of draws over 54 base/partner pairs, and five labelled images cut the loss from a wrong decision from 0.43 to 0.17 errors per image. Confirmation is a default for set error on over-reporting families only; for F1 the default is a single call (Sections~\ref{sec:res:set} and \ref{sec:res:transfer}).
\item \textbf{Repair the view.} Two-stage zoom cuts the miss rate by 65\% for Gemini 3.8 and 41\% for Sonnet 5.5 (not significant for GPT-6.1), where a three-click vote or zooming only on disagreement gains at most 13\%. Confirming a whole-image report with four crop reports cuts set error by up to 45.5\% and hurts where $p_1>1/2$ (Section~\ref{sec:res:view}).
\item \textbf{Detect.} A second, cheaper model catches 61\% of GPT-6.1's misses and a second call of the same model 17\%, at the price of flagging more items (Section~\ref{sec:res:persist}).
\item \textbf{Repair a click by choosing among models.} The miss rate of one answer chosen among several calls splits exactly into three terms: no correct candidate, no correct cluster representative, a wrong choice among representatives. FSR's GUI-grounding tier, built on this split, misses 9.33\% of 600 ScreenSpot-Pro targets, against 13.67\% for GPT-6.1's two-stage zoom, 12.67\% for three such zooms at about the same cost, and 12.33\% for a plain vote over the same three models' points; against a crop-first baseline the difference, $-2.17$ points, has an interval that reaches zero (Section~\ref{sec:res:choose}).
\item \textbf{Judge.} With the truth, choosing the best set among three calls of one model would remove 45\% of set error; selectors fitted with labels reach 16\%. Defaults chosen on other sources transfer: three-call unanimity improves set error significantly in 9 of 15 over-reporting cells and significantly worsens none, and a single-call configuration chosen on three detection sources raises macro F1 on the fourth from 42.1 to 49.1 (Sections~\ref{sec:res:select} and \ref{sec:res:transfer}).
\end{itemize}

\paragraph{Contributions.} Our contribution is a common task-loss model for visual reporting, procedures that act on its stages, and experiments that separate what information is available from the quality of the answer finally delivered. The model covers multi-record outputs and the single-record case of click grounding. Its exact relations do not prescribe one universal interface; they say what each operation can change and what must be measured to decide whether the change is useful.
\begin{enumerate}
\item \textbf{A common theory of reporting operations and task quality.} A chain from response to records, set and answer, in which task-compatible matching, scored record counts and the readout express the losses for counts, boxes, screen elements and clicks; exact relations for deletion, coordinate averaging and readout; conditions, stated as inequalities on confirmation rates, under which repetition, a new view or a second model helps, with a behavioural model of persistent errors; a constrained oracle for any declared family of candidate sets; an exact split of the error of one answer chosen among several calls, with the effect of adding a source and of switching an answer; and an impossibility result for certifying a report from outputs alone (Sections~\ref{sec:setup} and \ref{sec:theory}).
\item \textbf{Executable procedures for recovering, constructing and selecting visual records.} A default pipeline that dispatches by task and budget, and a several-model GUI-grounding tier, in which every input is named (task specification, shipped configuration, computed from the outputs, or optional labels), together with the revisions of the model that led to them (Section~\ref{sec:method}).
\item \textbf{Mechanism-resolving evidence and a fixed-protocol evaluation of complete task outputs.} Measurements on 67,364 calls to five proprietary models and one open-weight model that attribute gains and failures to single operations, and a 600-item click evaluation of the complete tier. All responses, the labels whose licences allow redistribution, a script that rebuilds the others from the official downloads, and the code that regenerates every table and figure are at \url{https://github.com/shijunzhe/vlm-set-reporting} (Sections~\ref{sec:experiments} and \ref{sec:results}).
\end{enumerate}

\section{Related Work}\label{sec:related}

\paragraph{Counting and set reporting by VLMs.} Counting is a long-standing weak spot of vision-language models \citep{paiss2023countbench,rahmanzadehgervi2024blind}, and recent work locates part of the failure after perception: count information can be present in internal representations yet misaligned with the emitted number \citep{countmisaligned2026}, and a visual counting bottleneck has been isolated in the model pipeline \citep{countbottleneck2026}. Training-free fixes divide the image and count per region \citep{qharabagh2026lvlmcount}, in the spirit of sliced inference for small-object detection \citep{akyon2022sahi}. Point-and-count supervision \citep{deitke2025molmo} and sequence-style detection \citep{chen2022pix2seq,lmmdet2025} make the list itself the output. Program-based visual reasoning counts by running code over the items a model finds \citep{gupta2023visprog,suris2023vipergpt}; we show that the same code readout repairs the counts a single model writes about its own list. We work with frozen API models and observe the same gap from the outside: the information is in the model's own list in the same response, so the fix is a readout rule rather than a probe or a training signal. We also extend the analysis from counts to boxes, screen elements and clicks, where the loss is a matching between sets.

\paragraph{Views: zoom, tiles and resolution.} Zooming in improves fine-grained perception \citep{wu2024vstar,shen2024zoomeye,zhang2025mllmsknow} and GUI grounding \citep{zoomclick2025,uizoomer2026}, and region-level consistency can be used at test time \citep{du2025guirc}. GUI-Lens lets a single general-purpose model crop from coarse to fine, with OCR and detector anchors and a verification step \citep{fu2026guilens}; Section~\ref{sec:res:choose} compares our GUI-grounding tier with a simplified crop-first baseline in its style and shows that the two err on different items. A companion study analyses when zooming in is safe for small objects \citep{shi2026zoomin}. Uncertainty-driven zoom on open 7B models found that zooming every item can hurt \citep{uizoomer2026}; on frontier API models we find two-stage zoom on every item, in the style of ReGround \citep{li2025screenspotpro}, not significantly different (GPT-6.1) or better (Sonnet 5.5, Gemini 3.8), and cheaper than zooming only on disagreement. The two observations are compatible: whether zoom hurts depends on whether the crop cuts the target and on the relative cost of the second call (Section~\ref{sec:res:view}).

\paragraph{Visual scaffolds.} Overlaid marks and coordinate grids are a popular way to help models localize \citep{yang2023setofmark,lei2025scaffold}. An overlay adds no Shannon information to the sent image and can only change how the model uses it \citep{shi2026zoomin}; we observe two such changes in current models (fewer omitted class fields for Gemini 3.8, more uses of a class name outside the vocabulary for Sonnet 5.5) and no coordinate correction, because on synthetic counting scenes their systematic coordinate offset is about 1\% of the matching tolerance.

\paragraph{Self-consistency, self-correction and cross-model agreement.} Sampling several answers and voting improves reasoning \citep{wang2023selfconsistency}, but models often fail to correct themselves \citep{huang2024cannot,madaan2023selfrefine,liao2025selfcorrect}, and their errors can be self-consistent and hard to detect \citep{tooconsistent2025,consistentwrong2026}. Disagreement between different models complements self-consistency for uncertainty estimation \citep{hamidieh2026complementing}, and multi-VLM agreement has been used to verify OCR \citep{zhang2026consensus}; we use disagreement among three frontier models' clicks as an alarm and choose among their points with a scoring model (Section~\ref{sec:method:clicks}); test-time scaling can also separate perception from reasoning \citep{sparc2026}. In set tasks these effects have a precise form: confirmation changes the set error by an exact identity, and whether a partner helps depends on its confirmation rates for the base model's true and false records, not on its overall quality (Proposition~\ref{prop:partner}).

\paragraph{Ensembles, fusion and selective prediction.} Weighted box fusion averages coordinates of matched detections \citep{solovyev2021wbf}, and detection ensembles combine models by affirmative, consensus or unanimous voting \citep{casadogarcia2020ensemble}; our three-call unanimity is the unanimous strategy applied to repeated calls, and we show that its effect splits into a selection term and an averaging term, the latter of either sign under threshold matching. Error decompositions for detection split error into classification, localization, duplicate and background parts \citep{bolya2020tide,oksuz2018lrp}; our split is instead by operation, so that each term belongs to one decision. Classical models for aggregating judges assume errors that are independent given the truth \citep{dawid1979maximum,grofman1983thirteen}, which object-level persistence violates. Work on ensemble diversity measures how correlated the members' errors are \citep{kuncheva2003diversity}, and deep ensembles obtain diversity by training members independently \citep{lakshminarayanan2017ensembles}; our members are repeated calls of fixed models, and we measure how strongly their errors on the same object are shared. Our alarms are evaluated as risk-coverage curves, as in selective classification \citep{geifman2017selective}. The matroid structure of matching rank \citep{edmonds1965transversals} and the assignment algorithm \citep{kuhn1955hungarian} underlie our identities and oracle, and one-to-one matching between predicted and true sets is also the loss of set-prediction detectors \citep{carion2020detr}.

\paragraph{What is known and what is new in the theory.} Several of our relations specialise known results to record sets. The deletion threshold is the cost-sensitive decision threshold \citep{elkan2001costsensitive}; its F1 form matches the result that the F1-optimal threshold is half the optimal F1 \citep{lipton2014f1}; diminishing coverage is the submodularity of coverage \citep{nemhauser1978submodular}; the certification bound is a two-point argument \citep{tsybakov2009nonparametric}; and the propensity model is related to labeler-expertise models \citep{whitehill2009glad}. The three propositions on choosing one answer among several calls are elementary: a partition of the miss event, the chain rule, and the law of total probability. The quantity $F$ is the miss counterpart of pass@$k$-style coverage of a candidate set; what the propositions add is the bookkeeping with representatives fixed before any choice, which makes the three terms measurable on stored calls (Section~\ref{sec:res:choose}). The set-reporting setting adds three things. First, a deleted record can lose fewer matches than it held, because a remaining record may take over its target, so the threshold concerns the marginal matching rate $p_1$ rather than precision, and $p_1$ estimated on five labelled images, with the task family's default as the starting point, predicts the effect of deletion on other images. Second, a selector and an oracle can be compared fairly only within the same feasible family, which the constrained oracle provides. Third, a view can confirm or reject only the records it contains, which restricts cross-view confirmation to each view's field. The chain itself is the unifying contribution: the same three quantities, matching size, scored record count and readout, carry these results across counts, boxes, screen elements and clicks.

\paragraph{Benchmarks.} We use SKU-110K \citep{goldman2019sku110k}, DOTA \citep{xia2018dota}, DocLayNet \citep{pfitzmann2022doclaynet}, FSC-147 \citep{ranjan2021fsc147}, PixMo-Count \citep{deitke2025molmo}, ScreenSpot-Pro \citep{li2025screenspotpro}, ScreenSpot-v2 (a corrected release of ScreenSpot; \citealp{cheng2024seeclick}), GroundCUA \citep{groundcua2026} and ScreenParse \citep{screenparse2026}, together with a synthetic counting probe whose labels are exact.

\section{Problem Setup}\label{sec:setup}

\subsection{Terms}\label{sec:setup:terms}
Table~\ref{tab:terms} defines the words used throughout the paper in plain language; the formal objects follow.

\begin{table*}[tp]
\centering\footnotesize\setlength{\tabcolsep}{4pt}\renewcommand{\arraystretch}{0.97}
\caption{Terms used in this paper.}\label{tab:terms}
\begin{tabular}{@{}>{\raggedright\arraybackslash}p{2.6cm}>{\raggedright\arraybackslash}p{12.8cm}@{}}\toprule
Term & Meaning \\\midrule
call, response & One request to a model API, and the raw text it returns; every attempted call is stored and costed. \\
interface & Everything that defines a call except the image: model, reasoning setting (\emph{lowest} or \emph{default}), output cap, prompt, output schema, and the view of the image that is sent. \\
target & A true object in the image, given by the annotation (a point or box with a class, or the box of the element a click should hit). \\
record & One object as reported by the model: a class name and a location (a point or a box). \\
list & The sequence of records in one response. A \emph{list-first} prompt asks for the list before any counts; a \emph{count-first} prompt asks for counts first. \\
stated count & The numbers the model writes for how many objects of each class it found (for example the field \texttt{counts\_by\_type}). \\
list count & The number of records of each class in the list, counted by code: the \emph{raw} list count before the alias map, the \emph{mapped} list count after it. Unqualified, it means the mapped list count. \\
readout gap & The difference between the list count and the stated count of the same response. A gap shows that the two disagree; it does not by itself say which is right. \\
alias map & A declared, task-specific mapping of class names the model uses to the task's classes (``diamond'' $\to$ ``square'' for rotated squares in our synthetic scenes). \\
decoding, status & Turning a response into records, stated counts and a status: whether the call returned, whether the text parses as the requested object, which records are valid, and four list-level \emph{degeneracy flags} (Section~\ref{sec:method}). A \emph{degenerate} list is one with at least one flag. \\
audit and delivered records & The audit collection holds every record the decoder could recover; the delivered collection holds the records that are valid under the task contract (Table~\ref{tab:contract}). Set scores evaluate the delivered collection. \\
matching, matched record & Pairing records with targets one-to-one under the task's rule (same class and within a distance or overlap threshold). The \emph{matching size} is the largest number of such pairs. A record in such a pair is \emph{matched}. \\
set error $E$, F1 & $E$ is the number of unmatched records plus the number of unmatched targets. F1 is the harmonic mean of precision and recall. \\
readout & The rule that turns a set or a response into the task answer, for example a count per class or one click point. \\
construction & A rule that builds one set from the records of several calls, for example keeping only records that all three calls agree on (\emph{unanimity}), at least two of three (\emph{majority}), both of two (\emph{intersection}), or any (\emph{union}). \emph{Averaged} coordinates average the agreeing records; \emph{deletion only} keeps the base call's own record. \\
base call, partner & In a confirmation, the call whose records may be deleted, and the call (same model or another model) used to confirm them. \\
$p_1$ & In a confirmation, the \emph{marginal matching rate} of the base call's unconfirmed records (those the partner does not report): how much the matching size falls per deleted record. It equals their precision unless another record can take over a deleted record's target. Deleting them lowers $E$ when $p_1<1/2$ (Proposition~\ref{thm:intersect}). \\
view, tile, two-stage zoom & The part of the image sent in a call. Tiles are overlapping crops sent in separate calls. Two-stage zoom (ReGround, \citealp{li2025screenspotpro}) first asks for a click on the whole screen, then asks again on a crop around that point. The \emph{apparent size} of an object is the number of visual tokens across its shorter side in the image as sent. A \emph{view planner} is any procedure that chooses the views, such as tiling or the safe decomposition of the companion paper. \\
disagreement alarm & An observable signal that two answers for the same item disagree by more than a threshold. In the risk-coverage analysis (Section~\ref{sec:res:persist}) alarmed items are set aside; in the GUI-grounding tier (Algorithm~\ref{alg:clicks}) the alarm triggers a visual comparison and every item stays in the denominator. \\
acceptance coverage & In the risk-coverage analysis, the share of items acted on without an alarm. It differs from \emph{candidate coverage}, the share of items with at least one correct candidate, which is measured with the annotations. \\
candidate, cluster, representative & For a click, a candidate is one point from one call; clusters group candidates within a distance; each cluster's representative is the member the tier would output for it (Section~\ref{sec:theory:points}). \\
scorer & In the GUI-grounding tier, the model (Gemini 3.8) that gives each candidate a hit score and compares representatives. \\
coverage and selection & What a set of calls has seen, the correct records or candidates present anywhere in their responses, and what is reported from it: the field read, the records kept, the point output. \\
$F$, $L$, $W$; rescue rate $\phi$ & For one click chosen among candidates: no correct candidate; a correct candidate but no correct representative; a correct representative not chosen (Proposition~\ref{prop:three}). $\phi$: how often an added source hits where every earlier candidate misses (Proposition~\ref{prop:coverage}). \\
source, cell & A dataset, and one model on one dataset. The six set-task sources are the four \emph{detection sources} (SKU-110K, DOTA, DocLayNet, FSC-147) and the two \emph{screen sources} (GroundCUA, ScreenParse). \\
shipped configuration & The fixed values the method uses on any new image (interface and setting per model and loss, flag thresholds, construction per loss, alarm partner, GUI-grounding tier), chosen on our evaluation data and released as a file with the code (Section~\ref{sec:method}). \\
over- and under-reporting & Task families where models mostly report false objects (detection, layout, screen parsing) or mostly miss objects (dense counting with points). The \emph{family default} for deleting unconfirmed records follows from it: delete for over-reporting families, keep for under-reporting ones. \\\bottomrule
\end{tabular}
\end{table*}

\subsection{Task}
A \emph{visual set-reporting task} $\tau$ fixes the following:
(i) a class vocabulary $\mathcal{C}_\tau$, with a declared alias map $\alpha_\tau$ from strings to $\mathcal{C}_\tau\cup\{\bot\}$ (the map is part of the task, not a global rule);
(ii) a record schema $(c,g)$ with class $c\in\mathcal{C}_\tau$ and geometry $g$ (a point, an axis-aligned box, or a click point);
(iii) a \emph{compatibility relation} $\rho_\tau(z,y)\in\{0,1\}$ between a record $z$ and a target $y$, which decides whether they may be matched;
(iv) an \emph{ignore region} $\Omega_\tau(x)$, possibly empty, in which an unmatched record is not scored;
(v) a loss $\ell_\tau$: weighted set error, F1, per-type count error, or click miss.
An image $x$ has a true target multiset $Y=Y(x)$ with $n=|Y|$.\footnote{Notation differs from the companion paper \citep{shi2026zoomin}: there $S$ is the number of visual tokens across a target side and $L$ the content of a call; here $S$ is a reported set and L1 to L4 are status layers. That paper's $S$ is the apparent size that enters the view effect $\gamma_v$ of Appendix~\ref{app:param}. Our point tolerances (20 or 30 pixels on the synthetic scenes) are about two to three times that paper's tolerance of half the target size.}
Table~\ref{tab:contract} lists, for every task in the paper, the compatibility relation used for scoring, the separate rule used to associate records across calls, and what the decoder does with records that are not valid.

\begin{table*}[tp]
\centering\small\setlength{\tabcolsep}{4pt}
\caption{Task contracts. Scoring compatibility decides matches with the truth; association decides which records of different calls are the same object, and uses no labels. Coordinates are mapped from each model's native frame to image pixels before scoring. The detection rows and the multi-class count row use overlapping datasets but different images and different contracts.}\label{tab:contract}
\begin{tabular}{@{}>{\raggedright\arraybackslash}p{2.6cm}>{\raggedright\arraybackslash}p{3.9cm}>{\raggedright\arraybackslash}p{3.4cm}>{\raggedright\arraybackslash}p{5.0cm}@{}}\toprule
Task & Scoring compatibility & Association across calls & Records that are not valid \\\midrule
SCP (synthetic, 100 images) & same class, within 30 px & not used & class unknown after the alias map: kept for status, excluded from the delivered set \\
SCP-dense (50 images) & same class, within 20 px & not used & as for SCP \\
Counts, one class (PixMo-Count, FSC-147; Section~\ref{sec:res:readout}) & absolute error of the count against the annotated number; no matching & not used & every decoded point is counted \\
Counts, several classes (30 DOTA windows, 50 DocLayNet pages; Section~\ref{sec:res:readout}) & per-class count error; DOTA objects marked difficult or less than half inside the window are not targets and there are no ignore regions, so records on them are counted like any other record; list recall (diagnostic only): a record centre inside a true box of the same class, one to one & not used & class unknown after the declared alias map: kept in the response, not counted \\
Detection boxes (SKU-110K, DOTA, DocLayNet) & same class, IoU $\ge 0.75$; unmatched records in ignore regions not scored & same class, IoU $\ge 0.5$ & label outside the vocabulary or box outside the image: kept for audit, excluded from the delivered set, counted as invalid \\
FSC-147 points & within half the mean short side of the three exemplar boxes & within the same distance & point outside the image: kept for audit, excluded and counted as invalid \\
Screen parsing (GroundCUA, ScreenParse) & class-agnostic, IoU $\ge 0.5$ & IoU $\ge 0.5$ & coordinates clamped to the image; boxes of zero area dropped \\
Clicks (ScreenSpot-Pro, ScreenSpot-v2) & point inside the target box & alarm: distance above a share of the longer image side & missing point counted as a miss \\\bottomrule
\end{tabular}
\end{table*}

\subsection{Reporting chain}
One call with interface $\eta$ consists of a model, a reasoning setting, an output cap, a prompt, an output schema and a view (crop or tile set). The chain (Figure~\ref{fig:chain}) is
\begin{align*}
&H_\eta\sim K_\eta(\cdot\mid x,\tau)\xrightarrow{D_{\eta,\tau}}(Z^{\mathrm{all}},Z,d,\sigma)\\
&\qquad\xrightarrow{C}S\xrightarrow{r_\tau}\hat y.
\end{align*}
$H_\eta$ is the raw response text, which is always stored. $D_{\eta,\tau}$ is a deterministic decoder. It returns the audit collection $Z^{\mathrm{all}}$ of every recoverable record; the delivered collection $Z\subseteq Z^{\mathrm{all}}$ of records that are valid under the task contract (Table~\ref{tab:contract}), with aliases normalised by $\alpha_\tau$ and coordinates in image pixels; the stated count vector $d\in\mathbb{N}^{\mathcal{C}_\tau}\cup\{\varnothing\}$; and a status vector $\sigma$ that records, among other things, how many records were excluded. Set scores in this paper evaluate the delivered collection; where exclusion is large we also report a sensitivity that keeps the excluded records and scores them as unmatched (Section~\ref{sec:res:degen}). $C$ is a construction that maps the delivered records of $K\ge1$ calls, possibly with different interfaces, to a final set $S$ inside a declared feasible family. $r_\tau$ is the task readout, for example the per-class count of $S$, or $d$ when a declared fallback applies. The objective of a procedure $\pi$ is
\[
\mathcal{R}(\pi)=\mathbb{E}\,\ell_\tau(\pi(x),Y)+\lambda\,\mathbb{E}\,\mathrm{cost}(\pi;x),
\]
where the cost counts every attempted call, including failed, truncated or retried ones.

\subsection{Records, clusters and feasible sets}\label{sec:setup:feasible}
Calls $1,\dots,K$ give delivered collections $Z_1,\dots,Z_K$. A label-free correspondence merges them into clusters $\mathcal{Q}=\{Q_1,\dots\}$: it first maximizes the number of associated pairs under the association rule of Table~\ref{tab:contract}, then their overlap, against cluster representatives. A cluster has a support set $\mathrm{sup}(Q)\subseteq\{1,\dots,K\}$ (the calls that reported it), member records, and an averaged record $\bar z_Q$ (the coordinate mean of its members, with the class of the first member). A construction chooses, for each cluster, either nothing or one representative from a declared candidate set $A(Q)$, so its feasible family is
\[
\SA=\big\{\textstyle\bigcup_Q s_Q:\ s_Q\in\{\varnothing\}\cup\{\{a\}:a\in A(Q)\}\big\}.
\]
Support-$k$ rules use $A(Q)=\{\bar z_Q\}$ and keep $Q$ iff $|\mathrm{sup}(Q)|\ge k$. Deletion-only intersection uses the base call's own record. A record gate uses $A(Q)=\{\bar z_Q\}\cup\mathrm{members}(Q)$. All oracles and fitted selectors compared with a construction use the same $\SA$.

\subsection{Output contract and status}\label{sec:setup:contract}
The decoder reports four status layers for every call. They are distinct, and passing a lower layer never implies a higher one.
\textbf{L1 transport}: the call returned, and its finish reason (stop, length) is recorded.
\textbf{L2 outer syntax}: the whole response parses as the requested JSON object after stripping a code fence.
\textbf{L3 record validity}: which records were recovered (strict parse, or a declared salvage of individual objects when the whole response does not parse), which of them are valid under the task contract, and four list-level degeneracy flags (Section~\ref{sec:method}).
\textbf{L4 task completeness}: whether $S$ contains every target. L1 to L3 are checked by code; L4 cannot in general be checked without labels (Section~\ref{sec:theory:cert}).
The procedures return $S$, $\hat y$ and $\sigma$ for every input, never drop a call from cost or denominator accounting, and report the strict-JSON failure rate, so the share of salvaged responses is visible.

\subsection{Losses}\label{sec:setup:loss}
For a scored collection $S$, let $r(S)$ be the matching size: the largest number of disjoint pairs $(z,y)$ with $\rho_\tau(z,y)=1$. Let $c(S)$ be the number of scored records: those matched in the maximum matching chosen by the evaluation code or lying outside $\Omega_\tau$. Without an ignore region, $c(S)=|S|$. With one, $c$ can depend on which maximum matching is chosen; the relations below hold for any fixed choice applied consistently, since they are computed from the same $c$ and $r$. Ignore regions occur in 34 of 240 detection items and in no screen item. The set losses are functions of $r$, $c$ and $n$; the count losses also depend on the readout. With costs $\cfp,\cfn>0$,
\begin{align*}
E_w(S)&=\cfp[c(S)-r(S)]+\cfn[n-r(S)],\\
\mathrm{F1}(S)&=2\,r(S)/(n+c(S)),
\end{align*}
with $\mathrm{F1}=1$ when $n=c(S)=0$. $E$ denotes $E_w$ with $\cfp=\cfn=1$; recall is $r(S)/n$. F1 in our tables is the mean of per-image F1. The per-type count error of a readout $\hat y$ is $\sum_c|\hat y_c-n_c|$, and the click loss is $1-\mathbb{1}[\hat p\in\mathrm{box}(y)]$. Every reported number names the number of calls $K$, the interface, the setting, and the loss.

\section{Theory}\label{sec:theory}

One principle runs through the results on operations. A library of calls fixes what can be reported, its coverage; a rule applied to the library decides what is reported, its selection. On a fixed library, an operation lowers the loss exactly when, on the records or items it changes, the errors it repairs outweigh the correct outputs it breaks, at the exchange rate the loss sets. Section~\ref{sec:theory:identities} states the exact relations for record sets: deleting records, averaging their coordinates, and reading a different field of one response. Section~\ref{sec:theory:behaviour} gives a behavioural model of how models report objects across repeated calls, views and models, which says when a partner adds coverage; unlike the relations, it is an assumption to be checked. Section~\ref{sec:theory:select} bounds what selection over a fixed library can reach. Section~\ref{sec:theory:points} treats a single answer chosen among several calls: the three places where a miss can arise, the coverage an added source brings, and the net effect of switching an answer, which is the principle's form for clicks. Section~\ref{sec:theory:cert} states what outputs alone cannot certify. Table~\ref{tab:theory_use} lists, for each result the method uses, the quantity it measures, the decision it makes and where it is checked. Proofs and supporting results are in Appendix~\ref{app:proofs}.

\begin{table*}[t]
\centering\footnotesize\setlength{\tabcolsep}{3.5pt}
\caption{What each result is used for: the quantity it measures, the decision it makes in the method, and the check in Section~\ref{sec:results}.}\label{tab:theory_use}
\begin{tabular}{@{}>{\raggedright\arraybackslash}p{2.6cm}>{\raggedright\arraybackslash}p{3.2cm}>{\raggedright\arraybackslash}p{4.5cm}>{\raggedright\arraybackslash}p{4.9cm}@{}}\toprule
Result & Quantity & Decision & Check \\\midrule
Intersection threshold (Prop.~\ref{thm:intersect}) & marginal matching rate $p_1$ of unconfirmed records & confirm and delete for set error on over-reporting sources, keep a single call on under-reporting sources; label rule of Step~7 & from 5 labelled images the threshold predicts the sign on the other images in 83\% of draws; the Step~7 rule lowers the loss against the better decision from 0.43 (family default alone) to 0.17 errors per image (\S\ref{sec:res:set}) \\
Geometry term (Prop.~\ref{prop:geometry}) & selection and geometry terms & coordinate averaging is a separate switch: on by default, the deletion-only variant in the menu & averaging helps in 37 of 54 pairs and hurts in 13; the threshold's only two failures on averaged intersections are geometry reversals (\S\ref{sec:res:set}) \\
Two readouts (Prop.~\ref{prop:readout}) & gap between stated count and list count, without labels & bounds what switching readouts can gain, without labels & on aerial and document images, for GPT-6.1 and Sonnet 5.5, the gap is 0.1 to 7.4 against errors of 5 to 46, and switching readouts changes little (\S\ref{sec:res:readout}) \\
Partner (Prop.~\ref{prop:partner}) & confirmation rates $h_+$, $h_-$ & a partner must reject more false than true records; cross-model partners in the menu & cross-model partners separate the two better in 32 of 36 comparisons (\S\ref{sec:res:set}) \\
Coverage (Prop.~\ref{prop:coverage}) & rescue rate $\phi$ on the menu's misses & three different models, not three calls of one & a GPT-6.1 repeat rescues 18\%, then 7.5\%; Sonnet 5.5 29\%, Gemini 3.8 37\%; given GPT-6.1's 82 misses, rates from the 260 GUI items give 46 items with no correct candidate among the 600, against 43 observed (\S\ref{sec:res:choose}) \\
Three sources (Prop.~\ref{prop:three}) & $F$, $L$, $W$ & the steps of Algorithm~\ref{alg:clicks}; three sources, not four & a fourth source lowers $F$ from 43 to 36 and raises $L$ from 3 to 7 and $W$ from 10 to 11 (\S\ref{sec:res:choose}) \\
Switch (Prop.~\ref{prop:switch}) & $e_T$, $r_{\mathrm{fix}}$, $r_{\mathrm{brk}}$ on the items acted on & compare clusters only on alarmed items; no switch to the crop-first point & the score alone misses 5.8\% of single-cluster and 31.7\% of multi-cluster items; switching can gain at most 10 items (\S\ref{sec:res:choose}) \\
Alarm and no certificate (Props.~\ref{prop:alarm}, \ref{prop:nocert}) & disagreement between reports & flag errors without labels; choose configurations with labels of other sources & alarms evaluated by risk and acceptance coverage (\S\ref{sec:res:persist}); configurations by leaving one source out (\S\ref{sec:res:transfer}) \\
Risk ordering (Prop.~\ref{prop:order}) & gap between a selector and the constrained oracle & selectors fitted with labels are optional (Step~7), not a default & median gap 26.1 points under cross-fitting (\S\ref{sec:res:select}) \\
\bottomrule
\end{tabular}
\end{table*}

\subsection{Exact relations between operations and losses}\label{sec:theory:identities}

\begin{proposition}[Deletion and addition]\label{prop:identity}
Let $S\subseteq A$, let $\Delta c=c(A)-c(S)$ be the net decrease in scored records and $\Delta r=r(A)-r(S)$ the number of matches lost. Then exactly
\[
E_w(A)-E_w(S)=\cfp\,\Delta c-(\cfp+\cfn)\,\Delta r,
\]
and in particular $E(A)-E(S)=\Delta c-2\Delta r$. Addition is the same relation read in reverse.
\end{proposition}

\begin{proposition}[Intersection threshold]\label{thm:intersect}
Deleting $A\setminus S$ lowers weighted set error if and only if $\cfp\,\Delta c>(\cfp+\cfn)\,\Delta r$, that is, for $\Delta c>0$, iff $\Delta r/\Delta c<\cfp/(\cfp+\cfn)$. Adding a non-empty set lowers weighted set error iff $(\cfp+\cfn)$ times the matches gained exceeds $\cfp$ times the scored records added.
\end{proposition}

In words: deleting records helps when few enough of them were matched. Over a population of images with $\mathbb{E}[\Delta c]>0$, the expected change has the sign of $\cfp/(\cfp+\cfn)-p_1$, where $p_1:=\mathbb{E}[\Delta r]/\mathbb{E}[\Delta c]$ is the ratio of expected matches lost to expected records removed. We call $p_1$ the \emph{marginal matching rate} of the deleted records; it is not the average of per-image precisions, and it equals the precision of the deleted records only when no remaining record can take over a deleted record's match. For $E$ the threshold is $1/2$. This is the cost-sensitive decision threshold \citep{elkan2001costsensitive} applied to the records a partner does not confirm; what is specific to record sets is that $\Delta r$ can be smaller than the number of matched records deleted, so the relevant rate is $p_1$ and not their precision. The proposition does not assume that the confirming report comes from the same model, the same view, or any probability model. It concerns deletion only: the retained records keep their coordinates.

For F1 the same argument gives a per-image threshold of $\mathrm{F1}(A)/2$ in place of $1/2$ (Corollary~\ref{cor:f1} in Appendix~\ref{app:proofs}), the marginal form of the known result that the F1-optimal threshold of a calibrated classifier is half its optimal F1 \citep{lipton2014f1}. Because $\mathrm{F1}(A)/2\le1/2$, a deletion that raises F1 on an image also lowers $E$ there, but not conversely. This agrees with the defaults of Section~\ref{sec:res:transfer}: confirmation for set error on over-reporting sources, a single call for F1. In the deletion case, the threshold is the same exchange rate as the price of a false output in the companion analysis of decomposition \citep{shi2026zoomin}.

\begin{proposition}[Geometry term]\label{prop:geometry}
Let $S_0\subseteq A$ be the confirmed records with their original coordinates and $S_f$ the same clusters with averaged coordinates. Then
\begin{align*}
E(A)-E(S_f)=\ &\underbrace{[E(A)-E(S_0)]}_{\text{selection (Prop.~\ref{prop:identity})}}\\
&+\underbrace{[E(S_0)-E(S_f)]}_{\text{geometry}},
\end{align*}
and the geometry term equals $[c(S_0)-c(S_f)]-2[r(S_0)-r(S_f)]$.
\end{proposition}
By convexity, the mean of a cluster's members is no farther, in mean squared distance, from any fixed target than the members are on average. That does not make every averaged record closer than its best member, and matching uses a threshold, so the geometry term can have either sign. On DocLayNet with GPT-6.1 confirmed by Sonnet 5.5 (60 images), $E(A)=8.567$, $E(S_0)=7.767$ and $E(S_f)=8.867$: selection gains $0.800$, geometry loses $1.100$, and the intersection with averaged coordinates is worse than the single call although deletion alone is better.

\begin{proposition}[Two readouts of one response]\label{prop:readout}
Let a decoded response have delivered records $Z$ and stated counts $d\neq\varnothing$, and let $\mathrm{lc}(Z)_c=\#\{z\in Z:\mathrm{class}(z)=c\}$ be the list count. Both readouts come from the same response, so the difference between their errors $\|d-n\|_1$ and $\|\mathrm{lc}(Z)-n\|_1$ isolates the readout and holds generation fixed. The gap $\|d-\mathrm{lc}(Z)\|_1$ is observable without labels, and
\[
\max\{\|d-n\|_1,\|\mathrm{lc}(Z)-n\|_1\}\ \ge\ \tfrac12\|d-\mathrm{lc}(Z)\|_1,
\]
\[
\big|\,\|d-n\|_1-\|\mathrm{lc}(Z)-n\|_1\big|\ \le\ \|d-\mathrm{lc}(Z)\|_1 .
\]
\end{proposition}
A positive gap shows that at least one readout is wrong; it does not show which. A small gap shows that no choice between the two readouts can change the error by more than the gap, whatever the truth: choosing a readout can never gain more than the gap, and errors that both readouts share, such as objects missing from the list and from the statement alike, are beyond its reach. With a list count of 9 and a stated count of 10, the gap is 1 whether the truth is 9 or 10. Which readout to use is therefore an empirical question per model and task family; the method's rule (list count unless the list is degenerate) is supported by the measurements of Section~\ref{sec:res:readout}, not by this proposition.

\subsection{Behaviour under repetition, changed views and changed models}\label{sec:theory:behaviour}

\paragraph{Joint reporting law.} Fix a complete interface $\eta$: model, prompt, schema, setting, cap and view. A call with interface $\eta$ reports a true object $o$ with probability $q_{o,\eta}$, and a candidate false object $f$ (a location where false records tend to appear) with probability $q_{f,\eta}$. Given these propensities, repeated calls with the same $\eta$ are independent. This joint law, with $q$ free to depend on every part of $\eta$, is the behavioural model; the consequences below use only it, together with conditional independence given the propensities for calls with different interfaces wherever that is stated, and not the explanatory logit parametrization (object, model and view terms) of Appendix~\ref{app:param}. A reported record contributes to the loss only through the matching: emitting a record near $o$, producing a match, and adding a match that no other record already provides are different events.

Three consequences are used in the paper.
\emph{Persistence}: for fixed $\eta$, the probability of missing $o$ in all $K$ calls is $(1-q_{o,\eta})^K$, and averaged over objects it is $\mathbb{E}[(1-q_{o,\eta})^K]$, which tends to the share of objects with $q=0$. The decay is slow when $q$ is small for some objects, so at a fixed budget many misses repeat.
\emph{Diminishing coverage}: for calls that are exchangeable given the image, the expected number of targets matched by the union of $K$ calls is concave in $K$; this follows from the submodularity of coverage \citep{nemhauser1978submodular} and of matching rank (Lemma~\ref{lem:rank}). The statement concerns raw coverage, not the loss of a constructed set: repeated calls can still remove false records that do not repeat. For example, with 20 true records that are always reported and 80 candidate false records reported independently with probability 0.2, one call has expected $E=16$ and three-call unanimity has expected $E=0.64$.
\emph{Confirmation rates}: if the records of two calls refer to the same candidate objects and the two calls are conditionally independent given the propensities, a partner call $B$ confirms a record of call $A$ with probability
\[
h_\pm=\frac{\mathbb{E}[q_Aq_B\mid\text{true / false}]}{\mathbb{E}[q_A\mid\text{true / false}]},
\]
with $h_+$ for true records and $h_-$ for false ones. For a second call with the same interface $q_B=q_A$, so $h=\mathbb{E}[q_A^2]/\mathbb{E}[q_A]$. A review call that receives the first answer has a different interface and is not covered by this formula.

\begin{proposition}[When a different partner is better]\label{prop:partner}
Assume deletion-only confirmation, no match takeover (so $\Delta r$ equals the number of true records deleted), no ignore region or all of $A$'s false records outside it, and expected counts $n_+$ (true) and $n_-$ (false) among $A$'s records, with $h^B_\pm$ averaged over these records. Then the expected reduction in set error $E$ (equal costs) from replacing $A$ by $A\cap B$ is
\[
G(B)=n_-\,(1-h_-^B)-n_+\,(1-h_+^B),
\]
and a partner $B$ beats a same-interface partner $A'$ iff $n_-\,(h_-^{A'}-h_-^{B})>n_+\,(h_+^{A'}-h_+^{B})$.
\end{proposition}
The partner must reject more of $A$'s false records than it loses of $A$'s true ones, weighted by their counts. Neither similar overall quality nor a larger separation $h_+-h_-$ is necessary or sufficient: with $n_+=100$ and $n_-=10$, the pair $(h_+,h_-)=(0.95,0.85)$ gives $G=-3.5$, while the more separated pair $(0.5,0)$ gives $G=-40$; and a weaker partner whose errors avoid $A$'s false records can satisfy the inequality. Without the assumptions, the exact relation of Proposition~\ref{prop:identity} replaces these expectations.

\paragraph{Reports from several views.} Nothing in the chain assumes that a report comes from one call on the whole image. A view is part of the interface $\eta$, so a view planner (tiling, the safe decomposition of the companion paper, zoom) produces calls like any other, and their records, mapped to image coordinates, enter the same constructions and losses. Merging the reports of several views is itself a construction: the union of their records, with duplicates in the overlaps removed by the association rule. Confirmation across views is covered by Proposition~\ref{thm:intersect} and Proposition~\ref{prop:partner} with one change. A view can report only objects that it contains, so for a record lying wholly inside no partner view, $h_+^B=h_-^B=0$: deleting it is a decision about that record's own $p_1$, on which the partner carries no information. A view should therefore confirm or reject only the records it contains. A different view changes the propensities, but it need not share fewer persistent misses: the coverage it adds splits into a gain in how often objects are reported and a gain in complementarity, as for the rescue rate after Proposition~\ref{prop:coverage} (worked example in Appendix~\ref{app:param}). Whether a view helps as a partner is again the inequality of Proposition~\ref{prop:partner}, which also counts its false records.

\begin{proposition}[Alarm from disagreement]\label{prop:alarm}
Consider a point task in which a record matches a target of the same class within distance $\theta$, with no ignore region, and let $E$ be its set error. Let $U_{2\theta}(A,B)=|A|+|B|-2\,m_{2\theta}(A,B)$, where $m_{2\theta}$ is the matching size between two reports when records of the same class are paired within $2\theta$. Then
\[
E(A)+E(B)\ \ge\ U_{2\theta}(A,B),
\]
hence $\max\{E(A),E(B)\}\ge\tfrac12\,U_{2\theta}(A,B)$.
\end{proposition}
The bound needs no labels: when $U>0$, at least one of the two reports is wrong, although the bound does not say which. Constructed instances attain it, and it can be loose: two identical wrong reports give $U=0$. The click alarm in our experiments (distance above a share of the longer image side) is an empirical alarm for the hit-in-box loss, evaluated by risk-coverage curves.

\subsection{Construction and selection}\label{sec:theory:select}

Fix a feasible family $\SA$ and a loss. Define the oracle risk $R_{\mathrm{or}}=\mathbb{E}[\min_{S\in\SA}\ell(S,Y)]$, which uses the truth to pick the best set; the best risk $R^*_\Phi=\inf_g\mathbb{E}[\ell(g(\Phi),Y)]$ of a rule that maps features $\Phi$ computed from outputs alone into $\SA$; and the risk $R_{\hat g}$ of a fitted rule.

\begin{proposition}[Risk ordering]\label{prop:order}
$R_{\mathrm{or}}\le R^*_\Phi\le R_{\hat g}$, and
\begin{align*}
R_{\hat g}-R_{\mathrm{or}}=\ &\underbrace{(R^*_\Phi-R_{\mathrm{or}})}_{\text{information not in }\Phi}\\
&+\underbrace{(R_{\hat g}-R^*_\Phi)}_{\text{selector and estimation}}.
\end{align*}
\end{proposition}
An observed gap between a selector and the oracle bounds the sum but does not apportion it. A rule whose fitting uses labels is a \emph{label-fitted output-feature selector}: it reads no labels at inference, but its fitting does. The gap concerns a fixed library of calls; it says nothing about what additional calls or a different library would add.

For the family with at most one representative per cluster, which the selectors of Section~\ref{sec:res:select} use, the oracle is a maximum matching between clusters and targets and is computed exactly (Proposition~\ref{prop:oracle} in Appendix~\ref{app:proofs}); unrestricted matching over all records can exceed it but is not attainable within the family (Section~\ref{sec:res:select} reports both).

\subsection{Choosing one answer among several calls}\label{sec:theory:points}

A click has one answer, so the constructions above reduce to choosing one point. Let several calls (models, settings, views) give an item a menu $A$ of candidate points. Group the candidates into \emph{clusters}, the connected components of the graph that joins two candidates within distance $\delta$; fix one \emph{representative} per cluster before any choice is made; and let a rule output one representative. For clicks these clusters replace the association rule of Section~\ref{sec:setup:feasible}. The loss is the miss, $\mathbb{1}[\hat y\notin\text{target}]$. In the GUI-grounding tier of Section~\ref{sec:method:clicks} a scoring model, the \emph{scorer}, gives each candidate a hit score and compares representatives; Section~\ref{sec:res:choose} reports it on 600 click items.

\begin{proposition}[Three sources of a miss]\label{prop:three}
For every rule that outputs a representative, the miss rate is exactly $F+L+W$, where $F$ is the probability that no candidate in $A$ hits, $L$ that some candidate hits but no representative does, and $W$ that some representative hits but the rule outputs one that misses.
\end{proposition}

The terms locate where a miss arises; they are not independent effects of interventions. With the candidate menu fixed, changing the representatives leaves $F$ unchanged but can move a miss between $L$ and $W$. With candidates and representatives fixed, changing only the final choice acts on $W$. Only changing or enlarging the candidate menu can lower $F$, and doing so can change all three terms at once, because a new candidate can also change the clusters, the representatives and the comparison. Section~\ref{sec:res:choose} measures the three terms separately when a source is added. How much an added candidate lowers $F$ needs no independence assumption:

\begin{proposition}[Coverage of an added source]\label{prop:coverage}
Let $F$ be the probability that every candidate of a menu misses, and $\phi=\Pr(\text{an added candidate hits}\mid\text{every earlier candidate misses})$. With the added candidate, the probability becomes $F(1-\phi)$. If the candidates are conditionally independent given per-item hit probabilities $q_1,\dots,q_k$, then $F=\mathbb{E}[\prod_{j=1}^{k}(1-q_j)]$.
\end{proposition}

The relevant quantity is the rescue rate $\phi$ on the items the existing menu misses, not the added source's own accuracy or its correlation alone. For the indicators $I$ that some earlier candidate hits and $J$ that the added candidate hits, with $\Pr(I=0)>0$,
\[
\phi=\Pr(J=1\mid I=0)=\mathbb{E}[J]-\frac{\mathrm{Cov}(I,J)}{1-\mathbb{E}[I]},
\]
the binary form of the decomposition into reporting rate and complementarity for views (Appendix~\ref{app:param}). At equal hit rates, lower covariance raises the rescue rate; when hit rates differ, both terms matter. Persistence makes the covariance of a repeated call with the same interface large, which is why a repeat tends to rescue little, but a weak partner can rescue less than a repeat. Section~\ref{sec:res:choose} therefore measures $\phi$ for each tested source.

\begin{proposition}[Net effect of a switch]\label{prop:switch}
Let a base rule output $b$, and let an action replace $b$ by $g$ on a set $T$ of items decided from outputs. With $e_T=\Pr(b\text{ misses}\mid T)$, the repair rate $r_{\mathrm{fix}}=\Pr(g\text{ hits}\mid b\text{ misses},T)$ and the break rate $r_{\mathrm{brk}}=\Pr(g\text{ misses}\mid b\text{ hits},T)$, the miss rate falls by exactly
\[
\Pr(T)\,\big[\,e_T\,r_{\mathrm{fix}}-(1-e_T)\,r_{\mathrm{brk}}\,\big].
\]
If, within an event $U\supseteq T$, membership in $T$ is independent of whether $b$ misses, then $e_T=\Pr(b\text{ misses}\mid U)$.
\end{proposition}

This is the click counterpart of Proposition~\ref{thm:intersect}: an action pays when the errors it repairs outweigh the correct answers it breaks, counted on the items it touches. Disagreement alone does not say which candidate is wrong (Proposition~\ref{prop:alarm} bounds the sum of two errors, not the base error), so where to act is an empirical question. Where the base miss rate $e_T$ is low, as on items whose candidates agree, an action must break very few correct answers to pay: with $e_T=5.8\%$, fewer than about 3.1\% if it repairs half of the misses (Section~\ref{sec:res:choose} measures $e_T$ with and without an alarm). The scorer is a design choice of the configuration and not a consequence of the theory: choosing the best of $N_{\mathrm{s}}$ candidate scorers to within a miss-rate gap $\varepsilon_{\mathrm{s}}$ with probability $1-\beta_{\mathrm{s}}$ takes on the order of $\log(N_{\mathrm{s}}/\beta_{\mathrm{s}})/\varepsilon_{\mathrm{s}}^2$ labelled items under independent sampling, so the pipeline ships one scorer and does not choose it per task.

\subsection{What outputs alone can and cannot certify}\label{sec:theory:cert}

\begin{proposition}[No output-only certificate]\label{prop:nocert}
Fix a reported set $S$, a loss $L$ and a tolerance $\epsilon$. Suppose two admissible worlds induce the same law $P$ of the observable output history $H$, while $L(S,Y_0)\le\epsilon$ in the first and $L(S,Y_1)>\epsilon$ in the second. Then every output-only acceptance rule $a(H)\in[0,1]$, possibly randomized, accepts with the same probability in both worlds, so no such rule can accept with probability at least $1-\beta$ in the first world and at most $\alpha$ in the second if $\alpha+\beta<1$. More generally, $P_1(\text{accept})\ge P_0(\text{accept})-\mathrm{TV}(P_0,P_1)$, where $\mathrm{TV}$ is the total variation distance.
\end{proposition}
The same argument applies to any property of $Y$ on which two such worlds disagree, for example whether $L(S,Y)<L(S',Y)$ for two reports $S$ and $S'$, so which of two reports is more accurate cannot be certified from outputs alone either. The argument is the two-point method of minimax lower bounds \citep{tsybakov2009nonparametric}; methods that estimate accuracy from the agreement of several classifiers without labels \citep{platanios2014unlabeled} rely on assumptions, such as weakly dependent errors, that rule such pairs out. Such pairs arise from objects that a model rarely reports. If the bad world adds one target that each of $K$ calls reports with probability at most $\vartheta$, given the earlier output, and the target does not change the rest of the output, then $\mathrm{TV}(P_0,P_1)\le K\vartheta$. Any test's advantage in telling the two worlds apart, the difference between its acceptance probabilities in them, is then at most $K\vartheta$ (with equal priors, the best probability of naming the right world is $(1+\mathrm{TV})/2$), and no test is reliable uniformly over admissible classes in which $\vartheta$ can be arbitrarily small. The practical reading: agreement alone does not certify correctness. This does not prevent running a default procedure, ordering interfaces under explicit assumptions tested on other data, as the companion paper does for decomposition \citep{shi2026zoomin}, or operating a configuration chosen on source data without new labels.

\begin{corollary}[Sign of a count gap]\label{cor:nonid}
Suppose that, for a given report $S$, some admissible world has $|S|>n$, and that the admissible worlds are closed under adding any number of targets that no interface reports and that do not change the rest of the output. Then the sign of $|S|-n$ is not determined by the outputs: that world has the same output law as one in which $S$ under-counts. (This differs from the sign of a recall gain between two interfaces, which assumptions such as monotonicity in tokens per object can fix.)
\end{corollary}
Without further information, neither deleting unconfirmed records nor keeping them dominates for a count objective. Our count procedure keeps them, a conservative choice whose quality we evaluate empirically.

\section{Method: A Default Pipeline with Stated Inputs}\label{sec:method}

The default pipeline of FSR turns an image, a task specification and a budget into a final set, a task answer, a status and an alarm score; it needs no labels from the target task. A summary table in Section~\ref{sec:discussion} gives when each component helps, its measured effect and its cost. Every input it uses comes from one of three named sources, or is computed from the current responses: \src{task} the task specification and budget given by the user (classes, alias map, matching rule, loss); \src{config} the shipped configuration, a fixed file released with the code (\texttt{config/default\_pipeline.json}) whose values were chosen on our evaluation data and are carried unchanged to new images; and \src{labels} optional labelled images. No step requires the user to know $p_1$, a partner's quality or the current F1. Algorithm~\ref{alg:pipeline} gives the procedure; its comments name the steps described below. Steps 1 to 6 repair the report, the optional label rule of Step 7 judges between configurations, and the alarm of Step 8 detects errors. Section~\ref{sec:method:clicks} adds the GUI-grounding tier, a second tier for clicks that combines several models, and Section~\ref{sec:method:revise} describes how the model behind these steps was revised.

\begin{algorithm*}[t]
\caption{FSR default reporting pipeline. Tags give the source of each input: \src{task} the user's task specification and budget, \src{config} the shipped configuration file, \src{labels} optional labelled images; everything else is computed from the current responses. All calls append their response status and incurred cost to a shared execution trace; every return includes the accumulated trace and its cost $c$. A missing click is represented by an empty set and an explicit no-point status. Comments name the step paragraphs of Section~\ref{sec:method}.}\label{alg:pipeline}
\small
\begin{algorithmic}[1]
\Require image $x$; task $\tau=(\mathcal{C}_\tau,\alpha_\tau,\rho_\tau,\Omega_\tau,\ell_\tau)$ and budget $B$ per item \src{task}; configuration $\Theta$ \src{config}; optional labelled images $\mathcal{D}_L$ \src{labels}
\Ensure final set $S$, answer $\hat y$, status $\sigma$ (readout branch, flags, recovery, views), alarm score $a$ (\emph{not computed} without a partner), cost $c$ of every attempt
\If{$|\mathcal{D}_L|\ge3$} \Comment{Step 7, before inference; never required}
  \State replace $\Theta$ by a menu configuration only if its mean paired advantage over $\Theta$ on $\mathcal{D}_L$ minus one standard error is positive
\EndIf
\State select the interface $\eta$ (model, setting, prompt, schema), the number of calls $K$ and the views $\mathcal{V}$ from $(\tau,B,\Theta)$ \Comment{Step 1}
\If{$\tau$ asks for one click}
  \If{$B$ covers the planned worst case of Algorithm~\ref{alg:clicks}}
    \State $(\hat p,\sigma,\mathit{alarm})\gets$ Algorithm~\ref{alg:clicks}$(x,\tau,B,\Theta)$;\ \ $\hat y\gets\hat p$;\ \ $a\gets\mathit{alarm}$ (1 or 0)
    \State $S\gets\{\hat p\}$, or $\varnothing$ if there is no point;\ \ \Return $S,\hat y,\sigma,a,c$
  \EndIf
  \State $\hat p\gets\textsc{Zoom}(x,\eta)$;\ \ $\hat y\gets\hat p$;\ \ $S\gets\{\hat p\}$, or $\varnothing$ with status no point \Comment{Step 4; no point is a miss}
\Else
  \For{$k=1,\dots,K$ and each view $v\in\mathcal{V}$} \Comment{$\mathcal{V}=\{x\}$ unless a view planner is configured}
    \State $R_{k,v}=(Z^{\mathrm{all}}_{k,v},Z_{k,v},d_{k,v},\sigma_{k,v})\gets\textsc{Report}(v,\eta)$, records mapped to image coordinates \Comment{Steps 2 and 3}
  \EndFor
  \State $R_k\gets$ merge of $R_{k,v}$ over $v$, dropping a record associated with one already kept from another view
  \State $S\gets C(R_1,\dots,R_K)$ with the construction $\Theta$ assigns to $\ell_\tau$; a view confirms only records inside it \Comment{Step 5}
  \State $\hat y\gets r_\tau(S)$, for counts the list count of $S$ \Comment{Step 6}
  \If{$\tau$ is a count task, $|\mathcal{V}|=1$, $\sigma_{1,x}$ flags the list of $R_1$ and its stated counts $d_{1,x}\neq\varnothing$}
    \State $\hat y\gets d_{1,x}$;\ \ readout branch $\gets$ \texttt{count\_from\_statement}
  \EndIf
  \State $\sigma\gets$ the readout branch, the call statuses $\sigma_{k,v}$ (flags, recovery used) and the views used
\EndIf
\If{a partner model is configured} \Comment{Step 8, alarm}
  \State $\hat y'\gets$ the same steps run by the partner with one call per view;\ \ $a\gets\mathrm{dis}_\tau(\hat y,\hat y')$
\Else
  \State $a\gets$ \emph{not computed}
\EndIf
\State \Return $S,\hat y,\sigma,a,c$
\Statex
\Function{Report}{$v,\eta$} \Comment{one call with bounded recovery}
  \State $H\gets$ call $\eta$ on $v$; on a transport failure (L1) resend once; if it fails again, \Return an empty report with status L1
  \State $(Z^{\mathrm{all}},Z,d,\sigma)\gets D_{\eta,\tau}(H)$ \Comment{$d$: stated counts; $\sigma$: this call's status, with the four degeneracy flags}
  \If{$\sigma$ is flagged or the response has no readable answer}
    \State $H'\gets$ one follow-up that attaches $H$; \textbf{if} $D_{\eta,\tau}(H')$ is readable and unflagged \textbf{then} \Return $D_{\eta,\tau}(H')$
  \EndIf
  \State \Return $(Z^{\mathrm{all}},Z,d,\sigma)$
\EndFunction
\Statex
\Function{Zoom}{$x,\eta$} \Comment{two-stage zoom in the style of ReGround}
  \State $q_1\gets$ point from one call on $x$ (resend once on L1); if there is none, \Return no point
  \State $w\gets$ crop of one third of the width and height of $x$, centred on $q_1$, shifted inside $x$, at native pixels
  \State $q_2\gets$ point from one call on $w$ (resend once on L1), mapped to $x$;\ \ \Return $q_2$ if it exists, else $q_1$
\EndFunction
\end{algorithmic}
\end{algorithm*}

\begin{table*}[tp]
\centering\small\setlength{\tabcolsep}{4pt}
\caption{Shipped defaults for one model (Algorithm~\ref{alg:clicks} combines three). \emph{Transferred construction}: the best construction on the other five sources, applied to the evaluated source (Section~\ref{sec:res:transfer}). \emph{Final default}: what the pipeline ships. They coincide for set error on over-reporting families and differ elsewhere; the last column gives the reason. The prompts are in Appendix~\ref{app:protocol}, the settings of the GUI-grounding tier in Algorithm~\ref{alg:clicks}, and every value in the released configuration file.}\label{tab:defaults}
\begin{tabular}{@{}>{\raggedright\arraybackslash}p{2.7cm}>{\raggedright\arraybackslash}p{3.0cm}>{\raggedright\arraybackslash}p{3.6cm}>{\raggedright\arraybackslash}p{5.7cm}@{}}\toprule
Loss and family & Transferred construction & Final default & Evidence and reason \\\midrule
$E$, over-reporting families (detection, layout, screen parsing) & three-call unanimity, averaged coordinates (chosen in all six folds) & the same; list interface, lowest setting & Significantly better in 9 of 15 cells, significantly worse in none, not significant in 6. \\
$E$, under-reporting family (dense points, FSC-147) & three-call unanimity, averaged coordinates & single call, no deletion; list interface, lowest setting & The transferred construction is worse in 2 of 3 cells. Deleting unconfirmed records raises $E$ when their marginal matching rate $p_1$ is above 1/2 (Proposition~\ref{thm:intersect}); on FSC-147, $p_1$ is 0.39 to 0.90 and deletion hurts in 7 of 9 pairs (Section~\ref{sec:res:set}). \\
F1 & three-call union (five folds) or majority (one fold) & single call; on detection sources, interface and setting per model (GPT-6.1: regions, lowest; Sonnet 5.5: regions, default; Gemini 3.8: list, default); on screen sources, list interface, lowest setting & The transferred construction is better in 2 of 18 cells, worse in 4 and not significant in 12. A configuration chosen on three detection sources is better on the fourth in 5 of 12 cells and worse in 1 (Section~\ref{sec:res:transfer}). \\
Count & not part of the transfer & single call, list-first request; GPT-6.1 and Sonnet 5.5 lowest, Gemini 3.8 default setting; list count, with the stated count as declared fallback & Sections~\ref{sec:res:readout} and \ref{sec:res:degen}. \\
Click (hit in the target) & not part of the transfer & the user's model, two-stage zoom; GPT-6.1 and Sonnet 5.5 default, Gemini 3.8 lowest setting; Algorithm~\ref{alg:clicks} when the budget covers its planned worst case & On 600 ScreenSpot-Pro items the default setting lowers Sonnet 5.5's miss rate by 4.0 points and does not change GPT-6.1's significantly; Algorithm~\ref{alg:clicks} lowers the miss rate by 4.33 points against GPT-6.1's zoom (Section~\ref{sec:res:choose}). \\\bottomrule
\end{tabular}
\end{table*}

\paragraph{Inputs \src{task, config}.} The task fixes the class vocabulary $\mathcal{C}_\tau$, the alias map $\alpha_\tau$ (for example \{diamond, rhombus, rotated square\} $\mapsto$ square), the matching rule $\rho_\tau$ of Table~\ref{tab:contract}, the ignore region $\Omega_\tau$ and the loss $\ell_\tau$. The loss selects the construction in Step 5. The user also gives a budget per item. The shipped configuration fixes everything else that the algorithm calls a value: the prompt, the reporting interface and reasoning setting per model and loss (Table~\ref{tab:defaults}), the four flag thresholds, the recovery request, the view settings, the construction per loss, the alarm partner, the default alarm threshold, and the models, settings, distance, scorer and number of compared clusters of Algorithm~\ref{alg:clicks}. For counts, a model known to write degenerate lists at its lowest setting (Gemini 3.8) is run at its default setting (Section~\ref{sec:res:degen}).

\paragraph{Steps 1 and 2: dispatch and decode.} Step 1 is a dispatcher, not one fixed request: what is shared across tasks is the chain of records, losses and readouts, not one output format. For set tasks scored by F1 the shipped configuration may select the regions interface, in which the model returns one row per region of a fixed coordinate registry (Appendix~\ref{app:protocol}). The count request asks for one JSON object in which the list of records comes first and \texttt{counts\_by\_type} follows, declared ``for the shapes you listed'' (Appendix~\ref{app:protocol} gives the prompt). The allowed class names are listed in the format; the model scans row by row from top to bottom and uses its native coordinate frame (0 to 1000 for GPT and Gemini; pixels for Sonnet, with the image pre-resized so that the provider does not resize it again). No grid, ticks or markers are drawn. The decoder keeps every recoverable record in the audit collection $Z^{\mathrm{all}}$ and passes the valid ones, after the alias map, to the delivered collection $Z$ (Section~\ref{sec:setup}). It sets four degeneracy flags on the list: more than 10\% of records without a valid class; a list length outside $[0.5,1.5]$ times the stated total; more than 10\% repeated coordinates; and the finish reason ``length''.

\paragraph{Step 3: bounded recovery.} The two kinds of failure get different recoveries. A transport failure, in which the call did not return, is a property of the connection and is resent unchanged. A flagged response is a property of how the model wrote this report, and such errors tend to recur, consistent with Section~\ref{sec:theory:behaviour}: in our detection experiments, resending a request whose flag had already repeated cleared the flag in 14.7\% of cases, while changing the request cleared it in 58.8\%. The change was a follow-up that attached the previous answer and asked for the complete answer under the same contract, or, for reports organised by image region, a new request for the flagged regions only. The pipeline sends the first kind of follow-up, once. If the follow-up still carries a flag, the first response is kept with its flags, and Step 6 decides how to read it. Every attempt enters the cost. The replay at the end of this section reports how often recovery was triggered and what the stored follow-ups changed.

\paragraph{Step 4: views.} For clicks we use a re-grounding strategy in the style of the ReGround method that \citet{li2025screenspotpro} evaluate. The first call gives a point on the whole screenshot. A crop one third of the screenshot's width and height, centred on that point and shifted to stay inside the image, is sent at native pixels, and the second call gives the point in the crop. If the second call gives no point, the first point is kept; if the first call gives no point, there is nothing to zoom on and the answer is no point, which counts as a miss. Using agreement to decide when to zoom costs more and is not more accurate (Section~\ref{sec:res:view}). For sets and counts, the pipeline does not plan views itself: it works on whatever reports it is given. By default it sends the whole image. When a \emph{view planner} is configured, for example a tiling scheme or the companion paper's safe decomposition, which places every target wholly inside some view, with neighbouring views overlapping by one object and no view having fewer tokens per object than the whole image, and cannot lower expected recall under that paper's monotonicity assumptions \citep{shi2026zoomin}, each view is one call, its records are mapped to image coordinates, and Step 5 merges them. We computed the apparent size of every annotated object in the whole-image calls on SCP-dense, DocLayNet, DOTA and SKU-110K, using the input geometry measured in the companion paper and its released calibration data (32-pixel tokens for GPT-6.1, 28-pixel tokens for Sonnet 5.5, a fixed grid of about 1,090 tokens for Gemini 3.8). The median object spans 0.62 to 1.43 tokens across its shorter side (the companion paper measures size as the square root of the area, which is never smaller). Below that paper's design value of $S^\star=1.5$ fall 55\% to 79\% of objects on the three detection sources and every shape on SCP-dense; yet GPT-6.1 and Sonnet 5.5 write nearly exact lists on SCP-dense. The companion paper likewise finds that these model families reach half recall at well under one token per object side. A fixed threshold would decompose images whose whole-image report is already complete, so when to call a planner is a property of the model and the task that the configuration supplies. These sizes come from the annotations and inform only this design choice; the pipeline reads no annotation at run time. Section~\ref{sec:res:view} reports how the pipeline handles reports from four crops, and how our own tiling arm, which did not meet the safe condition, fared on large objects.

\paragraph{Step 5: construction.} The construction is chosen per loss (Table~\ref{tab:defaults}). Records of different calls are associated without labels (Table~\ref{tab:contract}): boxes of the same class with IoU $\ge0.5$, screen elements at IoU $\ge0.5$ without class, and points of the same class within the task's tolerance. On over-reporting task families and set error, three calls are made and a record is kept only if all three report it, with its coordinates averaged over the three. Everywhere else a single call is delivered without deletion. Over- or under-reporting is a property of the task family, shipped with the configuration from our evaluation; the user does not supply it. Coordinate averaging is on by default because its contribution is positive in 37 of 54 base/partner pairs and negative in 13 (Section~\ref{sec:res:set}); the deletion-only variant is in the menu. Reports of several views are first merged into one report: the union of their records, where a record is dropped if it is associated with a record already kept from another view (a duplicate in an overlap). When reports from different views confirm one another, a view can confirm or reject only records that lie wholly inside it; a record that no partner view contains is kept (Section~\ref{sec:theory:behaviour}). A call that still has no readable answer after recovery counts as an empty report, so three-call unanimity keeps nothing on that image; this is the construction we evaluated, and dropping such a call from the support count is in the menu.

\begin{algorithm*}[t]
\caption{GUI-grounding tier of FSR: clicks from several models. Called by Algorithm~\ref{alg:pipeline} only when the budget covers its planned worst case. Comments give the term of Proposition~\ref{prop:three} that each step acts on and the number of calls.}\label{alg:clicks}
\small
\begin{algorithmic}[1]
\Require screenshot $x$, instruction, budget $B$ \src{task}; models $M_1,\dots,M_J$ with their settings, scorer $M_s$, distance $\delta$, number of compared clusters $m_{\max}$, model order \src{config}
\Ensure point $\hat p$ or no point (a miss); status $\sigma$ (no candidate, one cluster, or the comparison outcome); alarm flag (1 if the candidates form two or more clusters, else 0); cost $c$ of every attempt
\Statex \emph{Shipped instance:} $J=3$ (GPT-6.1 and Sonnet 5.5 at the default setting, Gemini 3.8 at the lowest), $M_s$ = Gemini 3.8 at the lowest setting, $\delta=1.5\%$ of the longer side, $m_{\max}=3$; planned worst case $3J+2\binom{m_{\max}}{2}=15$ calls, technical retries excluded.
\For{$j=1,\dots,J$} \Comment{generate, acts on $F$; $2J$ calls}
  \State $p_j\gets\textsc{Zoom}(x,M_j)$ (Algorithm~\ref{alg:pipeline}); if $p_j$ exists, register it as a candidate whose identifier is bound to its coordinates on $x$, its model, setting and call
\EndFor
\If{there is no candidate} \Return no point \EndIf
\State $Q_1,\dots,Q_n\gets$ connected components of the graph that joins candidates within $\delta$;\ \ alarm $\gets(n\ge2)$ \Comment{detect}
\For{each candidate $j$} \Comment{represent, acts on $L$; $J$ calls}
  \State $s_j\gets$ hit score in $[0,1]$ from $M_s$ on a native crop with candidate $j$ marked; a missing or unreadable score counts as $-1$
\EndFor
\State representative of each $Q_i\gets$ its member with the highest $s_j$, ties broken by the model order
\If{$n=1$} \Return the representative of $Q_1$ \Comment{no alarm, no comparison}
\EndIf
\State rank the clusters by (representative's $s_j$, cluster size, earliest model in the cluster)
\For{each pair of the first $m_{\max}$ clusters, in both orders} \Comment{compare, acts on $W$; at most $2\binom{m_{\max}}{2}$ calls}
  \State $M_s$ sees both representatives marked by identifier, on the downscaled screenshot and on a native crop around each, and names the one that hits the target
\EndFor
\State a representative wins a pair only if both orders name it; a failed, undecided or split pair gives no win
\State \Return the representative of the cluster with the most wins, ties broken by the ranking; if no cluster has a win, the first cluster of the ranking
\end{algorithmic}
\end{algorithm*}

\paragraph{Step 6: readout.} The answer is read from the final set. For counts it is the list count of $S$, the number of records of each class. When the list behind $S$ carries a degeneracy flag and the response stated counts, the answer falls back to the stated count $d$, and the status says so (\texttt{count\_from\_statement}); a flagged list without stated counts keeps the list count and its flag. With several views, each response states counts for its own crop, and overlapping views would count shared objects twice; stated counts are therefore never combined across views, and a flagged merged list keeps the list count, with the status saying that no stated count was used. The fallback changes only the count: $S$ itself is not repaired, so the count and the set each carry their own status. The flags, not the readout gap, trigger the fallback, because a gap shows that the two readouts disagree without saying which is right (Proposition~\ref{prop:readout}).

\paragraph{A second model \src{config}, optional.} When a second model is configured, its default role is the alarm of Step 8: its answer is compared with the user's, and items where they disagree by more than a threshold are flagged. The disagreement $\mathrm{dis}_\tau$ is the summed per-class difference of the two answers for counts, the number of records not associated across the two answers for sets (Table~\ref{tab:contract}), and the distance between the two points as a share of the longer image side for clicks, where a missing point always raises the alarm. The partner is fixed by a rule (GPT-6.1 and Sonnet 5.5: Gemini 3.8; Gemini 3.8: Sonnet 5.5), which picks the cheapest other model per click call (Appendix~\ref{app:protocol}), and the threshold $\delta$ is a point the user picks on the risk-coverage curve (Section~\ref{sec:res:persist}; shipped default 1.5\%). Using the second model to confirm records is a menu option, chosen with labels or a shipped family-level ranking rather than as a zero-label default, because it is worse only in cells where the user's own model is the strongest (Section~\ref{sec:res:transfer}).

\paragraph{Step 7: optional labels \src{labels}.} Labels are never required. With fewer than three labelled images the pipeline keeps the shipped default and reports the labelled error. With three or more, it switches to another configuration from the menu only if that configuration's mean advantage over the labelled images, paired by image, minus one standard error is positive. Section~\ref{sec:res:set} evaluates this rule for the decision to delete unconfirmed records. With 5 to 10 labelled images it may also fit the record gate, a logistic selector on output features (Section~\ref{sec:res:select}). Certifying the accuracy of a pipeline on a population is a separate service that needs its own labelled sample.

\paragraph{Step 8: delivery.} The pipeline returns the final set $S$, the answer $\hat y$, the readout branch, the status $\sigma$ (L1 to L3, flags, recovery used, views used), the alarm score when a second report exists, and the cost of every attempt.

\paragraph{One entry point, replayed on stored calls.} The repository implements Steps 1 to 6 and 8 of Algorithm~\ref{alg:pipeline} for counts, sets and single-model clicks, including merging views and cross-view confirmation, as a single entry point (\texttt{fsr\_pipeline/}) whose branches read only the task specification, the shipped configuration and the responses a client returns; the label rule of Step 7 is a function in the same module, and the record gate is fitted by the analysis code of Section~\ref{sec:res:select}. Algorithm~\ref{alg:clicks} runs as a separate function (\texttt{gui/code/decision.py}), and the per-model F1 configuration is a key of the shipped configuration file evaluated by the analysis of Section~\ref{sec:res:transfer}; neither is a branch of this entry point. A replay client serves the stored responses of this paper instead of calling a model. A request for which nothing was stored, such as a follow-up that was never sent, returns ``unavailable'', and the pipeline continues on its documented branch. Run on raw responses, the entry point reproduces the component analyses: the count errors of Sections~\ref{sec:res:readout} and \ref{sec:res:degen}, the click hit rates after two-stage zoom of Section~\ref{sec:res:view}, and, with recovery switched off, the set error of the shipped construction in all 18 transfer cells to within 0.07 per image, a difference that comes from the 0.1-pixel coordinate rounding of the component analysis. Follow-ups were stored for a sample of flagged detection calls. In the 18 set cells, 171 image runs had at least one flagged call; a stored follow-up cleared the flag in 24 of them, and replaying these changes set error by at most 0.55 per image in any cell. No follow-ups were stored for the count experiments, where 51 runs were flagged, so their results are those of the first call. Synthetic boundary cases go through the same code, and each produces an answer, a readout branch and a status: an empty or unreadable response, truncated JSON, a list without stated counts, a degenerate list with and without stated counts, two transport failures in a row, one unreadable call among three, and a click with no first point or no point in the crop. The replay output is released with the analysis outputs (\texttt{pipeline\_replay.json}).

\subsection{GUI grounding: clicks from several models}\label{sec:method:clicks}

Algorithm~\ref{alg:pipeline} delivers one model's two-stage zoom point. When the per-item budget covers the planned worst case of a second tier, the pipeline offers one built on the three sources of a miss (Proposition~\ref{prop:three}); its mean cost is about 3.2 times that of one zoom. Its run-time inputs are the image, the instruction and the budget; it reads no target label and does not abstain. The models, their settings, the distance $\delta$, the scorer and the number of compared clusters come from the shipped configuration \src{config}.

Algorithm~\ref{alg:clicks} gives the steps, organised by Proposition~\ref{prop:three}. Generation enlarges the candidate menu, which is the only way to lower $F$; Section~\ref{sec:res:choose} measures the rescue rate of each source on the others' misses (Proposition~\ref{prop:coverage}). The representative step fixes which member of each cluster can be output, and the comparison chooses among representatives, so these two steps divide the remaining misses between $L$ and $W$. The comparison runs only on items with an alarm, where the measured miss rate of the scorer's choice is highest (Sections~\ref{sec:theory:points} and \ref{sec:res:choose}); items whose candidates form one cluster keep that cluster's representative. The scorer's hit score is used only to rank. The registry makes the point that is scored, drawn, compared and returned the same point: every step refers to a candidate by its identifier, and the code asserts that the coordinates drawn for the scorer equal those returned. If a call fails, the procedure continues with the remaining candidates. It returns no point only when no candidate is available; a missing point counts as a miss, and every returned point is scored by the hit-in-target rule. The budget selects among planned call schedules before inference: the tier runs only if the budget covers its planned worst case of 15 calls at per-call output caps of 16,384 tokens at the lowest setting and 32,768 at the default. Technical retries are outside this bound, and the per-item costs we report are realized averages over all attempts, retries included. The rules are implemented as one deterministic function over stored or live responses (\texttt{gui/code/decision.py} in the release), with request builders for each step; with the evaluation script, run on the stored responses, it regenerates the results of Section~\ref{sec:res:choose}.

\subsection{Revising the model from failed predictions}\label{sec:method:revise}

The theory was built by iteration, and the iteration is part of the method. A model is stated, a consequence is derived and tested, and when a prediction fails we record three things: which assumption failed and how we know; how the revised assumption was checked directly; and what new consequence the revision predicts, and whether it holds. Table~\ref{tab:revisions} lists the five revisions that shaped the final theory. Each row names the kind of evidence: an exact identity, a controlled comparison of interventions on the same images, or an association observed across cells or records.

\begin{table*}[tp]
\centering\small\setlength{\tabcolsep}{4pt}
\caption{Revisions of the model. Each row records the failed assumption and the evidence against it, the revised assumption and its direct check, and the new consequence with its test. The kind of evidence is named in each cell.}\label{tab:revisions}
\begin{tabular}{@{}>{\raggedright\arraybackslash}p{3.1cm}>{\raggedright\arraybackslash}p{3.4cm}>{\raggedright\arraybackslash}p{4.0cm}>{\raggedright\arraybackslash}p{4.5cm}@{}}\toprule
Failed assumption & How we know & Revised assumption and direct check & New consequence and test \\\midrule
The intersection threshold, applied to intersections with averaged coordinates, predicts whether confirmation helps. & Wrong sign in 2 of 54 base/partner pairs. & The threshold governs deletion only; coordinate averaging is a separate term (Proposition~\ref{thm:intersect}, Proposition~\ref{prop:geometry}). Exact identity: both terms add up to the observed change on every image of all 54 pairs. & The averaging term can take either sign and can reverse the result. Observed: positive in 37 pairs, negative in 13; it reverses the deletion-only result in two pairs, which accounts for both former exceptions (Section~\ref{sec:res:set}). The deletion term, in turn, can be predicted from $p_1$ on 5 labelled images in 83\% of draws (Section~\ref{sec:res:set}). \\
Similar quality is what makes a second model a useful confirming partner. & On ScreenParse the three models have similar set error (16.2, 13.3, 16.0), yet among constructions with averaged coordinates the best cross-model construction is not significantly different from the best same-model one ($+1.8\%$, $[-3.0,7.6]$): similar aggregate quality does not establish a benefit. & A partner helps if it rejects more of the base model's false records than it loses of its true ones (Proposition~\ref{prop:partner}). Direct check across pairs: cross-model partners separate the two better than same-model partners in 32 of 36 comparisons; the 4 exceptions are partners that confirm few of the base model's true records. & Conditional consequence: a base with many true and few false records gains only if the partner keeps almost all of its true records ($n_+(1-h_+^B)$ dominates $G$), and a weaker partner usually keeps fewer, so cross-model confirmation is expected to do worst where the base model is strongest. Observed association across transfer cells: all 5 significantly worse cells are cells where the base model is strongest, and 20 of the 24 cells where it is not strongest improve (Section~\ref{sec:res:transfer}). \\
A count written first, or given in the prompt, pulls the list length like a quota for every model. & Giving a count moves list length with slope at most 0.004 for GPT-6.1 and Sonnet 5.5 on SCP and SCP-dense (controlled comparison). & The pull depends on the model and on how reliable its enumeration is. Direct check: slope 0.17 to 0.20 on PixMo, and 1.63 for Gemini 3.8 on SCP-dense, where its lists degenerate. & Asking for the list before the counts should not hurt and should help where enumeration is reliable. Observed in a controlled comparison under the same mapped readout: list-first lists beat count-first lists for all three models ($-39\%$, $-92\%$, $-25\%$; Section~\ref{sec:res:readout}). The list-first prompt also adds the row-by-row scan and the allowed class names, so this is a test of the protocol, not of order alone. \\
Gemini 3.8 has a coordinate spread that a different coordinate convention reduces. & Pixel versus grid convention: $+2\%$, n.s.\ (controlled comparison). & Current models make no coordinate-frame error on SCP. Direct check: systematic offset at most 0.3 pixels, slope 1.000, spread 0.9 to 1.5 pixels, against a 30-pixel tolerance. & References that correct coordinates (grids, ticks) should not improve localization. Observed in controlled comparisons: localization changes by at most 0.23 pixels in every arm. Any effect of an overlay must come through how the model writes its list (next row). \\
A grid helps Gemini 3.8 by improving localization. & The grid lowers set error by 3.5 per image at the lowest setting, but localization does not change, and at the default setting the grid makes detection slightly worse (controlled comparisons). & The grid acts on the record format: its gain falls on records that omit their class field. Association within lists: a record omits its class more often when the previous record has the same class (logistic regression controlling for position in the list: $+0.70$ $[0.57,0.84]$ in log-odds, arm fixed effects, intervals clustered by image). This path explains only a small part of the grid's gain. & An order that interleaves classes should lower omissions, and grouping by class should raise them. Observed in a controlled comparison: a row-by-row scan lowers set error by 11.5 per image against 3.5 for the grid, and grouping by class raises it (Section~\ref{sec:res:degen}). \\\bottomrule
\end{tabular}
\end{table*}

\section{Experimental Setup}\label{sec:experiments}

\paragraph{Models and settings.} The main experiments use three API models: GPT-6.1 (OpenAI), Claude Sonnet 5.5 (Anthropic) and Gemini 3.8 Flash (Google), written Sonnet 5.5 and Gemini 3.8. Pilots and the controlled-scene experiments also use GPT-5.6 and Opus 5.5, and an open-weight model, Qwen3-VL-235B, is tested in the pilots and on the readout tasks. Each call uses either the \emph{lowest} reasoning setting the API offers or the \emph{default} setting. Setting labels do not match behaviour, and we report what was measured (Appendix~\ref{app:settings}). Three facts matter for interpretation. First, the default setting also doubled the output cap (16,384 to 32,768 tokens); three runs separate the two: the Gemini 3.8 runs of Section~\ref{sec:res:degen}, a 90-call 16k control on the detection sources, and a 16k control on controlled scenes. Second, Sonnet 5.5's reasoning share at its default setting varies by task: 0\% to 23\% of calls in the reasoning experiment (dense scenes, FSC-147, screens) and 0\% to 90\% across detection cells; we attribute a Sonnet result to reasoning only where most of its calls reasoned. Reasoning tokens are read from the usage fields of every call after the pilots (Qwen3-VL is an instruct model and reports none); the two pilots logged no thinking tokens for the Claude models. Third, Gemini 3.8 at its lowest setting still reasoned on click and point outputs (about 99\% of GUI click calls), so its click results at the lowest setting include reasoning. GPT-6.1's lowest setting is \texttt{reasoning\_effort=low}, and most of its calls carry some reasoning tokens. Images sent to Sonnet 5.5 were resized to at most 1,568 pixels on the long side and 1.15 megapixels (by our code, or in the detection experiment by the provider), and images sent to the other models to at most 2,048 pixels; for Sonnet 5.5 this is below the 2,576-pixel long edge measured for it in the companion paper \citep{shi2026zoomin}.

\paragraph{Data.} The \emph{synthetic counting probe} (SCP) has 100 images of $1024\times1024$ pixels with five shape classes and 15 to 80 shapes each (mean 47.6), drawn over clutter lines; \emph{SCP-dense} uses the same generator with 50 images of 121 to 200 smaller shapes (mean 164.3). Labels are exact. Photographs, aerial images and document pages come from PixMo-Count (100 images, 2 to 10 objects), FSC-147 (100 images in three size strata for the readout and tiling experiments, and 60 further images for the detection experiment), and 60 images each from SKU-110K, DOTA v1.0 and DocLayNet v1.2 (three classes). Screen data are 200 ScreenSpot-Pro and 60 ScreenSpot-v2 click items, 60 GroundCUA screens and 60 ScreenParse screens; the click experiment of Section~\ref{sec:res:choose} adds 600 further ScreenSpot-Pro click items, sampled with a fixed seed and sharing no item and no screenshot with the 260 click items. The controlled-scene experiments add 96 procedurally drawn scenes with 16 to 384 targets each. A readout experiment on aerial and document images adds 30 DOTA v1.0 windows and 50 DocLayNet pages (financial reports, tenders and laws), selected with the annotations to contain at least three (DOTA) or four (DocLayNet) classes and at least 30 or 20 objects, disjoint from the detection images. In total the experiments use 1,746 images with 54,475 annotated targets (each click item counts as one target). We refer to SKU-110K, DOTA, DocLayNet and FSC-147 as the \emph{detection sources} and to GroundCUA and ScreenParse as the \emph{screen sources}.

\paragraph{Experiments.} Table~\ref{tab:experiments} lists the experiments that the results draw on. Every call of these experiments, including smoke tests, pilots and retries, is stored with its raw response; the full registry with costs is in Appendix~\ref{app:registry}. Merges of existing calls (intersections, unanimity, cross-model constructions) are deterministic post-processing of stored responses: the same items are re-scored under a changed construction.

\begin{table}[t]
\centering\small\setlength{\tabcolsep}{3pt}
\caption{All experiments with model calls, grouped (67,364 calls, including repairs and smoke tests); the registry in Appendix~\ref{app:registry} itemizes them.}\label{tab:experiments}
\begin{tabular}{@{}>{\raggedright\arraybackslash}p{1.75cm}>{\raggedright\arraybackslash}p{2.85cm}r@{}}\toprule
Experiment & What changes & Calls \\\midrule
Readout & 13 prompt arms on SCP, SCP-dense, PixMo & 7,900 \\
Readout, full protocol & SCP-dense; three models at the lowest setting, Gemini 3.8 also at the default (16k and 32k caps) & 250 \\
Multi-class readout & count-first vs.\ list-first, DOTA and DocLayNet & 574 \\
Grid & grids, ticks, cells, scan, type names & 1,340 \\
Tiling & count-first, list-only, $2\times2$ tiles (FSC-147) & 1,800 \\
Detection & list / regions $\times$ two settings, 4 datasets & 6,523 \\
GUI & click arms incl.\ zoom; parse arms incl.\ intersections; pilots & 7,232 \\
Zoom on second click & zoom around each model's second click & 772 \\
Click scores and choice & scores, default zooms, comparisons, crop-first baselines (260 items) & 5,964 \\
Clicks, 600 items & all arms of Section~\ref{sec:res:choose} & 13,126 \\
Fourth click source & crop-first point added to the tier, same items & 717 \\
Review & blind vs.\ draft review on crops & 1,440 \\
Same-view review & draft review on detection images & 663 \\
Comparator & pairwise comparison of disagreeing reports & 1,960 \\
Resolution & downscaled / native / upscaled & 600 \\
End-to-end & naive / common / proposed arms & 1,800 \\
Reasoning & default setting on dense, FSC, GUI & 1,110 \\
Open model & Qwen3-VL readout arms & 650 \\
Controlled scenes & pilots, subset execution, drift, grids & 12,296 \\
Pilots & small runs before the readout experiments & 647 \\\bottomrule
\end{tabular}
\end{table}

\paragraph{Statistics.} Intervals are 95\% paired bootstrap intervals over images (4,000 resamples unless stated), with relative changes computed on means. ``Significant'' means the interval excludes 0. Comparisons of two click flows on the same items also report the items repaired and broken and an exact McNemar test, with Holm adjustment where two comparisons share a claim. Fold assignments and random draws use seeds derived from SHA-256 of a fixed string and are persisted, so every table regenerates byte-identically from the saved responses with one script. Costs count every attempted call at list prices, with one exception: when the 600-item click run was interrupted and resumed, at most 8 requests were in flight without a stored response, at most about \$0.05, which is not in the totals.

\section{Results}\label{sec:results}

The results follow the three problems of Section~\ref{sec:intro}: a comparison with common practice (Section~\ref{sec:res:common}), repairing reports (Sections~\ref{sec:res:readout} to \ref{sec:res:view}), detecting errors (Section~\ref{sec:res:persist}), repairing a click by choosing among several models (Section~\ref{sec:res:choose}), and judging which report or configuration is more accurate (Sections~\ref{sec:res:select} and \ref{sec:res:transfer}). Each table caption states the operation, the comparator, the loss and any label use.

\subsection{Against common practice}\label{sec:res:common}

\begin{table*}[tp]\centering\footnotesize\setlength{\tabcolsep}{3pt}
\caption{Against common practice, from stored calls only. Columns give one call, the usual remedies (self-consistency: median of three stated counts, or a vote of three clicks; uncertainty-gated zoom: zoom only when two clicks disagree; majority vote and equal-weight box fusion over three calls), and our default. Relative changes are against one call, lower is better. \textbf{Bold}: the paired 95\% interval excludes 0 in the favourable direction; \underline{underlined}: it excludes 0 in the unfavourable direction. Our count readout costs about as much as one call and our click protocol less than a three-call vote. For Gemini 3.8, ``ours'' is the list-first protocol at its lowest setting; the method runs Gemini 3.8 at its default setting, which at the same 16k cap gives a readout error of 24.52 (Section~\ref{sec:res:degen}). For F1 our default is one call; over the 15 over-reporting cells, majority vote and box fusion change F1 by a median of $-0.3\%$ and $+0.2\%$. Per-cell set errors are in the released analysis outputs.}\label{tab:common}
\begin{tabular}{@{}>{\raggedright\arraybackslash}p{1.8cm}>{\raggedright\arraybackslash}p{2.6cm}>{\raggedright\arraybackslash}p{3.5cm}>{\raggedright\arraybackslash}p{3.6cm}>{\raggedright\arraybackslash}p{3.65cm}@{}}\toprule
& One call & Common remedy & Second common remedy & Ours (Section~\ref{sec:method}) \\\midrule
\multicolumn{5}{@{}p{15.2cm}@{}}{\emph{Repair the readout: dense counting (SCP-dense, 50 images, per-type count error; cost per image)}} \\
\quad GPT-6.1 & 10.84 (\$0.029) & 8.32 (\$0.089), {\bfseries\boldmath $-$23.2\%} & n/a & 1.60 (\$0.025), {\bfseries\boldmath $-$85.2\%} \\
\quad Sonnet 5.5 & 35.02 (\$0.031) & 34.56 (\$0.094), $-$1.3\% & n/a & 0.42 (\$0.033), {\bfseries\boldmath $-$98.8\%} \\
\quad Gemini 3.8 & 45.68 (\$0.026) & 45.22 (\$0.076), $-$1.0\% & n/a & 52.80 (\$0.019), $+$15.6\% \\
\midrule
\multicolumn{5}{@{}p{15.2cm}@{}}{\emph{Repair the view: clicks (ScreenSpot-Pro items with all four calls stored: 199, 196 and 200 items for GPT-6.1, Sonnet 5.5 and Gemini 3.8; miss rate; cost per item)}} \\
\quad GPT-6.1 & 11.1\% (\$0.0061) & 10.5\% (\$0.0182), $-$4.5\% & 11.1\% (\$0.0123), 0.0\% & 9.6\% (\$0.0083), $-$13.6\% \\
\quad Sonnet 5.5 & 20.4\% (\$0.0035) & 20.9\% (\$0.0104), $+$2.5\% & 19.4\% (\$0.0071), $-$5.0\% & 11.2\% (\$0.0058), {\bfseries\boldmath $-$45.0\%} \\
\quad Gemini 3.8 & 35.5\% (\$0.0030) & 35.0\% (\$0.0086), $-$1.4\% & 31.0\% (\$0.0063), {\bfseries\boldmath $-$12.7\%} & 12.5\% (\$0.0053), {\bfseries\boldmath $-$64.8\%} \\
\midrule
\multicolumn{5}{@{}p{15.2cm}@{}}{\emph{Repair the set: detection and screen parsing (15 over-reporting cells, set error $E$, 3 calls each; cells better / worse than one call, median change)}} \\
\quad 15 cells & one call & majority vote: 9 / 0, $-$2.9\% & box fusion: 8 / 0, $-$3.7\% & 9 / 0, $-$9.6\%; against box fusion: 11 / 0 \\
\midrule
\multicolumn{5}{@{}p{15.2cm}@{}}{\emph{Detect errors: clicks, share of a single call's misses caught by a disagreement alarm ($\delta=1.5\%$; Section~\ref{sec:res:persist})}} \\
\quad GPT-6.1 & n/a & same model: 17\% & n/a & second model: 61\% \\
\quad Sonnet 5.5 & n/a & same model: 36\% & n/a & second model: 71\% \\
\quad Gemini 3.8 & n/a & same model: 27\% & n/a & second model: 52\% \\
\bottomrule\end{tabular}\end{table*}

The comparisons in this subsection use the stored calls of the earlier experiments; the 600-item click evaluation is in Section~\ref{sec:res:choose}.

Table~\ref{tab:common} compares the defaults of Section~\ref{sec:method} with the remedies practitioners use most, on the same stored calls. The pattern follows the three problems. Self-consistency repeats the same source of evidence, so its gains are small: the median of three stated counts helps GPT-6.1 by 23\% and the other two models by about 1\%, at three times the cost, and a vote of three clicks changes the miss rate by $-4.5\%$ to $+2.5\%$. Uncertainty-gated zoom, which zooms only when two clicks of the same model disagree, inherits the same weakness, because persistent errors rarely produce disagreement; it gains 13\% for Gemini 3.8 and nothing significant for the other two. The remedies that change the information win: reading the list cuts dense-count error by 85\% and 99\% for about the cost of one call, and zooming every item cuts the miss rate by 45\% and 65\% for less than a three-click vote. For set error, three-call unanimity with averaged coordinates improves 9 of the 15 over-reporting cells (median $-9.6\%$), against $-2.9\%$ for majority vote and $-3.7\%$ for box fusion, and it beats box fusion directly in 11 of the 15 cells and is not significantly worse in any. For detecting missed clicks at the same distance threshold, a second model catches 1.9 to 3.5 times the share of errors that a second call of the same model catches, while flagging more items (Section~\ref{sec:res:persist}).

\subsection{Repair the readout: read the list, not the number}\label{sec:res:readout}

\begin{table*}[tp]\centering\footnotesize\setlength{\tabcolsep}{3pt}
\caption{Readout effects kept separate (SCP-dense, 50 images, lowest setting, one call; per-type count error, lower is better; paired image bootstrap 95\% intervals). The \emph{raw} list count counts the records of each class before the alias map, the \emph{mapped} list count after it. \emph{Pure readout}: stated count $\to$ raw list count of the same response. \emph{Alias map}: raw $\to$ mapped list count of the same list. \emph{Protocol}: mapped list count of a count-first response $\to$ that of a list-first response. \emph{Full method}: stated count of a count-first response $\to$ readout of Section~\ref{sec:method} on a list-first response, the effect of the reporting and decoding protocol. \emph{Totals}: mean absolute difference between the list length and the true total, and between the stated total and the true total. \textbf{Bold}: the paired 95\% interval excludes 0 in the favourable direction; \underline{underlined}: it excludes 0 in the unfavourable direction.}\label{tab:readout}
\begin{tabular}{@{}>{\raggedright\arraybackslash}p{2.2cm}>{\raggedright\arraybackslash}p{2.4cm}ccc@{}}\toprule
Effect & Comparison & GPT-6.1 & Sonnet 5.5 & Gemini 3.8 (lowest) \\\midrule
Pure readout, count-first & stated $\to$ raw list count & \begin{tabular}{@{}c@{}}10.84 $\to$ 2.64\\ {\bfseries\boldmath $-$75.6\% [$-$83.2, $-$65.8]}\end{tabular} & \begin{tabular}{@{}c@{}}35.02 $\to$ 7.80\\ {\bfseries\boldmath $-$77.7\% [$-$91.3, $-$58.4]}\end{tabular} & \begin{tabular}{@{}c@{}}45.68 $\to$ 129.04\\ \underline{$+$182.5\% [163.3, 202.3]}\end{tabular} \\
Pure readout, list-first & stated $\to$ raw list count & \begin{tabular}{@{}c@{}}10.10 $\to$ 1.60\\ {\bfseries\boldmath $-$84.2\% [$-$88.2, $-$79.6]}\end{tabular} & \begin{tabular}{@{}c@{}}18.52 $\to$ 5.42\\ {\bfseries\boldmath $-$70.7\% [$-$76.9, $-$64.8]}\end{tabular} & \begin{tabular}{@{}c@{}}46.56 $\to$ 95.04 ($n=45$)\\ \underline{$+$104.2\% [72.7, 139.5]}\end{tabular} \\
Alias map, count-first & raw $\to$ mapped list count & 2.64 $\to$ 2.64 (no change) & \begin{tabular}{@{}c@{}}7.80 $\to$ 5.42\\ {\bfseries\boldmath $-$30.5\% [$-$76.3, $-$12.7]}\end{tabular} & 129.04 $\to$ 129.04 (no change) \\
Alias map, list-first & raw $\to$ mapped list count & 1.60 $\to$ 1.60 (no change) & \begin{tabular}{@{}c@{}}5.42 $\to$ 0.42\\ {\bfseries\boldmath $-$92.3\% [$-$97.0, $-$86.2]}\end{tabular} & 96.78 $\to$ 96.78 (no change) \\
Protocol & mapped list count, count-first $\to$ list-first & \begin{tabular}{@{}c@{}}2.64 $\to$ 1.60\\ {\bfseries\boldmath $-$39.4\% [$-$57.6, $-$13.5]}\end{tabular} & \begin{tabular}{@{}c@{}}5.42 $\to$ 0.42\\ {\bfseries\boldmath $-$92.3\% [$-$97.7, $-$44.4]}\end{tabular} & \begin{tabular}{@{}c@{}}129.04 $\to$ 96.78\\ {\bfseries\boldmath $-$25.0\% [$-$32.7, $-$17.6]}\end{tabular} \\
Full method & count-first stated $\to$ readout & \begin{tabular}{@{}c@{}}10.84 $\to$ 1.60\\ {\bfseries\boldmath $-$85.2\% [$-$89.1, $-$80.3]}\end{tabular} & \begin{tabular}{@{}c@{}}35.02 $\to$ 0.42\\ {\bfseries\boldmath $-$98.8\% [$-$99.5, $-$98.1]}\end{tabular} & \begin{tabular}{@{}c@{}}45.68 $\to$ 52.80\\ $+$15.6\% [$-$5.3, 39.4]\end{tabular} \\
\midrule
Totals, list-first & list length vs.\ stated total & 0.04 vs.\ 7.78 & 0.02 vs.\ 10.48 & 109.38 vs.\ 46.11 \\
\bottomrule\end{tabular}\end{table*}

\begin{figure*}[tp]
\centering
\includegraphics[width=\textwidth,alt={A dense synthetic scene with 173 shapes, a zoomed detail with the records of GPT-6.1 drawn as coloured dots that each match a different shape, and the abridged text response, whose list has 173 records while its stated counts sum to 188.}]{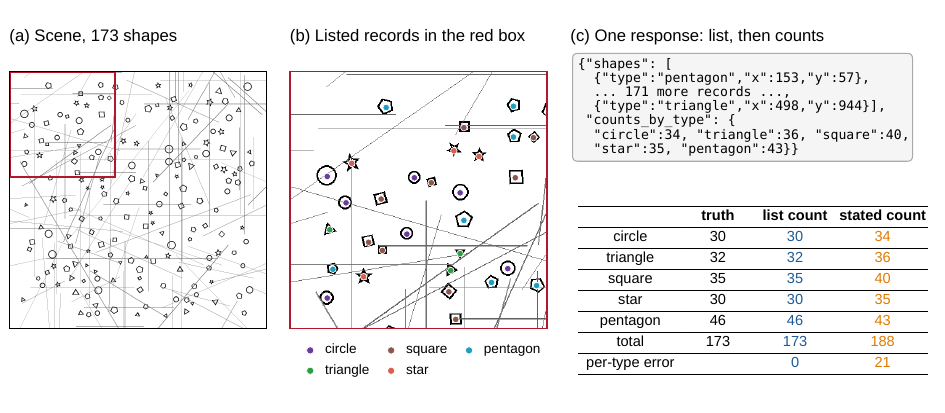}
\caption{One response, two answers. GPT-6.1 is asked to list every shape in a dense synthetic scene (SCP-dense image 28) and then give counts by type ``for the shapes you listed''. (a) The scene has 173 shapes. (b) A detail of the scene with the records of the model's list as dots, one colour per class: each record matches a different shape of the same class within 20 pixels. (c) The response itself, abridged; coordinates are on the model's native 0 to 1000 grid. The list has 173 records, so the list count of every class is exact, while the counts the model states after it sum to 188, a per-type error of 21. Across 50 such scenes, list length misses the true total by 0.04 objects on average and the stated total misses by 7.78 (Table~\ref{tab:readout}).}
\label{fig:example}
\end{figure*}

\paragraph{Same response, two answers.} Figure~\ref{fig:example} shows one response in full. When a list-first response has both a list and stated counts, the list length misses the true total by 0.04 objects on average for GPT-6.1 and 0.02 for Sonnet 5.5, while the stated totals miss by 7.78 and 10.48 (Table~\ref{tab:readout}, last row). Reading per-type counts from the list (the list count) instead of the stated counts in the same response cuts per-type count error by 75.6\% (GPT-6.1) and 77.7\% (Sonnet 5.5) under a count-first prompt, and by 84.2\% and 70.7\% under the list-first prompt. Generation is held fixed in these comparisons, so the gain is a readout effect: the model's list carries information that its own summary loses.

\begin{figure*}[tp]
\centering
\includegraphics[width=\textwidth,alt={Three panels, one per model (GPT-6.1, Sonnet 5.5, Gemini 3.8), each with four bars giving the relative change of count error from the stated count to the list count in the same response, with 95 percent intervals, for PixMo-Count, FSC-147, SCP and SCP-dense ordered by objects per image. For GPT-6.1 the change is minus 6.3, minus 1.1 (both not significant), minus 44.1 and minus 75.6 percent; for Sonnet 5.5 minus 17.6 and minus 4.2 (both not significant), minus 98.5 and minus 77.7 percent; for Gemini 3.8 minus 18.2 (not significant), minus 42.8, plus 191.4 and plus 182.5 percent.}]{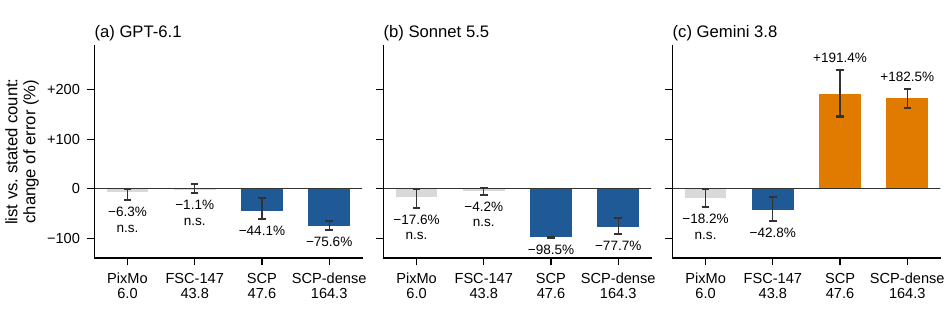}
\caption{Same-response readout across datasets (count-first prompts, lowest reasoning setting, one call; 100 images per dataset, 50 on SCP-dense). Bars: relative change of count error from the stated count to the raw list count of the same response, with paired image-bootstrap 95\% intervals; negative means the list count is better. Blue: interval below 0; orange: interval above 0; grey and n.s.: interval includes 0. Error is per type on SCP and SCP-dense and total on PixMo-Count and FSC-147; numbers under dataset names are mean objects per image. Absolute errors are in Table~\ref{tab:density}. The list count is better on the multi-class synthetic scenes for GPT-6.1 and Sonnet 5.5; Gemini 3.8's lists degenerate there, so reading them is worse (Section~\ref{sec:res:degen}).}
\label{fig:density}
\end{figure*}

\paragraph{The gap grows with the length of a multi-type list.} Figure~\ref{fig:density} (values in Table~\ref{tab:density}) repeats the same-response comparison on four datasets. On PixMo (6 objects per image, one class) and FSC-147 (43.8 objects, one class), the stated total tracks the list for GPT-6.1 and Sonnet 5.5 and the difference is not significant. On SCP (47.6 objects, five classes) the list count is better by 44\% for GPT-6.1 and by 98.5\% for Sonnet 5.5, whose stated per-type counts are 65 times as far from the truth as its list. On SCP-dense (164.3 objects) both models gain 76\% to 78\%. The gap therefore appears when the stated count must summarize a long list across several classes.

\paragraph{Aliases and protocol.} In list-first responses, Sonnet 5.5 wrote 258 ``diamond'' records across 50 images although the format listed the allowed names. The declared alias map, applied to the same list, takes its error from 5.42 to 0.42 ($-92.3\%$). The prompt alone does not do this: the map is part of the task, and the decoder applies it. Holding the mapped readout fixed, list-first lists are better than count-first lists by 39.4\% for GPT-6.1 and 92.3\% for Sonnet 5.5. The complete protocol, compared with the stated count of a count-first response, cuts error by 85.2\% $[-89.1,-80.3]$ for GPT-6.1 and 98.8\% $[-99.5,-98.1]$ for Sonnet 5.5. For Gemini 3.8 at its lowest setting the change is $+15.6\%$ $[-5.3,39.4]$, not significant: the degeneracy check routes 40 of its 50 responses to the stated count. These are effects of the reporting and decoding protocol. The same-response and same-list comparisons isolate the readout and the alias map; the full protocol also changes how the list is generated.

\paragraph{Aerial and document images with several classes.} The readout repair needs a gap to work with. We ran both prompts on 30 DOTA windows and 50 DocLayNet pages chosen to contain several classes and many objects (Table~\ref{tab:natural}; 574 calls). There the lists themselves miss objects: GPT-6.1 and Sonnet 5.5 list 68\% to 88\% of the true objects. The stated count also stays close to the list count of the same response, differing by 0.1 to 7.4 per image against per-type errors of 5 to 46. Switching readouts can change the error by at most this gap (Proposition~\ref{prop:readout}), so the change in the same response is small: none for GPT-6.1, and $-13.5\%$ and $-19.9\%$ for Sonnet 5.5. The full list-first protocol, here without the recovery step, changes count error by $+0.4\%$ and $-2.4\%$ for GPT-6.1 and by $-15.7\%$ and $-20.1\%$ for Sonnet 5.5, intervals reaching zero. Gemini 3.8 at its lowest setting again writes degenerate lists; the fallback to the stated count holds its change to $+18.0\%$ and $-7.9\%$, both not significant, and at its default setting, the shipped configuration, it changes count error, together with the setting, by $-28.2\%$ on DOTA (interval reaching zero) and by $-25.8\%$ on DocLayNet. The two kinds of scene differ in where the error sits. On dense synthetic scenes the lists are nearly complete and the statement loses information, so the readout repairs most of the error; on these aerial and document images most of the error is already in the list, so switching between the two readouts cannot remove it. The label-free readout gap tells the two cases apart in one direction: when it is small, switching readouts cannot matter (Proposition~\ref{prop:readout}).

\begin{table*}[tp]\centering\footnotesize\setlength{\tabcolsep}{3pt}
\caption{Readout on aerial and document images with several classes: 30 DOTA windows (71 objects and 3.3 classes per image on average) and 50 DocLayNet pages (25 elements, 5.0 classes); one call per image and arm, lowest setting unless stated, 16k cap. Per-type count error, lower is better; paired image bootstrap 95\% intervals. Difficult DOTA objects and objects less than half inside the window are not targets, so records on them are counted like any other record, for every readout alike. \emph{List recall}: share of true objects matched by a record centre inside a true box of the same class (list-first arm). \emph{Readout gap}: mean per-type difference between the stated count and the list count of the same count-first response, which bounds how much switching readouts can change the error. \emph{Same response}: stated count $\to$ list count of the count-first response. \emph{Full method}: readout of the list-first response (list count, or the stated count if the list is flagged) against the stated count of the count-first response; the Gemini 3.8 default row is the shipped configuration. \textbf{Bold}: the paired 95\% interval excludes 0 in the favourable direction; \underline{underlined}: it excludes 0 in the unfavourable direction.}\label{tab:natural}
\begin{tabular}{@{}llrrrrrr@{}}\toprule
& & & & \multicolumn{2}{c}{Same response} & \multicolumn{2}{c}{Full method} \\\cmidrule(lr){5-6}\cmidrule(l){7-8}
Images & Model & \begin{tabular}[b]{@{}c@{}}List\\recall (\%)\end{tabular} & \begin{tabular}[b]{@{}c@{}}Readout\\gap\end{tabular} & Error & Change, \% [95\% CI] & Error & Change, \% [95\% CI] \\\midrule
DOTA & GPT-6.1 & 88 & 2.2 & 40.5 $\to$ 40.8 & $+$0.7 [$-$5.0, 4.5] & 40.7 & $+$0.4 [$-$17.7, 16.4] \\
 & Sonnet 5.5 & 68 & 7.4 & 45.8 $\to$ 39.7 & {\bfseries\boldmath $-$13.5 [$-$34.7, $-$0.6]} & 38.6 & $-$15.7 [$-$38.2, 7.7] \\
 & Gemini 3.8 & 26 & 44.2 & 33.2 $\to$ 63.9 & \underline{$+$92.4 [54.2, 145.0]} & 39.2 & $+$18.0 [$-$13.4, 61.2] \\
 & Gemini 3.8, default & n/a & n/a & n/a & n/a & 23.9 & $-$28.2 [$-$48.6, 0.1] \\
\midrule
DocLayNet & GPT-6.1 & 87 & 0.1 & 4.9 $\to$ 4.9 & $+$0.4 [$-$1.6, 2.2] & 4.8 & $-$2.4 [$-$12.4, 4.5] \\
 & Sonnet 5.5 & 83 & 3.1 & 8.6 $\to$ 6.9 & {\bfseries\boldmath $-$19.9 [$-$32.5, $-$10.1]} & 6.8 & $-$20.1 [$-$36.7, 3.3] \\
 & Gemini 3.8 & 68 & 5.3 & 9.4 $\to$ 11.2 & \underline{$+$19.2 [1.6, 41.0]} & 8.6 & $-$7.9 [$-$24.7, 10.6] \\
 & Gemini 3.8, default & n/a & n/a & n/a & n/a & 7.0 & {\bfseries\boldmath $-$25.8 [$-$42.5, $-$8.8]} \\
\bottomrule\end{tabular}\end{table*}

\subsection{Repair the readout when the list degenerates}\label{sec:res:degen}

\begin{table*}[tp]\centering\footnotesize\setlength{\tabcolsep}{3pt}
\caption{Gemini 3.8 on SCP-dense under the list-first protocol (50 images, one call each). At a fixed 16k output cap, the default reasoning setting changes the count readout error by $-$53.6\% [$-$72.4, $-$29.5]; raising the cap to 32k changes it by a further $-$37.0\% [$-$62.4, 7.7] (not significant). Readout error: per-type count error of the Step-6 readout (stated count if the list is flagged, list count otherwise). List-count error: the same for the list count alone. Set error $E$ scores the delivered records, those whose class is valid after the alias map (equal class, 20 px); unknown-class records are excluded from the delivered set and counted in the next column; the column after it is a sensitivity that keeps them and scores them as unmatched. Flagged: share of lists with at least one degeneracy flag. JSON fail: share of responses that do not parse as the requested object (records are then recovered by the declared salvage step). Hit cap: share of responses that stopped at the output cap. Reasoning: share of calls with reasoning tokens. Output tokens: mean per call. Cost: all 50 calls, counting every attempted call. The setting comparison is cap-matched, not cost-matched: output tokens and cost rise with reasoning.}\label{tab:setting}
\begin{tabular}{@{}lrrrrrrrrrrr@{}}\toprule
\begin{tabular}[b]{@{}c@{}}Setting,\\cap\end{tabular} & \begin{tabular}[b]{@{}c@{}}Readout\\error\end{tabular} & \begin{tabular}[b]{@{}c@{}}List-count\\error\end{tabular} & \begin{tabular}[b]{@{}c@{}}Set\\error $E$\end{tabular} & \begin{tabular}[b]{@{}c@{}}Unknown\\records\end{tabular} & \begin{tabular}[b]{@{}c@{}}$E$, unknown\\kept\end{tabular} & \begin{tabular}[b]{@{}c@{}}Flagged\\(\%)\end{tabular} & \begin{tabular}[b]{@{}c@{}}JSON\\fail (\%)\end{tabular} & \begin{tabular}[b]{@{}c@{}}Hit\\cap (\%)\end{tabular} & \begin{tabular}[b]{@{}c@{}}Reasoning\\(\%)\end{tabular} & \begin{tabular}[b]{@{}c@{}}Output\\tokens\end{tabular} & \begin{tabular}[b]{@{}c@{}}Cost\\(\$)\end{tabular} \\\midrule
lowest, 16k & 52.80 & 96.78 & 103.46 & 184.76 & 288.22 & 90 & 72 & 4 & 0 & 4,796 & 0.95 \\
default, 16k & 24.52 & 25.90 & 34.58 & 4.90 & 39.48 & 12 & 70 & 8 & 100 & 8,831 & 1.71 \\
default, 32k & 15.44 & 20.14 & 28.26 & 8.84 & 37.10 & 6 & 78 & 0 & 98 & 9,487 & 1.83 \\
\bottomrule\end{tabular}\end{table*}

Gemini 3.8 at its lowest setting writes long lists in which most records lack a valid class and some coordinates repeat, so the list count is far worse than its stated count ($+182.5\%$ in the same response; Figure~\ref{fig:density}). The four-part degeneracy check of Section~\ref{sec:method} flags 45 of its 50 dense-scene lists. For 40 of them the readout falls back to the stated count; the other 5 have no stated count and keep the list count with a status flag. On the GPT-6.1 and Sonnet 5.5 list-first responses it flags none.

For Gemini 3.8, degeneration depends mainly on the reasoning setting. At the same 16k output cap, Gemini 3.8's default setting reduces flagged lists from 90\% to 12\% and the readout error from 52.80 to 24.52 ($-53.6\%$ $[-72.4,-29.5]$; Table~\ref{tab:setting}). Doubling the cap to 32k gives a point estimate of 15.44, a further change that is not significant. Even then, Gemini 3.8's dense-count error is about ten times GPT-6.1's, and none of its 50 counts is exact. Strict JSON parsing fails for 70\% to 78\% of Gemini 3.8 responses at every setting, so the declared record salvage of Section~\ref{sec:setup:contract} is a necessary part of the decoder, and its share is reported.

On the list-first responses of the readout experiment, applying the same check to existing responses recovers 68\% (SCP) and 58\% (SCP-dense) of Gemini 3.8's error relative to reading its list, and 42\% relative to the stated count on FSC-147. For Sonnet 5.5, whose lists are not degenerate, the cost of the check reaches $+23\%$ on SCP and $+13\%$ on SCP-dense (not significant; intervals $[0,57]$ and $[0,26]$). On the open model Qwen3-VL, with thresholds unchanged, it lowers SCP-dense error by 7\% $[-13,-1]$ relative to the stated count. Gemini 3.8's missing class fields follow a pattern: a record is more likely to omit its type when the previous record has the same type, and a row-by-row scan that interleaves types cuts its count error by 60\%, while grouping by type makes it worse.

\subsection{Repair the set: delete below the threshold, keep geometry separate}\label{sec:res:set}

\begin{table*}[tp]\centering\footnotesize\setlength{\tabcolsep}{4pt}
\caption{Confirmation by a partner call, split into selection and geometry (Propositions~\ref{prop:identity} and \ref{prop:geometry}). Each source has 9 base/partner pairs (3 base models $\times$ same-model or one of two other models), 60 images, one call each; selection, geometry and their total are per-image reductions of $E$ (positive is better). Both identities hold exactly on every image of every pair. \emph{Sign from 5 labels}: $p_1$ is estimated on 5 labelled images and Proposition~\ref{thm:intersect} predicts whether deletion lowers $E$ on the other 55; share of correct predictions over 200 draws per pair, averaged over the pairs (Section~\ref{sec:res:set}). \emph{Deletion helps}: pairs whose selection term is positive. \emph{Geometry $+$ / $-$}: pairs whose geometry term is positive / negative. \emph{Sign flips}: pairs in which the total with averaged coordinates has the opposite sign of the deletion-only selection term. Per-pair values are in Table~\ref{tab:intersect_full}.}\label{tab:intersect}
\begin{tabular}{@{}lrrrrrrr@{}}\toprule
Source & \begin{tabular}[b]{@{}c@{}}Sign from\\5 labels (\%)\end{tabular} & \begin{tabular}[b]{@{}c@{}}Deletion\\helps\end{tabular} & \begin{tabular}[b]{@{}c@{}}$p_1$\\range\end{tabular} & \begin{tabular}[b]{@{}c@{}}Selection\\range\end{tabular} & \begin{tabular}[b]{@{}c@{}}Geometry\\$+$ / $-$\end{tabular} & \begin{tabular}[b]{@{}c@{}}Geometry\\range\end{tabular} & \begin{tabular}[b]{@{}c@{}}Sign\\flips\end{tabular} \\\midrule
DocLayNet & 74 & 6 & 0.11 to 0.64 & $-$2.2 to 2.1 & 6 / 2 & $-$1.1 to 3.1 & 1 \\
DOTA & 98 & 9 & 0.05 to 0.22 & 4.4 to 16.3 & 7 / 2 & $-$1.5 to 3.9 & 0 \\
FSC-147 & 81 & 2 & 0.39 to 0.90 & $-$21.8 to 0.3 & 8 / 0 & 0.0 to 0.4 & 0 \\
SKU-110K & 96 & 9 & 0.00 to 0.39 & 10.8 to 68.2 & 7 / 2 & $-$14.1 to 13.8 & 0 \\
GroundCUA & 86 & 6 & 0.06 to 0.72 & $-$16.8 to 37.7 & 4 / 5 & $-$7.7 to 15.4 & 1 \\
ScreenParse & 63 & 8 & 0.19 to 0.50 & 0.0 to 2.1 & 5 / 2 & $-$0.5 to 0.9 & 0 \\
\midrule
All 54 pairs & 83 & 40 & 0.00 to 0.90 & $-$21.8 to 68.2 & 37 / 13 & $-$14.1 to 15.4 & 2 \\
\bottomrule\end{tabular}\end{table*}

\begin{figure*}[tp]
\centering
\includegraphics[width=\textwidth,alt={(a) Scatter of selection gain against the marginal matching rate p1 of deleted records for 54 base and partner pairs, with a dashed line at p1 equal to one half; every pair left of the line has a positive gain and every pair right of it a negative gain. (b) Bars of the reduction of set error from selecting among three calls of one model on DocLayNet, DOTA, FSC-147 and SKU-110K: the constrained oracle reduces it most, the cross-fitted logistic selector and the selector fitted on 10 labelled images less.}]{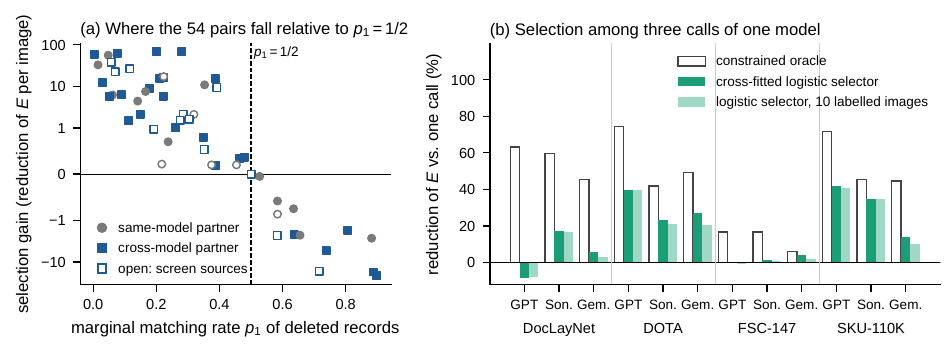}
\caption{Selection and the deletion threshold. (a) Selection gain of deleting unconfirmed records, as a reduction of set error $E$ per image (positive: deletion lowers $E$), against the marginal matching rate $p_1$ of the deleted records, for the 54 base/partner pairs (symmetric log scale; grey circles: same-model partner; blue squares: cross-model partner; open markers: screen sources). With equal error costs the gain per image equals $\mathbb{E}[\Delta c](1-2p_1)$ (Proposition~\ref{thm:intersect}), so its sign follows the side of $p_1=1/2$; the panel shows where each pair falls. (b) Reduction of $E$ relative to one call when selecting among three calls of one model (detection sources, 60 images per cell): the constrained oracle, the cross-fitted logistic selector, and the logistic selector fitted on 10 labelled images (median of 40 draws). GPT, Son., Gem.: GPT-6.1, Sonnet 5.5, Gemini 3.8.}
\label{fig:select}
\end{figure*}

\paragraph{Deletion and geometry, measured separately.} Table~\ref{tab:intersect} (per-pair values in Table~\ref{tab:intersect_full}) splits the effect of confirming a base call with a partner call into the selection term of Proposition~\ref{prop:identity} and the geometry term of Proposition~\ref{prop:geometry}, for 54 base/partner pairs on six sources. Both identities hold exactly on every image, so the selection term of a pair has the sign of $1-2p_1$ by construction (Figure~\ref{fig:select}a). What the data add is where $p_1$ lies: 0.00 to 0.39 on SKU-110K and 0.05 to 0.22 on DOTA, where deletion removes 4 to 68 errors per image, against 0.39 to 0.90 on FSC-147. On FSC-147, where models under-report, $p_1$ is 0.39 to 0.90 and deletion hurts in 7 of 9 pairs, which is why counting families keep a single call. Coordinate averaging is positive in 37 pairs and negative in 13. In two pairs it reverses the sign of the deletion-only result: DocLayNet with GPT-6.1 confirmed by Sonnet 5.5 (selection $+0.80$, geometry $-1.10$ per image) and GroundCUA with Sonnet 5.5 confirmed by itself (selection $-0.87$, geometry $+1.20$). These are the two pairs in which an earlier check of the threshold on intersections with averaged coordinates failed; separating the terms accounts for both.

\paragraph{The threshold from a few labels.} The useful question is whether $p_1$ measured on a handful of labelled images predicts what deletion will do on other images. For each pair we drew 5 labelled images 200 times, estimated $p_1$ on them, and predicted from Proposition~\ref{thm:intersect} whether deleting unconfirmed records lowers set error on the other 55 images of the pair. The prediction is right in 83\% of draws (80\% with 3 labelled images, 87\% with 10), and the loss against the better decision on the other images is 0.33 errors per image (0.48 and 0.15). The shipped family default alone, which deletes on over-reporting families and keeps on FSC-147 without any label, is right in 85\% of draws with a loss of 0.43. The label rule of Step 7 combines the two: it starts from the family default and switches only if the labelled images favour the other decision by more than one standard error. It is right in 86\%, 87\% and 88\% of draws with 3, 5 and 10 labelled images, with losses of 0.22, 0.17 and 0.10 errors per image. In the 8 pairs whose $p_1$ lies within 0.1 of $1/2$, the gain from deleting is small and its sign is noisy: the plain threshold is right in 47\% of draws there, and the family default and the Step 7 rule, which lean towards deletion, in 34\% and 37\%. Table~\ref{tab:intersect} gives the share by source.

\paragraph{Which partner helps.} Proposition~\ref{prop:partner} says a partner helps if it rejects more of the base call's false records than it loses of the true ones. Cross-model partners separate the two better than same-model partners in 32 of 36 base/partner comparisons on the detection and screen-parsing sources. The four exceptions are GPT-6.1's two much weaker partners on FSC-147, and Gemini 3.8 confirming Sonnet 5.5 on FSC-147 and SKU-110K, where Gemini 3.8 confirms few of Sonnet 5.5's true records. The misses are consistent with a component shared across models: of the objects that GPT-6.1, the highest-recall model, misses in all three of its calls, a single call of either other model recovers only 0.7\% to 9\% on SKU-110K, DOTA and FSC-147, and 18\% to 26\% on DocLayNet. New information flows mainly from higher-recall to lower-recall models.

\subsection{Repair the view; fixes that do not help}\label{sec:res:view}

\paragraph{Zoom and tiles.} On ScreenSpot-Pro (200 items), two-stage zoom (Step 4) raises the click hit rate from 64.5\% to 87.5\% for Gemini 3.8 ($+23$ points $[16.5,29.5]$, a 65\% cut in the miss rate) and from 78\% to 87\% for Sonnet 5.5 ($+9$ $[5,13.5]$, a 41\% cut; 45\% on the 196 items that every arm of Table~\ref{tab:common} answered); for GPT-6.1, 88.5\% to 90.0\% is not significant. Zooming only when two different models disagree, a cross-model version of the gated zoom in Table~\ref{tab:common}, is less accurate than zooming every item for Sonnet 5.5 (2.3 to 2.7 points) and Gemini 3.8 (9.6 to 12.3 points), not significantly different for GPT-6.1, and more expensive. Our tiling arm on FSC-147 used $2\times2$ crops that overlap by 10\% of the side, each enlarged $2\times$ and sent in a separate call. It raises the tokens per object side for every object, but its overlap is smaller than one object for the larger objects (median exemplar side 17.5\% of the image side in the largest third), so it does not meet the companion paper's safe-decomposition condition that every target lie inside some view \citep{shi2026zoomin}. The results follow that condition. On the smallest third of images, Sonnet 5.5's count error falls by 70\% against whole-image lists; on the largest third, where many objects lie wholly inside no crop, GPT-6.1's rises by 78\%; over all images Sonnet 5.5's change ($-21\%$, $[-50,27]$) is not significant. The small-object gain mixes more tokens per object with less content per call, and the large-object loss is consistent with objects that no crop contains rather than with object size as such. On screens, where tiles overlap by 6\%, objects cut by a tile boundary lose 0.16 to 0.42 in report probability in all 6 cells, and separate calls per tile beat one call with several tiles for GPT-6.1 ($-15\%$, $[-23,-5]$). For native-resolution click crops, downscaling lowers Sonnet 5.5's hit rate by 9 points $[3,16]$, the loss for GPT-6.1 reaches 3 points ($[0,7]$, not significant), Gemini 3.8 shows no significant response to resizing, consistent with the fixed token grid measured for it in the companion paper, and enlarging these crops does not help.

\paragraph{Reports from a view planner.} In the crop calls of the Review experiment (Table~\ref{tab:experiments}) the pipeline receives, for each image, a whole-image report and four reports from four native-resolution crops (6\% overlap, one call per crop), on 30 GroundCUA screens and 30 SKU-110K images, scored at IoU $\ge0.5$ without class. Merging the four crop reports alone lowers set error relative to one whole-image call for GPT-6.1 on screens ($-19.4\%$ $[-27.5,-10.6]$) and for Sonnet 5.5 on SKU-110K ($-64.0\%$ $[-70.6,-56.6]$), raises it for Sonnet 5.5 on screens ($+44.9\%$) and for Gemini 3.8 on SKU-110K ($+24.7\%$), and is not significant in the other two cells. Confirming the whole-image report with the merged crops under the shipped rule, in which a crop confirms or rejects only the records it contains wholly, lowers set error in 4 of 6 cells: GPT-6.1 ($-22.8\%$) and Gemini 3.8 ($-20.3\%$) on screens, and GPT-6.1 ($-45.5\%$) and Sonnet 5.5 ($-36.6\%$) on SKU-110K. In these 4 cells it is also better than confirmation by a second whole-image call ($-11.8\%$ to $-22.6\%$), and better than the shipped three-call unanimity in 3 of them ($-8.0\%$ to $-20.2\%$; Gemini 3.8 on screens $-4.7\%$, not significant), at the price of five calls instead of three. In the other two cells it raises set error: Sonnet 5.5 on screens ($+12.9\%$) and Gemini 3.8 on SKU-110K ($+37.9\%$). The threshold of Proposition~\ref{thm:intersect} accounts for both groups. Among the whole-image records that the crops do not confirm, the share matched to a true object ($p_1$) is 0.03 to 0.24 in the four cells where confirmation helps, and 0.62 and 0.76 in the two where it hurts, as the identity requires. The plain rule, in which every record needs confirmation, differs from the shipped rule by at most 4 percentage points, because with a 6\% overlap few records lie wholly inside no crop.

\paragraph{Grids and coordinates.} On SCP, the systematic coordinate offset of current models is at most 0.3 pixels, with slope 1.000 and spread 0.9 to 1.5 pixels, against a matching tolerance of 30 pixels. A grid, ticks or cell labels therefore have no coordinate error to correct: in every arm, localization changes by at most 0.23 pixels, and at the lowest setting they do not significantly change detection for GPT-6.1 and Sonnet 5.5. For Gemini 3.8 at its lowest setting, a grid lowers set error by 3.5 per image by reducing records that omit their class, and a row-by-row scan lowers it by 11.5; at its default setting the grid makes detection and localization slightly worse. ``No overlay'' is therefore the shipped default, chosen from these measurements; the theory does not forbid an overlay from helping a model whose list format it changes. Markers also make Sonnet 5.5 write ``diamond'' more often ($+0.3$ per image). An overlay adds no Shannon information and can only change how the model uses its input \citep{shi2026zoomin}; the two changes we saw are of that kind.

\paragraph{Wording, instructions and review.} ``Do not miss any'' wording, compared with a symmetric instruction not to miss or invent, has no measurable effect on SCP for any of the three models. Asking the model to localize each item relative to structures in the image has no significant effect, because current models have almost no location misses on these scenes. Same-view self-review, in which the model edits its own draft, confirms true and false records at nearly the same rate ($h_+\approx h_-$) and has no net gain on screen parsing; on FSC-147 it gave significant gains for 2 of 3 models, which we do not explain. Pairwise comparators that ask a model which of two reports is right judge whether something is an object rather than whether the box is precise, and are not used. Reasoning is not uniformly helpful: on 20 GroundCUA screens, GPT-6.1's default setting raises set error by 18.7 per screen $[5.8,34.3]$, through 16.3 additional false records and no fewer misses.

\subsection{Detect errors: they persist, so ask someone else}\label{sec:res:persist}

\paragraph{Persistence.} Object-level errors repeat across calls. Intraclass correlations of reporting across repeated calls are 0.60 to 0.98 for true objects and 0.23 to 0.93 for false reports. On the detection sources, a binomial model with a common reporting probability is rejected in all 12 dataset and model cells, while a beta-binomial model fits. The share of misses that recur in every repeated call is 0.76 to 0.99 on the detection sources and 0.73 to 0.94 on the screen sources. In clicks, 64\% to 83\% of a model's errors are repeated by a second call of the same model within the agreement tolerance.

\begin{figure*}[tp]
\centering
\includegraphics[width=\textwidth,alt={Three panels, one per model, of risk-coverage curves of click alarms on 260 items: share of items acted on against miss rate on those items, for alarms from a second call of the same model (grey) and from a second, cheaper model (blue), on first clicks (solid) and after two-stage zoom (dashed), with rings at the default threshold and diamonds for acting on every item. Alarms from a different model reach much lower miss rates than alarms from a second call of the same model, at the price of setting more items aside.}]{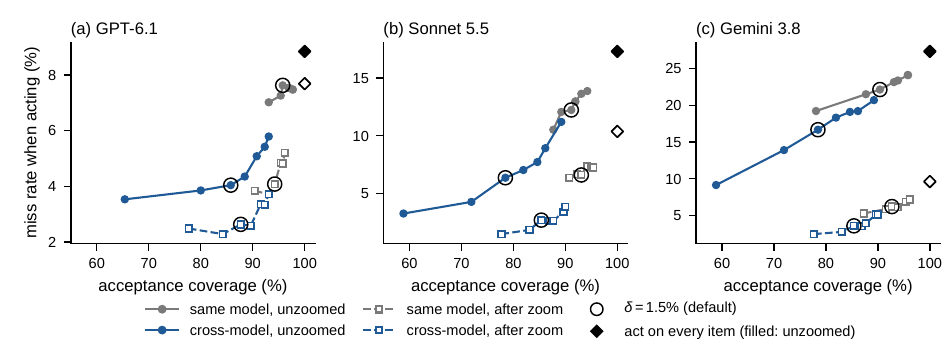}
\caption{Risk-coverage of click alarms (260 items; point estimates). An alarm sets an item aside when the acted-on point and a second point for the same item are farther apart than $\delta$ of the longer image side; the system acts on the other items. \emph{Acceptance coverage} is the share of items acted on, and the \emph{miss rate when acting} the share of those items whose point is outside the target; lower and further right is better. Each curve traces $\delta\in\{0.5,1,1.5,2,3,5,8\}\%$, and a smaller $\delta$ sets more items aside. Grey: the second point comes from a second call of the same model; blue: from the partner, the cheapest other model (Gemini 3.8 for GPT-6.1 and Sonnet 5.5, Sonnet 5.5 for Gemini 3.8). Solid lines and filled markers: first clicks on the whole screenshot; dashed lines and open markers: both points after two-stage zoom. Rings: the shipped default $\delta=1.5\%$ (Table~\ref{tab:alarm}). Diamonds: acting on every item. The cross-model alarm reaches lower miss rates than the same-model alarm, at the price of setting more items aside.}
\label{fig:alarm}
\end{figure*}

\paragraph{A second model is a better alarm.} Persistence means that a second call of the same model rarely disagrees with the first where the first is wrong. A second model shares only part of the propensity. With the partner set by a fixed rule as the cheapest other model, the share of errors in which both calls agree on the same wrong point falls from 83\% to 39\% for GPT-6.1, from 64\% to 29\% for Sonnet 5.5, and from 73\% to 48\% for Gemini 3.8. In the risk-coverage analysis (Figure~\ref{fig:alarm}), at $\delta=1.5\%$, the same-model alarm catches 17\% of GPT-6.1's errors (4 of 23) and the cross-model alarm catches 61\% (14 of 23); for Sonnet 5.5 the shares are 36\% and 71\% (Table~\ref{tab:alarm}). The cross-model alarm sets more items aside, and Figure~\ref{fig:alarm} shows the trade-off: where the two curves overlap, its miss rate at matched acceptance coverage is lower by 0.5 to 2.7 points (point estimates on 260 items, without intervals; for GPT-6.1 at 93.1\% acceptance coverage it catches 9 of the 23 errors against 6 for the same-model alarm), and for GPT-6.1 and Sonnet 5.5 it reaches operating points with a miss rate of 3\% to 4\%, below the lowest same-model values (7.0\% and 10.5\%). After two-stage zoom the same pattern holds, so the advantage comes from the second model and not from zoom alone. We zoomed each model's second click as well (772 calls), so that both alarms compare zoomed points. At $\delta=1.5\%$ a second zoomed call of the same model catches 40\% to 50\% of the remaining errors, and the partner's zoomed point catches 68\% to 78\%, with miss rates of 2.6\% to 3.6\% on the 85\% to 88\% of items acted on (Table~\ref{tab:alarm}). The same-model alarm again acts on more items (93\% to 94\%); where the two curves overlap, the cross-model miss rate at matched acceptance coverage is lower by 0 to 1.5 points.

\begin{table*}[tp]\centering\footnotesize\setlength{\tabcolsep}{4pt}
\caption{Disagreement alarms for clicks (ScreenSpot-Pro and ScreenSpot-v2, 260 items; partner set by a fixed rule as the cheapest other model; point estimates). The alarm compares the acted-on point with a second point for the same item and fires when the two are farther apart than $\delta=1.5\%$ of the longer image side; in this risk-coverage analysis the item is then set aside instead of acted on (in Algorithm~\ref{alg:clicks} it triggers a comparison instead). Unzoomed alarms compare first clicks; after zoom, the acted-on point is the zoomed first click, and the second point is the same model's zoomed second click or the partner's zoomed first click. Miss rate, all items: miss rate when acting on every item. Acceptance coverage: share of items acted on. Miss rate when acting: share of acted-on items whose point is outside the target. Errors caught: share of the acted-on prediction's misses that the alarm sets aside. Other values of $\delta$ are in Figure~\ref{fig:alarm}, and miss rates at matched acceptance coverage in Table~\ref{tab:alarm_matched}.}\label{tab:alarm}
\begin{tabular}{@{}lllrrrr@{}}\toprule
Model & Partner & Alarm & \begin{tabular}[b]{@{}c@{}}Miss rate,\\all items (\%) $\downarrow$\end{tabular} & \begin{tabular}[b]{@{}c@{}}Acceptance\\coverage (\%) $\uparrow$\end{tabular} & \begin{tabular}[b]{@{}c@{}}Miss rate when\\acting (\%) $\downarrow$\end{tabular} & \begin{tabular}[b]{@{}c@{}}Errors\\caught (\%) $\uparrow$\end{tabular} \\\midrule
GPT-6.1 & Gemini 3.8 & same model, unzoomed & 8.9 & 95.8 & 7.6 & 17 \\
 &  & cross-model, unzoomed & 8.9 & 85.8 & 4.0 & 61 \\
 &  & same model, after zoom & 7.7 & 94.2 & 4.1 & 50 \\
 &  & cross-model, after zoom & 7.7 & 87.7 & 2.6 & 70 \\
\midrule
Sonnet 5.5 & Gemini 3.8 & same model, unzoomed & 17.3 & 91.1 & 12.2 & 36 \\
 &  & cross-model, unzoomed & 17.3 & 78.5 & 6.4 & 71 \\
 &  & same model, after zoom & 10.4 & 93.1 & 6.6 & 41 \\
 &  & cross-model, after zoom & 10.4 & 85.4 & 2.7 & 78 \\
\midrule
Gemini 3.8 & Sonnet 5.5 & same model, unzoomed & 27.3 & 90.4 & 22.1 & 27 \\
 &  & cross-model, unzoomed & 27.3 & 78.5 & 16.7 & 52 \\
 &  & same model, after zoom & 9.6 & 92.7 & 6.2 & 40 \\
 &  & cross-model, after zoom & 9.6 & 85.4 & 3.6 & 68 \\
\bottomrule\end{tabular}\end{table*}

\subsection{Repair a click by choosing among models}\label{sec:res:choose}

\begin{table*}[tp]
\centering\footnotesize\setlength{\tabcolsep}{4pt}
\caption{Choosing one click among several calls, on 600 ScreenSpot-Pro items that share no item or screenshot with the 260 click items of the GUI experiment. (a) Every flow, with every item in the denominator; a flow that returns no point counts as a miss, which happened only in single-model flows (column No point). Calls and cost: what the flow needs per item, every attempt counted, at list prices. The three-model flows use GPT-6.1 and Sonnet 5.5 at the default setting and Gemini 3.8 at the lowest, unless marked ``all lowest''. Cluster vote: the largest cluster of points within $\delta$ wins, ties go to the earliest model (GPT-6.1, Sonnet 5.5, Gemini 3.8), and the earliest point of the winning cluster is returned; computed from the stored calls, no scorer. (b) Paired comparisons: $\Delta$ is the miss rate of the first flow minus that of the second, in percentage points (negative favours the first), with a 95\% paired bootstrap interval over screenshots, items repaired and broken by the first flow, and the exact McNemar $p$ (Holm-adjusted over the two comparisons marked $\dagger$). \textbf{Bold}: the paired 95\% interval excludes 0 in the favourable direction; \underline{underlined}: it excludes 0 in the unfavourable direction.}\label{tab:choose}
\begin{tabular}{@{}lrrrrr@{}}\toprule
\multicolumn{6}{@{}l}{\emph{(a) Flows}} \\
Flow & Misses & No point & Miss rate (\%) & Calls & Cost / item (\$) \\\midrule
GPT-6.1 zoom, lowest & 86 & 4 & 14.33 & 1.99 & 0.0081 \\
Sonnet 5.5 zoom, lowest & 104 & 15 & 17.33 & 1.98 & 0.0057 \\
Gemini 3.8 zoom, lowest & 88 & 0 & 14.67 & 2.00 & 0.0054 \\
Sonnet 5.5 zoom, default & 80 & 0 & 13.33 & 2.00 & 0.0066 \\
GPT-6.1 zoom, default & 82 & 2 & 13.67 & 2.00 & 0.0085 \\
GPT-6.1 zoom $\times$3, default; cluster vote & 76 & 1 & 12.67 & 5.99 & 0.0255 \\
Crop-first, GUI-Lens style (our implementation) & 69 & 0 & 11.50 & 2.00 & 0.0117 \\
Three models; cluster vote & 74 & 0 & 12.33 & 6.00 & 0.0205 \\
Three models, all lowest; score & 66 & 0 & 11.00 & 8.94 & 0.0244 \\
Three models, all lowest; score and comparison & 61 & 0 & 10.17 & 9.42 & 0.0261 \\
Three models; score & 61 & 0 & 10.17 & 8.99 & 0.0257 \\
GUI-grounding tier (Alg.~\ref{alg:clicks}): three models; score and comparison & 56 & 0 & 9.33 & 9.45 & 0.0273 \\
\bottomrule\end{tabular}

\medskip

\begin{tabular}{@{}lrrrr@{}}\toprule
\multicolumn{5}{@{}l}{\emph{(b) Paired comparisons}} \\
First flow vs.\ second flow & $\Delta$ (pp) & 95\% interval & Repaired / broken & $p$ \\\midrule
GUI-grounding tier vs.\ GPT-6.1 zoom, default & {\bfseries\boldmath $-$4.33} & [$-$6.33, $-$2.33] & 34 / 8 & $7\times10^{-5}$ \\
GUI-grounding tier vs.\ GPT-6.1 zoom $\times$3, vote$^\dagger$ & {\bfseries\boldmath $-$3.33} & [$-$5.17, $-$1.50] & 27 / 7 & 0.002 \\
GUI-grounding tier vs.\ crop-first baseline$^\dagger$ & $-$2.17 & [$-$4.17, 0.00] & 27 / 14 & 0.060 \\
GUI-grounding tier vs.\ three models, cluster vote & {\bfseries\boldmath $-$3.00} & [$-$4.67, $-$1.50] & 22 / 4 & $5\times10^{-4}$ \\
GUI-grounding tier vs.\ three models, score & $-$0.83 & [$-$2.00, 0.17] & 8 / 3 & 0.227 \\
All lowest, score and comparison vs.\ all lowest, score & $-$0.83 & [$-$2.00, 0.33] & 9 / 4 & 0.267 \\
GPT-6.1 zoom, default vs.\ GPT-6.1 zoom, lowest & $-$0.67 & [$-$2.67, 1.33] & 19 / 15 & 0.608 \\
Sonnet 5.5 zoom, default vs.\ Sonnet 5.5 zoom, lowest & {\bfseries\boldmath $-$4.00} & [$-$6.17, $-$1.83] & 34 / 10 & $4\times10^{-4}$ \\
\bottomrule\end{tabular}
\end{table*}

On the 260 click items of the GUI experiment, the GUI-grounding tier of Section~\ref{sec:method:clicks} missed 11 targets against 16 for GPT-6.1's two-stage zoom at its default setting. Table~\ref{tab:choose} reports it, with the baselines, on 600 further ScreenSpot-Pro items that share no item or screenshot with those 260. All 13,120 requests completed (13,126 attempts with retries; \$44.93 at list prices for all arms), none of the fallbacks for a failed call, zoom, score or comparison was used, and every item stays in the denominator. Single-model flows sometimes returned no point, counted as a miss (Table~\ref{tab:choose}); the three-model flows always returned one.

\paragraph{The tier against single-model practice.} On the 600 items the tier misses 56 targets (9.33\%), against 82 (13.67\%) for GPT-6.1's two-stage zoom at its default setting: $-4.33$ points $[-6.33,-2.33]$, with 34 items repaired and 8 broken (exact McNemar $p=7\times10^{-5}$). At about the same cost, three independent default zooms of GPT-6.1 with a cluster vote miss 76, so the tier's miss rate is lower by 3.33 points ($-3.33$ $[-5.17,-1.50]$, Holm-adjusted $p=0.002$). Against a crop-first baseline that we implemented in the style of GUI-Lens \citep{fu2026guilens}, without its OCR and detector anchors or its verification step and with our own prompts, the tier misses 56 against 69: the difference is $-2.17$ points with an interval that reaches zero ($[-4.17,0.00]$; Holm-adjusted $p=0.060$), at 2.3 times that baseline's cost.

\paragraph{Where the gain comes from.} Proposition~\ref{prop:three} splits the tier's 56 misses into 43 items on which no model's zoomed point hits ($F$), 3 on which a correct point exists but no cluster representative is correct ($L$), and 10 on which the comparison picks the wrong representative ($W$). With the scorer's score alone, $W$ is 15 and the other two terms are unchanged. The comparison therefore repairs 8 items and breaks 3: its increment is $-0.83$ points $[-2.00,0.17]$, and $-0.83$ $[-2.00,0.33]$ with all three models at their lowest setting, neither significant. On the representative pairs in which exactly one point hits, a single comparison call names the right one in 85.9\% of calls, and the two orders agree on 71 of 78 pairs, 64 of them correctly; on the 260 items of the GUI experiment the single-call rate was 88\%. Against three GPT-6.1 zooms, most of the gain is in $F$. One default zoom of GPT-6.1 has no correct point on 82 items; three of them, at about the tier's cost, still have none on 62; the tier's three models have none on 43. Proposition~\ref{prop:coverage} names the rate behind this: on the 82 items that GPT-6.1 missed, a second GPT-6.1 zoom hits on 18\% and a third, after two misses, on 7.5\%, while Sonnet 5.5 hits on 29\% and Gemini 3.8 on 37\%, and Gemini 3.8 still hits on 26\% of the 58 items that both GPT-6.1 and Sonnet 5.5 missed. On these items, repeating GPT-6.1 adds little coverage, consistent with its persistent errors (Section~\ref{sec:res:persist}); the partners rescue more here, which Proposition~\ref{prop:coverage} does not guarantee in general. The partners' rates are close to those on the 260 click items of the GUI experiment, where one GPT-6.1 zoom has no correct point on 16 items: Sonnet 5.5 hits on 4 of them (25\%), Gemini 3.8 on 6 (38\%), and Gemini 3.8 on 3 of the 12 that both GPT-6.1 and Sonnet 5.5 missed (25\%). Given the 82 items GPT-6.1 misses here, the rates from those items give $82\times0.75\times0.75\approx46$ items with no correct candidate among the 600, against 43 observed, although GPT-6.1 misses more than twice as often here (13.7\% against 6.2\%). Against the three-zoom vote, whose 76 misses are 62 items with no correct candidate and 14 later losses, 19 of the tier's 20 fewer misses lie in candidate coverage.

\paragraph{Coverage needs selection.} A plain cluster vote over the same three models' points, with no scorer (the largest cluster wins, ties go to the earliest source, whose point is returned), computed from the stored calls, misses 74 items (12.33\%, \$0.0205 per item), not significantly different from three GPT-6.1 zooms ($-0.33$ points $[-1.83,1.17]$). It has the tier's candidates, so its 74 misses are the same 43 items with no correct candidate and 31 later losses. Choosing by the scorer's score and the comparison cuts those later losses to 13 (56 misses, 9.33\%, \$0.0273 per item): the tier's miss rate is lower than the vote's by 3.00 points ($-3.00$ $[-4.67,-1.50]$), with 22 items repaired and 4 broken (exact McNemar $p=5\times10^{-4}$). The other two models supply the candidates; the selection turns them into fewer misses.

\paragraph{Items without an alarm.} On 499 of the 600 items the three points form one cluster and no comparison is made. The tier misses 29 of them (5.8\%), which is 52\% of its misses. An action applied to all of them, or through a trigger independent of whether the base misses, improves accuracy only if $r_{\mathrm{brk}}<e_T\,r_{\mathrm{fix}}/(1-e_T)$ with $e_T=5.8\%$ (Proposition~\ref{prop:switch}): about 3.1\% of the correct answers it touches if it repairs half of the misses. The scorer's score alone misses 32 of the other 101 items (31.7\%), which is where the comparison acts; the 260 click items of the GUI experiment show the same pattern (10 of 39 against 3 of 221). Nine clusters are chained, with two members farther apart than $\delta=1.5\%$, and none is wider than 2.13\% of the longer side.

\paragraph{Settings.} On the 600 items the default setting helps Sonnet 5.5 (misses $-4.00$ points $[-6.17,-1.83]$ against its lowest setting) and does not change GPT-6.1 significantly ($-0.67$ $[-2.67,1.33]$), the same direction as on the 260 items of the GUI experiment; the default setting also doubles the output cap (16,384 to 32,768 tokens).

\paragraph{The crop-first baseline errs elsewhere.} The crop-first baseline and the tier miss different items: both miss 42, only the baseline 27, only the tier 14. Of those 14, the baseline supplies a correct point on 7 items where all three of the tier's models miss, so it is a new source in the sense of Proposition~\ref{prop:coverage}: added to the tier's menu, it lowers the items with no correct candidate from 43 to 36 ($\phi=7/43$). On the other 7 the tier had a correct candidate and lost it ($L$, 2 items) or chose wrongly ($W$, 5). Conversely, on 14 of the 27 items that only the baseline misses, its wrong point lies within $\delta$ of one of the tier's candidates, so adding it to the vote would mostly strengthen wrong clusters (on 12 of the 14 every nearby candidate is wrong).

\paragraph{What a fourth source can add.} On the 57 items where the crop-first point and the tier's point are more than $\delta$ apart, the crop-first point is right and the tier's wrong on 10, the reverse holds on 25, both are right on 12 and both wrong on 10. A rule that takes the crop-first point on $a$ of the first 10 items and on $b$ of the 25 changes the number of misses by $b-a$, the count form of Proposition~\ref{prop:switch}. Within this two-candidate, threshold-triggered menu the largest possible improvement is therefore 10 items (1.67 points). The comparison's accuracy on the representative pairs above does not determine $a$ and $b$ on these different pairs, so we evaluated an enlarged candidate menu directly. We added the crop-first point as a fourth candidate, ranked last among sources, scored by the same scorer and handled by the same rules, on the same 600 items (717 further calls, \$1.44). Items with no correct candidate fall from 43 to 36 ($\phi=7/43$). Lost representatives rise from 3 to 7 and wrong choices from 10 to 11. Of the 7 items where only the crop-first point is correct, the four-source tier gets 3 right; on 3 the point joins a cluster of wrong points and loses the representative choice to a wrong member with the same score, and on 1 the comparison picks a wrong representative. On the other items the added point repairs 1 and breaks 2: on one it displaces a correct representative, on the other the comparison turns to a wrong one. The four-source tier misses 54 items against 56 ($-0.33$ points $[-1.17,0.34]$; 4 repaired, 2 broken) at \$0.041 per item (12.5 calls) instead of \$0.027. Proposition~\ref{prop:three} accounts for the result term by term: the added source pays in $F$ and is charged in $L$ and $W$. The tier of Algorithm~\ref{alg:clicks} therefore stays at three sources, and the crop-first baseline is reported as a baseline.

\subsection{Judge: selection has room left, but labels do not close it}\label{sec:res:select}

\begin{table*}[tp]\centering\footnotesize\setlength{\tabcolsep}{2pt}
\caption{Selection among three calls of one model (detection sources, list interface, lowest setting, 60 images per cell; reduction of $E$ relative to the first call, in \%, higher is better). The constrained oracle optimizes over the same feasible family as the selectors (Proposition~\ref{prop:oracle}); the unrestricted value lets two members of one cluster match two targets and is not attainable. The two oracles differ on 25 of 720 images. Selectors are logistic or boosted-tree record gates on output features; they read no labels at inference but are fitted with labels: cross-fitted (5 folds by image, persisted), in-sample, or on 10 labelled images (median over 40 draws, each evaluated on the images it did not use). Gap: constrained oracle minus cross-fitted logistic, with a 95\% image-bootstrap interval for the fitted selector; it mixes information absent from the features with selector and estimation shortfall (Proposition~\ref{prop:order}).}\label{tab:select}
\begin{tabular}{@{}llrrrrrrr@{\hspace{8pt}}r@{}}\toprule
& & \multicolumn{2}{c}{Oracles} & \multicolumn{5}{c}{Selectors fitted with labels} & \\\cmidrule(lr){3-4}\cmidrule(lr){5-9}
Source & Model & Constrained & Unrestricted & \begin{tabular}[b]{@{}c@{}}Cross-fit\\logistic\end{tabular} & \begin{tabular}[b]{@{}c@{}}Cross-fit\\trees\end{tabular} & \begin{tabular}[b]{@{}c@{}}In-sample\\logistic\end{tabular} & \begin{tabular}[b]{@{}c@{}}In-sample\\trees\end{tabular} & \begin{tabular}[b]{@{}c@{}}10 labels\\logistic\end{tabular} & \begin{tabular}[b]{@{}c@{}}Gap\\{}[95\% CI]\end{tabular} \\\midrule
DocLayNet & GPT-6.1 & 63.2 & 63.2 & $-$8.4 & $-$7.4 & 1.6 & 42.8 & $-$7.7 & 71.6 [60.1, 87.4] \\
DocLayNet & Sonnet 5.5 & 59.8 & 59.8 & 17.3 & 14.8 & 22.6 & 50.1 & 16.6 & 42.5 [34.2, 54.0] \\
DocLayNet & Gemini 3.8 & 45.2 & 45.2 & 5.9 & 1.3 & 6.9 & 37.6 & 3.2 & 39.3 [30.1, 50.3] \\
DOTA & GPT-6.1 & 74.3 & 74.4 & 39.9 & 39.6 & 40.4 & 52.1 & 39.6 & 34.4 [25.1, 42.6] \\
DOTA & Sonnet 5.5 & 41.9 & 41.9 & 23.3 & 22.6 & 24.1 & 31.1 & 21.3 & 18.6 [13.7, 25.1] \\
DOTA & Gemini 3.8 & 49.3 & 49.3 & 26.8 & 24.6 & 27.5 & 36.0 & 20.5 & 22.5 [16.7, 30.1] \\
FSC-147 & GPT-6.1 & 16.7 & 19.0 & 0.2 & $-$0.7 & 1.6 & 5.0 & $-$0.7 & 16.5 [7.6, 31.7] \\
FSC-147 & Sonnet 5.5 & 16.6 & 17.3 & 1.1 & 1.0 & 1.2 & 5.5 & 0.8 & 15.5 [10.6, 21.4] \\
FSC-147 & Gemini 3.8 & 6.0 & 6.2 & 3.9 & 4.1 & 4.2 & 4.6 & 2.0 & 2.1 [0.8, 4.0] \\
SKU-110K & GPT-6.1 & 71.8 & 71.8 & 42.0 & 39.9 & 42.2 & 46.8 & 40.7 & 29.8 [27.2, 32.7] \\
SKU-110K & Sonnet 5.5 & 45.3 & 45.3 & 34.9 & 34.8 & 35.1 & 37.0 & 34.6 & 10.3 [8.4, 12.6] \\
SKU-110K & Gemini 3.8 & 44.6 & 44.6 & 13.9 & 14.3 & 14.3 & 22.7 & 9.8 & 30.6 [26.4, 35.6] \\
\midrule
\multicolumn{2}{@{}l}{Median of 12 cells} & 45.2 & 45.2 & 15.6 & 14.6 & 18.4 & 36.5 & 13.2 & 26.1 \\
\bottomrule\end{tabular}\end{table*}

\paragraph{Selection among a model's own outputs.} Choosing the best feasible set from a model's own three calls would remove 45.2\% of set error relative to one call (median over 12 cells; Table~\ref{tab:select}), and 72\% to 74\% for GPT-6.1 on SKU-110K and DOTA. The constrained oracle differs from the unrestricted value on 25 of 720 images and has the same median. Record gates on output features, fitted with labels and evaluated by cross-fitting, reach 15.6\% (logistic) and 14.6\% (boosted trees); fitted on 10 labelled images they reach 13.2\%. The more flexible selector does not help under cross-fitting, and its in-sample value (36.5\%) is overfit. The empirical gap, median 26.1 points, combines information absent from the output features with selector and estimation shortfall (Proposition~\ref{prop:order}); our results bound it but do not apportion it. It measures the distance between the tested selectors and the best set in a fixed library of three calls, so it does not bound what adding calls or changing the library could gain. Within the operations we compared, the larger gains came from changing the information, a different view, model or readout, rather than from selecting among a model's own outputs.

\subsection{Judge: defaults that transfer}\label{sec:res:transfer}

\begin{table*}[tp]\centering\footnotesize\setlength{\tabcolsep}{3pt}
\caption{Defaults evaluated by leave-one-source-out transfer: for each source, the construction is the one that is best on the other five sources (same model only, or any construction including cross-model ones), and it is evaluated on the held-out source against a single call of the user's model. Entries are relative changes of the loss in \% (for $E$, negative is better; for F1, positive is better). \textbf{Bold}: the paired 95\% interval excludes 0 in the favourable direction; \underline{underlined}: it excludes 0 in the unfavourable direction. $^{\star}$: the user's model is the strongest single model on that source and loss; every significantly worse cross-model entry carries a $\star$. GPT, Son., Gem.: GPT-6.1, Sonnet 5.5, Gemini 3.8. Costs, recall and the chosen construction per cell are in Table~\ref{tab:transfer_full}.}\label{tab:transfer}
\begin{tabular}{@{}lrrrrrrrrrrrr@{}}\toprule
& \multicolumn{6}{c}{Same model only} & \multicolumn{6}{c}{Any construction (incl.\ cross-model)} \\\cmidrule(lr){2-7}\cmidrule(l){8-13}
& \multicolumn{3}{c}{$E$} & \multicolumn{3}{c}{F1} & \multicolumn{3}{c}{$E$} & \multicolumn{3}{c}{F1} \\\cmidrule(lr){2-4}\cmidrule(lr){5-7}\cmidrule(lr){8-10}\cmidrule(l){11-13}
Source & GPT & Son. & Gem. & GPT & Son. & Gem. & GPT & Son. & Gem. & GPT & Son. & Gem. \\\midrule
DocLayNet & $+$7$^{\star}$ & {\bfseries\boldmath $-$10} & $+$3 & \underline{$-$3}$^{\star}$ & $-$1 & {\bfseries\boldmath $+$19} & $+$4$^{\star}$ & {\bfseries\boldmath $-$36} & {\bfseries\boldmath $-$13} & $-$7$^{\star}$ & {\bfseries\boldmath $+$15} & {\bfseries\boldmath $+$75} \\
DOTA & {\bfseries\boldmath $-$22} & {\bfseries\boldmath $-$16}$^{\star}$ & {\bfseries\boldmath $-$25} & $-$1$^{\star}$ & $-$1 & $+$2 & {\bfseries\boldmath $-$30} & {\bfseries\boldmath $-$17}$^{\star}$ & {\bfseries\boldmath $-$22} & $-$6$^{\star}$ & $+$11 & {\bfseries\boldmath $+$17} \\
FSC-147 & $+$3$^{\star}$ & \underline{$+$7} & \underline{$+$5} & $-$2$^{\star}$ & \underline{$-$6} & $+$1 & \underline{$+$58}$^{\star}$ & $-$1 & $-$1 & \underline{$-$35}$^{\star}$ & $+$1 & {\bfseries\boldmath $+$68} \\
SKU-110K & {\bfseries\boldmath $-$25} & {\bfseries\boldmath $-$32} & {\bfseries\boldmath $-$10}$^{\star}$ & \underline{$-$8}$^{\star}$ & $+$1 & {\bfseries\boldmath $+$13} & {\bfseries\boldmath $-$30} & {\bfseries\boldmath $-$36} & {\bfseries\boldmath $-$8}$^{\star}$ & \underline{$-$22}$^{\star}$ & {\bfseries\boldmath $+$165} & {\bfseries\boldmath $+$21} \\
GroundCUA & $-$4 & $-$1$^{\star}$ & {\bfseries\boldmath $-$21} & \underline{$-$4} & $+$2$^{\star}$ & $+$12 & {\bfseries\boldmath $-$26} & \underline{$+$52}$^{\star}$ & {\bfseries\boldmath $-$42} & {\bfseries\boldmath $+$24} & \underline{$-$17}$^{\star}$ & {\bfseries\boldmath $+$61} \\
ScreenParse & {\bfseries\boldmath $-$3} & $-$2$^{\star}$ & $-$3 & 0 & 0$^{\star}$ & $+$2 & {\bfseries\boldmath $-$14} & $+$4$^{\star}$ & {\bfseries\boldmath $-$12} & {\bfseries\boldmath $+$6} & $-$2$^{\star}$ & {\bfseries\boldmath $+$11} \\
\midrule
Better / worse / n.s. & \multicolumn{3}{c}{9 / 2 / 7} & \multicolumn{3}{c}{2 / 4 / 12} & \multicolumn{3}{c}{12 / 2 / 4} & \multicolumn{3}{c}{10 / 3 / 5} \\
\bottomrule\end{tabular}\end{table*}

Table~\ref{tab:transfer} (full values in Table~\ref{tab:transfer_full}) evaluates construction transfer on stored reports: for each source, the construction that does best on the other five sources is applied to the stored calls of the sixth, so it is never chosen on the source it is evaluated on. Table~\ref{tab:defaults} shows where the final defaults follow these transferred constructions and where they differ. For set error, the best same-model construction on the other five sources is three-call unanimity with averaged coordinates on every evaluated source. It lowers set error significantly in 9 of the 15 over-reporting cells and raises it significantly in none; on FSC-147, the under-reporting family, it is worse in 2 of 3 cells, consistent with Proposition~\ref{thm:intersect} and Corollary~\ref{cor:nonid}. For F1, the transferred same-model construction is better in 2 cells, worse in 4 and not significant in 12, so the default is a single call. Three calls cost three times one call. From logged token usage at list prices, unanimity costs \$0.029 to \$0.21 per image for GPT-6.1 across the five over-reporting sources, \$0.032 to \$0.13 for Sonnet 5.5 and \$0.011 to \$0.058 for Gemini 3.8.

\begin{table*}[tp]
\centering\footnotesize\setlength{\tabcolsep}{5pt}
\caption{Single-call configuration chosen on the other three detection sources and applied to the fourth. F1 at the primary scoring of Table~\ref{tab:contract} (boxes: same class, IoU $\ge0.75$; FSC-147: point tolerance), in \%. Reference: list interface at the lowest setting. Change: chosen minus reference, in F1 points, with a 95\% paired bootstrap interval over images. \textbf{Bold}: the paired 95\% interval excludes 0 in the favourable direction; \underline{underlined}: it excludes 0 in the unfavourable direction. Macro F1, the mean over the 12 cells: 42.1 (reference) and 49.1 (chosen); the best configuration of each cell chosen with its own labels gives 51.7. The shipped configuration (Table~\ref{tab:defaults}) is chosen by the same rule on all four sources (Table~\ref{tab:config_menu}).}\label{tab:config}
\begin{tabular}{@{}llllrr@{}}\toprule
& & \multicolumn{2}{c}{Chosen} & & \\\cmidrule(lr){3-4}
Source & Model & Interface & Setting & Change & 95\% CI \\\midrule
DocLayNet & GPT-6.1 & regions & lowest & $+$1.5 & [$-$2.6, 6.0] \\
DocLayNet & Sonnet 5.5 & regions & default & \underline{$-$6.5} & [$-$10.4, $-$3.1] \\
DocLayNet & Gemini 3.8 & list & default & {\bfseries\boldmath $+$14.4} & [6.6, 22.5] \\
DOTA & GPT-6.1 & regions & lowest & $-$1.7 & [$-$5.3, 1.7] \\
DOTA & Sonnet 5.5 & regions & default & $+$3.1 & [$-$1.8, 7.4] \\
DOTA & Gemini 3.8 & regions & default & $+$4.9 & [$-$0.5, 10.9] \\
FSC-147 & GPT-6.1 & regions & lowest & {\bfseries\boldmath $+$9.6} & [2.2, 17.6] \\
FSC-147 & Sonnet 5.5 & regions & default & {\bfseries\boldmath $+$19.0} & [12.3, 26.0] \\
FSC-147 & Gemini 3.8 & regions & default & {\bfseries\boldmath $+$22.6} & [14.3, 31.4] \\
SKU-110K & GPT-6.1 & regions & lowest & $+$0.4 & [$-$0.5, 1.4] \\
SKU-110K & Sonnet 5.5 & list & default & {\bfseries\boldmath $+$13.2} & [8.9, 17.8] \\
SKU-110K & Gemini 3.8 & list & default & $+$2.9 & [$-$2.2, 8.1] \\
\bottomrule\end{tabular}
\end{table*}

\paragraph{Configurations that transfer.} The same logic applies to the configuration of a single call: the reporting interface (one list for the whole image, or one row for each region of a fixed registry) and the reasoning setting. For each model, we chose among the four combinations on three detection sources and applied the choice to the fourth, so the labels of the evaluated source are used only to score it (Table~\ref{tab:config}). For F1, the chosen configuration is better than the list interface at the lowest setting in 5 of 12 cells and worse in 1 (Sonnet 5.5 on DocLayNet, $-6.5$ points); macro F1 rises from 42.1 to 49.1, against 51.7 for the best configuration of each cell chosen with its own labels, and the chosen calls cost 1.44 times as much. For set error, the same rule is better in 2 cells, worse in 1 and not significant in 9 (median change 0\%), so the set-error defaults stay as in Table~\ref{tab:defaults}. One configuration shared by the three models transfers worse (F1: 5 cells better, 3 worse). The configuration that ships is chosen by the same rule on all four sources (GPT-6.1: regions, lowest; Sonnet 5.5: regions, default; Gemini 3.8: list, default; Table~\ref{tab:config_menu}).

When cross-model constructions are allowed, the transferred default improves set error in 12 of 18 cells and F1 in 10 of 18. It is better in 20 of the 24 cells in which the user's model is not the strongest single model for that source and loss, and not significantly different in the other 4. All 5 cells in which it is significantly worse are cells where the user's model is the strongest; for example, a three-model majority raises Sonnet 5.5's set error on GroundCUA by 52\% relative to one call, and a two-model intersection raises GPT-6.1's on FSC-147 by 58\%. Among constructions with averaged coordinates, against the best same-model construction chosen with all labels (a reference, not a deployable rule), the best cross-model construction is significantly better only on DOTA ($-5.8\%$) and significantly worse on GroundCUA ($+26.0\%$). When the user does not know which model is strongest, three-model unanimity is significantly better than three-call unanimity of a randomly chosen model on 4 of 5 over-reporting sources ($-4\%$ to $-11\%$). Voting a strong model together with weaker ones is not a safe default: a three-model majority is 52\% worse than GPT-6.1's own three-call majority on FSC-147 and 59\% worse than Sonnet 5.5's on GroundCUA, an unzoomed three-model click vote on the 260 click items of the GUI experiment is 3.5 points below GPT-6.1 alone, and the per-type median of three models' dense counts is 0.60 worse than Sonnet 5.5 alone. On aerial and shelf detection the same vote helps instead: it is 9.5\% better than Sonnet 5.5's three-call majority on DOTA and 4.7\% better than Gemini 3.8's on SKU-110K; on DocLayNet and ScreenParse it is not significantly different. Cross-model confirmation is therefore a menu option chosen with a few labels or a family-level ranking, while the cross-model alarm is a default whenever a second model is available.

\section{Discussion}\label{sec:discussion}

\paragraph{What a practitioner can take away.} Table~\ref{tab:recipe} lists each component of the method with when to use it, its measured effect, where it does not help, and its cost; Table~\ref{tab:defaults} gives the shipped settings and Appendix~\ref{app:protocol} the prompts. Three quantities decide most choices at run time: the degeneracy flags decide which count to read; agreement between two calls of one model measures stability, not correctness, so high agreement means another such call adds little; and the family default, or $p_1$ estimated on a few labelled images, decides whether to delete unconfirmed records.

\begin{table*}[tp]
\iflatexml\else\centering\fi\footnotesize\setlength{\tabcolsep}{3.5pt}\renewcommand{\arraystretch}{1.15}
\caption{FSR at a glance: what each component is for, its measured effect, where it gives no significant gain or harms, and its cost. Effects are relative changes of the stated loss against one call of the same model unless stated; ``n.s.'': not significant. The last row lists remedies with small or no significant gains.}\label{tab:recipe}
\begin{tabular}{@{}>{\raggedright\arraybackslash}p{2.5cm}>{\raggedright\arraybackslash}p{2.6cm}>{\raggedright\arraybackslash}p{3.8cm}>{\raggedright\arraybackslash}p{4.0cm}>{\raggedright\arraybackslash}p{1.9cm}@{}}\toprule
Component & Use when & Measured effect & No significant gain, or harm, when & Cost \\\midrule
List first, read the list count; stated count only if the list is flagged & counting, above all several classes in dense scenes & per-type count error $-85.2\%$ (GPT-6.1), $-98.8\%$ (Sonnet 5.5); same response, readout only: $-71\%$ to $-84\%$ (\S\ref{sec:res:readout}) & one-class photographs, and aerial or document images whose lists miss objects: small, n.s.\ for GPT-6.1 and Sonnet 5.5 (\S\ref{sec:res:readout}) & one call \\
Allowed names in the prompt, aliases mapped in code & the model uses names outside the vocabulary & Sonnet 5.5's ``diamond'': error 5.42 to 0.42 on the same lists (Table~\ref{tab:readout}) & names already in the vocabulary & no extra calls \\
Degeneracy flags; Gemini 3.8 at its default setting for counts & long lists that may break & flagged lists 90\% to 12\%, count error $-53.6\%$ at the same cap (Table~\ref{tab:setting}) & Gemini 3.8 at its lowest setting, even with the fallback: $+15.6\%$, n.s. & more reasoning tokens \\
Two-stage zoom & clicks, small targets & miss rate $-65\%$ (Gemini 3.8), $-41\%$ (Sonnet 5.5) (\S\ref{sec:res:view}) & GPT-6.1: n.s. & about 2 calls \\
Views from a planner, merged; each view confirms only records it contains & small objects; views from a planner or crops & four crops (6\% overlap) confirming a whole-image report: set error up to $-45.5\%$, better in 4 of 6 cells (\S\ref{sec:res:view}) & where unconfirmed records are mostly true ($p_1>1/2$): worse in 2 of 6 cells; tiles without a one-object overlap lose large objects & one call per view \\
Three-call unanimity, coordinates averaged & set error on over-reporting families (detection, layout, screens) & significantly better in 9 of 15 cells, significantly worse in none (\S\ref{sec:res:transfer}) & F1 (2 better, 4 worse, 12 n.s.); under-reporting dense points (worse in 2 of 3) & 3 calls \\
Per-model single-call configuration for F1 & F1 on detection-type sources & macro F1 42.1 to 49.1; significantly better in 5 of 12 cells, worse in 1 (\S\ref{sec:res:transfer}) & set error: 2 better, 1 worse, 9 n.s. & $1.44\times$ the cost of the same model's list call at the lowest setting \\
Cross-model alarm & deciding which items to check & catches 61\% of GPT-6.1's misses against 17\% for a same-model call; 68\% to 78\% after zoom (\S\ref{sec:res:persist}) & flags more items; disagreement does not say which answer is right, and agreement does not certify & one call of a cheaper model \\
Three-model GUI-grounding tier (Algorithm~\ref{alg:clicks}) & higher click accuracy is worth up to 15 planned calls per item & miss rate 13.67\% to 9.33\%; 3.00 points better than a plain vote over the same points (\S\ref{sec:res:choose}) & against crop-first: n.s.; a fourth source: n.s.; the comparison step alone: n.s. & \$0.0273 per item, $3.2\times$ one zoom \\
Remedies with small or no significant gains & \multicolumn{4}{>{\raggedright\arraybackslash}p{12.7cm}}{a three-click vote or zoom gated on disagreement (miss-rate reductions of at most 13\%); grids, ticks and cell labels (localization changes $\le0.23$ px); ``do not miss any'' wording; votes over repeated calls of one model (error reductions of at most 23\% for counts and 4.5\% for clicks); same-view self-review on screens; an equal-weight vote of all models when one is strongest (\S\ref{sec:res:view}, \S\ref{sec:res:common}, \S\ref{sec:res:transfer})} \\\bottomrule
\end{tabular}
\end{table*}

\paragraph{Why the gains come from changing information.} The theory and the measurements point the same way. Object-level persistence means that repeated calls of one model share most of their errors. In our comparisons, votes over the same outputs gained at most 23\% for counts and 4.5\% for clicks, and even a perfect selector over three calls can only choose among the records those calls contain. The largest gains we measured came from operations that change what is available: reading a different field of the same response, looking at a different view, or asking a model with a different model-specific component. Where those operations fail, the failures have the predicted form: deletion hurts when $p_1>1/2$, tiling can lose objects that no tile contains whole, and a partner hurts when its additional loss of true matches outweighs its additional rejection of false records, with the effect of averaged coordinates counted separately.

\paragraph{Where click accuracy can still come from.} Of the tier's 56 remaining misses on the 600 click items, 43 have no correct candidate and 13 are lost in representation or choice (Proposition~\ref{prop:three}). A better rule over the same candidates can act only on the 13. The 43 need a source with a high rescue rate $\phi$ on exactly those items, and its gain must survive representation and choice: of the crop-first source's 7 rescues, 3 did (Section~\ref{sec:res:choose}). On items where the models agree, any further action must meet the break-rate condition of Proposition~\ref{prop:switch}. These two measurable quantities, the rescue rate on the items no candidate covers and the break rate on agreeing items, decide whether a further source or action raises accuracy, and both are measured from stored calls on labelled items.

\paragraph{What future models should change.} The measurements point to properties that model builders can target directly. A model's stated summary should be computed from its own enumeration: GPT-6.1 and Sonnet 5.5 already list almost exactly the right number of objects, and their stated counts lose that information. Long lists should not degrade with position; Gemini 3.8 at its lowest setting drops class fields more often when consecutive records share a class; an order that interleaves classes lowers its count error and grouping by class raises it (Section~\ref{sec:res:degen}). Class vocabularies given in the prompt should be respected, since Sonnet 5.5 still wrote ``diamond'' 258 times when the format named the allowed classes. Finally, errors that persist across calls limit what a fixed budget of resampling can repair; diversity across calls, or calibrated uncertainty at the object level, could make the cheaper same-model alarm as useful as a second model.

\paragraph{Scope of the results.} The theory's guarantees hold under their stated conditions. The identities of Section~\ref{sec:theory:identities} hold for any outputs and any fixed choice of maximum matching. Proposition~\ref{prop:partner} assumes deletion-only confirmation, no match takeover, and no ignore region or none of the base report's false records inside one; outside these conditions the exact identity of Proposition~\ref{prop:identity} replaces it. The disagreement floor is a bound for point tasks and for boxes at IoU thresholds above 0.5; the click alarm is empirical and is evaluated by risk and acceptance coverage. Proposition~\ref{prop:nocert} concerns model classes that admit scenes with identical or nearly identical output laws; when an object is reported with probability at most $\vartheta$ per call, $K$ calls separate the two scenes by at most $K\vartheta$ in total variation. The readout repair applies where a readout gap exists: on our dense synthetic scenes the gap is most of the error, while on aerial and document images it is a small part of it for GPT-6.1 and Sonnet 5.5. Section~\ref{sec:experiments} states how the reasoning setting and the output cap were separated. All results are for the model versions listed in Appendix~\ref{app:registry}, at a total cost of about \$758; we release every response and the analysis code, so the reported identities and tables can be reproduced without new model calls, while evaluating a new model version requires its own responses, analysed the same way.

\paragraph{Broader impact.} More accurate and auditable visual reports can make inventory, document and screen-automation systems more reliable, and an explicit status and alarm make failures visible instead of silent. The same techniques could make automated agents act with more confidence; our pipeline always returns a status and never claims certification without labels.

\section{Conclusion}\label{sec:conclusion}

Reporting is part of the visual task, not merely its formatting. When vision-language models report sets of objects, part of their error is in the report rather than in the seeing. FSR treats reporting as a chain from responses to records, sets and readouts, which makes that error analysable. The chain gives exact identities for what each operation does, a behavioural model of persistent errors, and a proof that outputs alone cannot certify correctness when two scenes produce the same outputs. It also separates three problems. Repair works where the information is already present: read the count from the model's list of records, not from the numbers it states, unless the list has degenerated; delete unconfirmed records only below the cost threshold; change the report structure, its decoding (declared aliases applied in code), the view or the model, rather than restating the same request more insistently. Error detection benefits most from a different source of evidence; on our click tasks a second model provided it better than a second call. For one click, the error splits exactly into missing candidates, lost representatives and wrong choices, which tells where to spend calls: FSR's GUI-grounding tier, which adds two other models' candidates and compares them only where they disagree, lowered the miss rate on 600 ScreenSpot-Pro items to 9.33\%, against 13.67\% for one model's zoom and 12.33\% for a plain vote over the same three models; against a crop-first baseline the difference was not significant. Judging between reports needs labels or a configuration transferred from other data; three-call unanimity, chosen on five sources, improved set error on most over-reporting cells of the sixth and did not significantly worsen it in any of them, while for F1 a single call remained the default, with an interface and setting per model on the detection sources. Every number in this paper can be regenerated from the released responses, with the DOTA, SKU-110K and FSC-147 labels rebuilt from their official downloads.

\bibliography{refs}

\appendix
\onecolumn
\counterwithin{table}{section}
\section{Proofs and Supporting Results}\label{app:proofs}

Throughout, $E(S)=c(S)+n-2r(S)$ and $E_w(S)=\cfp[c(S)-r(S)]+\cfn[n-r(S)]$, with $c$ and $r$ computed from one fixed choice of maximum matching.

\begin{lemma}[Matching size]\label{lem:rank}
Fix the targets, a ground set of candidate records and the compatibility graph between them. Then $S\mapsto r(S)$ is the rank function of a transversal matroid \citep{edmonds1965transversals}: it never decreases when records are added, increases by 0 or 1 when one record is added, and is submodular (adding a record helps a smaller set at least as much as a larger one).
\end{lemma}

\paragraph{Proof of Lemma~\ref{lem:rank}.} Consider the bipartite graph between records and targets with edges given by $\rho_\tau$. The subsets of records that can be matched simultaneously form the independent sets of a transversal matroid \citep{edmonds1965transversals}, and the maximum matching size of $S$ is the rank of $S$ in that matroid. Matroid rank functions are monotone, satisfy $r(S\cup\{z\})-r(S)\in\{0,1\}$, and are submodular. \qed

\paragraph{Proposition~\ref{prop:identity}.} Substituting the definition, $E_w(A)-E_w(S)=\cfp[(c(A)-c(S))-(r(A)-r(S))]-\cfn[r(A)-r(S)]=\cfp\Delta c-(\cfp+\cfn)\Delta r$. Without an ignore region every record is scored, so $\Delta c=|A|-|S|$. \qed

\paragraph{Proposition~\ref{thm:intersect}.} Deleting $A\setminus S$ lowers $E_w$ iff $E_w(A)-E_w(S)>0$, which by Proposition~\ref{prop:identity} is $\cfp\Delta c>(\cfp+\cfn)\Delta r$. If $\Delta c>0$, divide by $(\cfp+\cfn)\Delta c$. If $\Delta c=0$, deletion lowers $E_w$ iff $\Delta r<0$, which cannot happen by monotonicity (Lemma~\ref{lem:rank}), so deletion of no scored record never helps. Addition of $B'$ to $A$ is the same identity with the roles of $A\cup B'$ and $A$: it lowers $E_w$ iff $(\cfp+\cfn)[r(A\cup B')-r(A)]>\cfp[c(A\cup B')-c(A)]$. For the population rule, the expected change is $\cfp\mathbb{E}[\Delta c]-(\cfp+\cfn)\mathbb{E}[\Delta r]$, whose sign is that of $\cfp/(\cfp+\cfn)-p_1$ when $\mathbb{E}[\Delta c]>0$. \qed

\begin{corollary}[F1 under deletion]\label{cor:f1}
If $n+c(A)>0$ and $n+c(S)>0$, then
\[
\mathrm{F1}(S)-\mathrm{F1}(A)=\frac{2\{r(A)\Delta c-[n+c(A)]\Delta r\}}{[n+c(A)][n+c(S)]},
\]
so for $\Delta c>0$ deletion raises F1 on that image iff $\Delta r/\Delta c<\mathrm{F1}(A)/2$. If $n=0$, $c(A)>0$ and every scored record is deleted, F1 rises from 0 to 1.
\end{corollary}
In 22 of our 54 base/partner pairs, the pooled marginal matching rate lies between half the base call's mean F1 and $1/2$; since the condition is per image, this count indicates a direction only. A mean F1 over images changes by the mean of the per-image differences, not by comparing a pooled rate with an average threshold.

\paragraph{Proof of Corollary~\ref{cor:f1}.} Write $r=r(A)$, $c=c(A)$, so that $n+c(S)=n+c-\Delta c$. Then $\mathrm{F1}(S)-\mathrm{F1}(A)=\frac{2(r-\Delta r)}{n+c-\Delta c}-\frac{2r}{n+c}=\frac{2[(r-\Delta r)(n+c)-r(n+c-\Delta c)]}{(n+c)(n+c-\Delta c)}=\frac{2[r\,\Delta c-(n+c)\Delta r]}{(n+c)(n+c-\Delta c)}$. Both denominators are positive, so for $\Delta c>0$ the sign is that of $r/(n+c)-\Delta r/\Delta c=\mathrm{F1}(A)/2-\Delta r/\Delta c$. If $n=0$, then $r=0$ and $\mathrm{F1}(A)=0$ whenever $c(A)>0$; deleting every scored record gives $n=c=0$ and $\mathrm{F1}=1$ by convention. \qed

\paragraph{Proposition~\ref{prop:geometry}.} The decomposition is a telescoping sum. Applying $E=c+n-2r$ to $S_0$ and $S_f$ gives the geometry term $[c(S_0)-c(S_f)]-2[r(S_0)-r(S_f)]$. Convexity of the squared distance gives $\|\bar z-y\|^2\le\frac1m\sum_i\|z_i-y\|^2$ for any target $y$, so averaging does not increase mean squared error to any fixed target. A threshold matching is not monotone in squared error, however. If one member is within tolerance of $y$ and the other is far from every target, their mean can leave the tolerance ball of $y$ and a match is lost; conversely, two members just outside the tolerance on opposite sides of $y$ have a mean inside it, and a match is gained. Hence the geometry term can have either sign. \qed

\paragraph{Proposition~\ref{prop:readout}.} Both errors are evaluated against the same $n$ for the same response, so their difference depends only on which field is read. By the triangle inequality, $\|d-\mathrm{lc}(Z)\|_1\le\|d-n\|_1+\|n-\mathrm{lc}(Z)\|_1\le2\max\{\|d-n\|_1,\|\mathrm{lc}(Z)-n\|_1\}$. For the second inequality, the reverse triangle inequality gives $\big||d_c-n_c|-|\mathrm{lc}(Z)_c-n_c|\big|\le|d_c-\mathrm{lc}(Z)_c|$ for every class $c$; summing over classes and using $|\sum_c x_c|\le\sum_c|x_c|$ gives the bound. \qed

\paragraph{Diminishing coverage.} Let $U_K=Z_1\cup\dots\cup Z_K$ and $\delta_K=\mathbb{E}[r(U_K)-r(U_{K-1})]$. By submodularity, $r(U_K\cup Z_{K+1})-r(U_K)\le r(U_{K-1}\cup Z_{K+1})-r(U_{K-1})$. Taking expectations and using exchangeability of $Z_K$ and $Z_{K+1}$ given the image, the right side has expectation $\delta_K$, so $\delta_{K+1}\le\delta_K$. \qed

\paragraph{Proposition~\ref{prop:partner}.} Under deletion-only confirmation, the records of $A$ not confirmed by $B$ are deleted. In expectation, $n_+(1-h_+^B)$ true and $n_-(1-h_-^B)$ false records are deleted. With no takeover, $\Delta r$ is the number of deleted true records, and with all false records scored, $\Delta c$ is the number of deleted records. By Proposition~\ref{prop:identity}, $\mathbb{E}[E(A)-E(A\cap B)]=n_+(1-h_+^B)+n_-(1-h_-^B)-2n_+(1-h_+^B)=n_-(1-h_-^B)-n_+(1-h_+^B)$. The comparison follows from $G(B)-G(A')=n_-(h_-^{A'}-h_-^{B})-n_+(h_+^{A'}-h_+^{B})$. For the examples, $10(0.15)-100(0.05)=-3.5$ and $10(1)-100(0.5)=-40$. \qed

\paragraph{Proposition~\ref{prop:alarm}.} Let $M_A$ and $M_B$ be maximum matchings of $A$ and $B$ to $Y$ at tolerance $\theta$. At least $r(A)+r(B)-n$ targets are matched in both. For each such target $y$, its partners $a\in A$ and $b\in B$ satisfy $\|a-b\|\le\|a-y\|+\|y-b\|\le2\theta$, and distinct targets give disjoint pairs. Hence $m_{2\theta}(A,B)\ge r(A)+r(B)-n$. With no ignore region, $E(A)+E(B)=|A|+|B|+2n-2r(A)-2r(B)\ge|A|+|B|-2m_{2\theta}(A,B)=U_{2\theta}(A,B)$. The bound on the maximum follows. For boxes, if $\mathrm{IoU}(a,y)\ge t$ and $\mathrm{IoU}(b,y)\ge t$, then $1-\mathrm{IoU}$ is a metric (the Jaccard distance), so $\mathrm{IoU}(a,b)\ge2t-1$, and the same argument applies. At $t=0.5$ the overlap threshold becomes 0, so the bound for boxes keeps differences in the number of records per class but loses all geometric information. \qed

\paragraph{Proposition~\ref{prop:order}.} Every rule $g(\Phi)$ takes values in $\SA$, so pointwise $\ell(g(\Phi),Y)\ge\min_{S\in\SA}\ell(S,Y)$; taking expectations and the infimum over $g$ gives $R_{\mathrm{or}}\le R^*_\Phi$. A fitted rule is one such $g$, so $R^*_\Phi\le R_{\hat g}$. The decomposition is an identity. \qed

\begin{proposition}[Constrained oracle]\label{prop:oracle}
For $\SA$ with at most one representative per cluster, including the option of choosing none, $\min_{S\in\SA}E(S)=n-|M|$, where $M$ is a maximum matching in the bipartite graph between clusters and targets that has an edge whenever some $a\in A(Q)$ is compatible with the target. It is attained by selecting one compatible representative for each matched cluster and nothing for the others.
\end{proposition}
Unrestricted matching over all records can exceed $|M|$ when two members of one cluster match two targets; that value is not attainable within $\SA$ (Table~\ref{tab:select} reports both).

\paragraph{Proof of Proposition~\ref{prop:oracle}.} Any $S\in\SA$ contains at most one record per cluster, so a maximum matching of $S$ to targets induces a cluster-target matching of size $r(S)$ in the stated graph; hence $r(S)\le|M|$. Records unmatched in the chosen maximum matching and lying outside $\Omega$ add to $c$, so $c(S)\ge r(S)$ with or without ignore regions, and $E(S)=c(S)+n-2r(S)\ge n-r(S)\ge n-|M|$. The construction that selects, for each cluster matched in $M$, a representative compatible with its target, and nothing else, has $r=c=|M|$ and attains the bound. \qed

\paragraph{Proposition~\ref{prop:three}.} The output is a representative. If no candidate hits, the output misses (event of $F$). Otherwise, if no representative hits, the output misses (event of $L$). Otherwise some representative hits, and the output misses exactly in the event of $W$. The three events are disjoint and their union is the event that the output misses.

\paragraph{Proposition~\ref{prop:coverage}.} Every candidate of the enlarged menu misses iff every earlier candidate misses and the added one misses, so the probability is $F\,\Pr(\text{added misses}\mid\text{all earlier miss})=F(1-\phi)$ when $F>0$, and $0$ when $F=0$. Under conditional independence given $q$, $\Pr(\text{all miss}\mid q)=\prod_{j=1}^{k}(1-q_j)$; taking expectations gives $F$.

\paragraph{Proposition~\ref{prop:switch}.} The change in the miss indicator is $\mathbb{1}_T\,(\mathbb{1}[b\text{ misses}]-\mathbb{1}[g\text{ misses}])$, which is $+1$ when $b$ misses and $g$ hits, $-1$ when $b$ hits and $g$ misses, and $0$ otherwise. Its expectation is $\Pr(T)[\Pr(b\text{ misses},g\text{ hits}\mid T)-\Pr(b\text{ hits},g\text{ misses}\mid T)]=\Pr(T)[e_Tr_{\mathrm{fix}}-(1-e_T)r_{\mathrm{brk}}]$. If $\mathbb{1}_T$ and $\mathbb{1}[b\text{ misses}]$ are independent given $U$ and $T\subseteq U$, then $\Pr(b\text{ misses}\mid T)=\Pr(b\text{ misses}\mid U,T)=\Pr(b\text{ misses}\mid U)$. The sample size for choosing a scorer follows from Hoeffding's inequality for each scorer's empirical miss rate with a union bound over the $N_{\mathrm{s}}$ candidate scorers.

\paragraph{Proposition~\ref{prop:nocert}.} For any acceptance rule $a(H)\in[0,1]$, the acceptance probability in world $j$ is $\int a\,dP_j$. If $P_0=P_1$ the two are equal, so acceptance at least $1-\beta$ in world 0 forces acceptance at least $1-\beta>\alpha$ in world 1 when $\alpha+\beta<1$. In general, $|\int a\,dP_0-\int a\,dP_1|\le\mathrm{TV}(P_0,P_1)$ for $0\le a\le1$. For the hidden-target construction, couple the two worlds so that the output histories coincide unless the added target is reported in at least one of the $K$ calls, an event of probability at most $K\vartheta$ by the union bound, since the per-call bound $\vartheta$ holds given the earlier history; non-interference means that, conditional on the target not being reported, the rest of the output has the same law in both worlds. The coupling inequality gives $\mathrm{TV}(P_0,P_1)\le K\vartheta$. \qed

\paragraph{Corollary~\ref{cor:nonid}.} By assumption there is an admissible world in which the report over-counts, $|S|>n_0$. Add $m>|S|-n_0$ targets that no interface reports and that do not change the rest of the output ($\vartheta=0$); by closure the result is admissible. The output law is unchanged and $|S|<n_0+m$, so the sign of the count gap differs between two worlds with the same output law. \qed

\subsection{Explanatory parametrization and views}\label{app:param}

To describe how propensities vary with the model and the view when the rest of the interface is held fixed, that is, for calls that share the prompt, schema, setting and cap, we use the explanatory parametrization
\[
\mathrm{logit}\,q_{o,\eta}=a_o+b_{o,m(\eta)}+\gamma_{v(\eta)}(o),\]
\[\textstyle\sum_m b_{o,m}=0,\qquad \gamma_{v_0}(o)=0,
\]
where $v_0$ is the whole image, $a_o$ is common to all models (how hard the object is), $b_{o,m}$ is specific to model $m$, and $\gamma_v(o)$ is the change from the whole image to view $v$, which depends on the apparent size of $o$ in $v$. All three terms may differ between prompts, schemas, settings and caps; a comparison that changes one of these compares two such parametrizations, not two terms of one. The form is related to item-response and labeler-expertise models, in which an item's difficulty and a rater's ability combine on the logit scale \citep{whitehill2009glad} (there as a product, here additively, as in Rasch-type models); here the items are objects and the raters are models and views. The two constraints are conventions that make the parts well defined: centering separates the shared part from the model-specific one, and the reference view separates both from the view effect. We do not estimate the parts; the parametrization motivates the predictions of Section~\ref{sec:theory:behaviour}, each of which is tested directly.

A view can raise how often objects are reported while the misses of the two calls stay as correlated, or become more so. For a true object drawn at random from the image, let $A$ and $A'$ indicate that two calls with the same interface report it and $B$ that a call on another view does, with means $\mu_A=\mu_{A'}$ and $\mu_B$, the calls being conditionally independent given the propensities. Then
\[
\begin{aligned}
&P(A=0,B=1)-P(A=0,A'=1)\\
&\quad=(1-\mu_A)(\mu_B-\mu_A)\\
&\qquad+\mathrm{Cov}(A,A')-\mathrm{Cov}(A,B):
\end{aligned}
\]
the first term is the gain in how often objects are reported, the second the gain in complementarity. For example, with two equally likely kinds of object reported with probability 0.2 and 0.4 on the whole image, a view that adds $\log 2$ to every logit gives $1/3$ and $4/7$: the mean rises from 0.30 to 0.45 and the covariance with the first call rises from 0.0100 to 0.0119, yet the probability that the first call misses an object and the second reports it is 0.305 for the new view against 0.200 for a repeated call: the gain in how often objects are reported (0.107) more than offsets a small loss of complementarity (0.002).

\section{Full Tables}\label{app:tables}

All tables in this appendix are written by \path{paper/make_tables.py} and \path{stageB/analysis/accfirst/make_accuracy_tables.py} from the outputs of \path{stageB/analysis/run_all.sh}, which recomputes them from saved responses and labels without model calls. The same numbers are available in machine-readable form in the JSON files of \path{stageB/analysis/out/}. The click and configuration tables (Tables~\ref{tab:choose}, \ref{tab:config} and \ref{tab:config_menu}) are written by \path{stageB/analysis/accfirst/make_accuracy_tables.py}.

\begin{table}[H]\centering\footnotesize\setlength{\tabcolsep}{3pt}
\caption{Same-response readout across datasets (count-first prompts, lowest setting, one call; data of Figure~\ref{fig:density}). Count error of the stated count and of the list count in the same response, without and with the declared alias map (lower is better; total error for one-class datasets, per-type error otherwise). Change: relative change from the stated count to the list count, with its paired image-bootstrap 95\% interval; this is a separate run from Table~\ref{tab:readout}, so shared rows can differ in the last digit of an interval, and the two list-count columns use separate draws. \textbf{Bold}: the paired 95\% interval excludes 0 in the favourable direction; \underline{underlined}: it excludes 0 in the unfavourable direction. Under each dataset: the error measure and the mean number of objects per image.}\label{tab:density}
\begin{tabular}{@{}llrrrrrrrr@{}}\toprule
& & & & \multicolumn{3}{c}{Raw list count} & \multicolumn{3}{c}{Mapped list count} \\\cmidrule(lr){5-7}\cmidrule(l){8-10}
Dataset & Model & $n$ & Stated & Error & Change (\%) & 95\% CI & Error & Change (\%) & 95\% CI \\\midrule
\begin{tabular}[t]{@{}l@{}}PixMo-Count\\total, 6.0 obj.\end{tabular} & GPT-6.1 & 100 & 0.16 & 0.15 & $-$6.3 & [$-$23.1, 0.0] & \multicolumn{3}{c}{not applicable (one class)} \\
 & Sonnet 5.5 & 100 & 0.17 & 0.14 & $-$17.6 & [$-$38.5, 0.0] & \multicolumn{3}{c}{not applicable (one class)} \\
 & Gemini 3.8 & 100 & 0.33 & 0.27 & $-$18.2 & [$-$37.5, 0.0] & \multicolumn{3}{c}{not applicable (one class)} \\
\midrule
\begin{tabular}[t]{@{}l@{}}FSC-147\\total, 43.8 obj.\end{tabular} & GPT-6.1 & 100 & 2.73 & 2.70 & $-$1.1 & [$-$8.5, 8.8] & \multicolumn{3}{c}{not applicable (one class)} \\
 & Sonnet 5.5 & 100 & 9.24 & 8.85 & $-$4.2 & [$-$11.9, 2.1] & \multicolumn{3}{c}{not applicable (one class)} \\
 & Gemini 3.8 & 100 & 6.87 & 3.93 & {\bfseries\boldmath $-$42.8} & [$-$65.7, $-$17.4] & \multicolumn{3}{c}{not applicable (one class)} \\
\midrule
\begin{tabular}[t]{@{}l@{}}SCP\\per type, 47.6 obj.\end{tabular} & GPT-6.1 & 100 & 0.59 & 0.33 & {\bfseries\boldmath $-$44.1} & [$-$60.8, $-$18.6] & 0.33 & {\bfseries\boldmath $-$44.1} & [$-$61.1, $-$19.0] \\
 & Sonnet 5.5 & 100 & 4.52 & 0.07 & {\bfseries\boldmath $-$98.5} & [$-$99.6, $-$96.9] & 0.07 & {\bfseries\boldmath $-$98.5} & [$-$99.6, $-$96.9] \\
 & Gemini 3.8 & 100 & 5.92 & 17.25 & \underline{$+$191.4} & [145.8, 239.3] & 17.25 & \underline{$+$191.4} & [146.5, 234.5] \\
\midrule
\begin{tabular}[t]{@{}l@{}}SCP-dense\\per type, 164.3 obj.\end{tabular} & GPT-6.1 & 50 & 10.84 & 2.64 & {\bfseries\boldmath $-$75.6} & [$-$83.0, $-$65.9] & 2.64 & {\bfseries\boldmath $-$75.6} & [$-$83.0, $-$65.7] \\
 & Sonnet 5.5 & 50 & 35.02 & 7.80 & {\bfseries\boldmath $-$77.7} & [$-$91.5, $-$58.4] & 5.42 & {\bfseries\boldmath $-$84.5} & [$-$97.6, $-$65.9] \\
 & Gemini 3.8 & 50 & 45.68 & 129.04 & \underline{$+$182.5} & [163.4, 202.0] & 129.04 & \underline{$+$182.5} & [163.2, 202.4] \\
\bottomrule\end{tabular}\end{table}

{\footnotesize\setlength{\tabcolsep}{4pt}\begin{longtable}{@{}lllrrrrrr@{}}
\caption{Selection and geometry terms for every base/partner pair (60 images per source, one call each; data of Table~\ref{tab:intersect}). $\Delta c$: deleted scored records per image; $\Delta r$: matches lost per image; $p_1=\Delta r/\Delta c$, the marginal matching rate of the deleted records. Selection, geometry and total are per-image reductions of $E$ (positive is better); selection plus geometry equals the total on every row (Propositions~\ref{prop:identity} and \ref{prop:geometry}).}\ltlabel{tab:intersect_full}\\\toprule
Source & Base & Partner & $\Delta c$ & $\Delta r$ & $p_1$ & Selection & Geometry & Total \\\midrule\endfirsthead
\multicolumn{9}{@{}l}{\emph{Table~\ref{tab:intersect_full}, continued}}\\\toprule Source & Base & Partner & $\Delta c$ & $\Delta r$ & $p_1$ & Selection & Geometry & Total \\\midrule\endhead
\midrule\multicolumn{9}{r}{\emph{continued on the next page}}\\\endfoot
\bottomrule\endlastfoot
DocLayNet & GPT-6.1 & same model & 0.92 & 0.48 & 0.527 & $-$0.05 & $-$0.30 & $-$0.35 \\
DocLayNet & GPT-6.1 & Sonnet 5.5 & 2.63 & 0.92 & 0.348 & 0.80 & $-$1.10 & $-$0.30 \\
DocLayNet & GPT-6.1 & Gemini 3.8 & 8.08 & 5.15 & 0.637 & $-$2.22 & 0.43 & $-$1.78 \\
DocLayNet & Sonnet 5.5 & same model & 1.33 & 0.32 & 0.237 & 0.70 & 0.10 & 0.80 \\
DocLayNet & Sonnet 5.5 & GPT-6.1 & 3.03 & 0.45 & 0.148 & 2.13 & 2.77 & 4.90 \\
DocLayNet & Sonnet 5.5 & Gemini 3.8 & 8.30 & 3.97 & 0.478 & 0.37 & 3.07 & 3.43 \\
DocLayNet & Gemini 3.8 & same model & 3.48 & 2.03 & 0.584 & $-$0.58 & 0.30 & $-$0.28 \\
DocLayNet & Gemini 3.8 & GPT-6.1 & 2.12 & 0.55 & 0.260 & 1.02 & 0.47 & 1.48 \\
DocLayNet & Gemini 3.8 & Sonnet 5.5 & 1.93 & 0.22 & 0.112 & 1.50 & 0.00 & 1.50 \\
\midrule
DOTA & GPT-6.1 & same model & 7.07 & 0.43 & 0.061 & 6.20 & 0.50 & 6.70 \\
DOTA & GPT-6.1 & Sonnet 5.5 & 29.33 & 6.53 & 0.223 & 16.27 & $-$0.65 & 15.62 \\
DOTA & GPT-6.1 & Gemini 3.8 & 27.08 & 5.72 & 0.211 & 15.65 & $-$1.50 & 14.15 \\
DOTA & Sonnet 5.5 & same model & 6.15 & 0.87 & 0.141 & 4.42 & 0.05 & 4.47 \\
DOTA & Sonnet 5.5 & GPT-6.1 & 6.20 & 0.32 & 0.051 & 5.57 & 2.92 & 8.48 \\
DOTA & Sonnet 5.5 & Gemini 3.8 & 10.30 & 2.28 & 0.222 & 5.73 & 1.25 & 6.98 \\
DOTA & Gemini 3.8 & same model & 11.25 & 1.87 & 0.166 & 7.52 & 0.82 & 8.33 \\
DOTA & Gemini 3.8 & GPT-6.1 & 7.60 & 0.68 & 0.090 & 6.23 & 3.90 & 10.13 \\
DOTA & Gemini 3.8 & Sonnet 5.5 & 13.93 & 2.50 & 0.179 & 8.93 & 1.17 & 10.10 \\
\midrule
FSC-147 & GPT-6.1 & same model & 2.78 & 1.77 & 0.635 & $-$0.75 & 0.27 & $-$0.48 \\
FSC-147 & GPT-6.1 & Sonnet 5.5 & 22.60 & 20.05 & 0.887 & $-$17.50 & 0.00 & $-$17.50 \\
FSC-147 & GPT-6.1 & Gemini 3.8 & 27.28 & 24.52 & 0.899 & $-$21.75 & 0.10 & $-$21.65 \\
FSC-147 & Sonnet 5.5 & same model & 7.35 & 4.82 & 0.655 & $-$2.28 & 0.10 & $-$2.18 \\
FSC-147 & Sonnet 5.5 & GPT-6.1 & 4.53 & 2.10 & 0.463 & 0.33 & 0.37 & 0.70 \\
FSC-147 & Sonnet 5.5 & Gemini 3.8 & 11.23 & 8.30 & 0.739 & $-$5.37 & 0.03 & $-$5.33 \\
FSC-147 & Gemini 3.8 & same model & 3.53 & 3.12 & 0.882 & $-$2.70 & 0.07 & $-$2.63 \\
FSC-147 & Gemini 3.8 & GPT-6.1 & 0.82 & 0.32 & 0.388 & 0.18 & 0.10 & 0.28 \\
FSC-147 & Gemini 3.8 & Sonnet 5.5 & 2.83 & 2.28 & 0.806 & $-$1.73 & 0.13 & $-$1.60 \\
\midrule
SKU-110K & GPT-6.1 & same model & 33.92 & 0.52 & 0.015 & 32.88 & 5.83 & 38.72 \\
SKU-110K & GPT-6.1 & Sonnet 5.5 & 153.82 & 43.05 & 0.280 & 67.72 & $-$14.07 & 53.65 \\
SKU-110K & GPT-6.1 & Gemini 3.8 & 113.82 & 22.83 & 0.201 & 68.15 & 1.17 & 69.32 \\
SKU-110K & Sonnet 5.5 & same model & 61.95 & 2.95 & 0.048 & 56.05 & 0.87 & 56.92 \\
SKU-110K & Sonnet 5.5 & GPT-6.1 & 59.23 & 0.23 & 0.004 & 58.77 & 13.83 & 72.60 \\
SKU-110K & Sonnet 5.5 & Gemini 3.8 & 73.15 & 5.57 & 0.076 & 62.02 & 9.47 & 71.48 \\
SKU-110K & Gemini 3.8 & same model & 36.88 & 13.02 & 0.353 & 10.85 & 1.43 & 12.28 \\
SKU-110K & Gemini 3.8 & GPT-6.1 & 13.07 & 0.40 & 0.031 & 12.27 & 9.67 & 21.93 \\
SKU-110K & Gemini 3.8 & Sonnet 5.5 & 66.98 & 25.87 & 0.386 & 15.25 & $-$10.10 & 5.15 \\
\midrule
GroundCUA & GPT-6.1 & same model & 5.85 & 1.87 & 0.319 & 2.12 & $-$0.67 & 1.45 \\
GroundCUA & GPT-6.1 & Sonnet 5.5 & 34.58 & 4.00 & 0.116 & 26.58 & 13.03 & 39.62 \\
GroundCUA & GPT-6.1 & Gemini 3.8 & 43.42 & 17.02 & 0.392 & 9.38 & $-$6.30 & 3.08 \\
GroundCUA & Sonnet 5.5 & same model & 5.17 & 3.02 & 0.584 & $-$0.87 & 1.20 & 0.33 \\
GroundCUA & Sonnet 5.5 & GPT-6.1 & 13.82 & 8.07 & 0.584 & $-$2.32 & $-$6.50 & $-$8.82 \\
GroundCUA & Sonnet 5.5 & Gemini 3.8 & 38.90 & 27.83 & 0.716 & $-$16.77 & $-$7.73 & $-$24.50 \\
GroundCUA & Gemini 3.8 & same model & 30.98 & 6.92 & 0.223 & 17.15 & $-$0.93 & 16.22 \\
GroundCUA & Gemini 3.8 & GPT-6.1 & 26.32 & 1.83 & 0.070 & 22.65 & 9.60 & 32.25 \\
GroundCUA & Gemini 3.8 & Sonnet 5.5 & 42.57 & 2.45 & 0.058 & 37.67 & 15.43 & 53.10 \\
\midrule
ScreenParse & GPT-6.1 & same model & 0.38 & 0.08 & 0.217 & 0.22 & 0.07 & 0.28 \\
ScreenParse & GPT-6.1 & Sonnet 5.5 & 4.90 & 1.40 & 0.286 & 2.10 & 0.93 & 3.03 \\
ScreenParse & GPT-6.1 & Gemini 3.8 & 4.12 & 1.25 & 0.304 & 1.62 & 0.03 & 1.65 \\
ScreenParse & Sonnet 5.5 & same model & 0.80 & 0.30 & 0.375 & 0.20 & 0.00 & 0.20 \\
ScreenParse & Sonnet 5.5 & GPT-6.1 & 1.80 & 0.63 & 0.352 & 0.53 & $-$0.47 & 0.07 \\
ScreenParse & Sonnet 5.5 & Gemini 3.8 & 2.83 & 1.42 & 0.500 & 0.00 & $-$0.37 & $-$0.37 \\
ScreenParse & Gemini 3.8 & same model & 2.17 & 0.98 & 0.454 & 0.20 & 0.00 & 0.20 \\
ScreenParse & Gemini 3.8 & GPT-6.1 & 1.57 & 0.30 & 0.192 & 0.97 & 0.47 & 1.43 \\
ScreenParse & Gemini 3.8 & Sonnet 5.5 & 3.38 & 0.93 & 0.276 & 1.52 & 0.87 & 2.38 \\
\end{longtable}\ltlabelafter{tab:intersect_full}}

\begin{table}[H]\centering\footnotesize\setlength{\tabcolsep}{4pt}
\caption{Click alarms at matched acceptance coverage (260 items; data of Table~\ref{tab:alarm} and Figure~\ref{fig:alarm}; point estimates without intervals). Rows are the acceptance coverages that the same-model and the cross-model alarm both reach on the grid of $\delta$; miss rates are those of the acted-on items, lower is better.}\label{tab:alarm_matched}
\begin{tabular}{@{}llrrr@{}}\toprule
Model & Points & \begin{tabular}[b]{@{}c@{}}Acceptance\\coverage (\%)\end{tabular} & \begin{tabular}[b]{@{}c@{}}Miss rate,\\same model (\%)\end{tabular} & \begin{tabular}[b]{@{}c@{}}Miss rate,\\cross-model (\%)\end{tabular} \\\midrule
GPT-6.1 & unzoomed & 93.1 & 7.0 & 5.8 \\
 & after zoom & 90.4 & 3.8 & 2.9 \\
 &  & 91.7 & 3.8 & 3.4 \\
 &  & 93.1 & 3.7 & 3.7 \\
\midrule
Sonnet 5.5 & unzoomed & 87.7 & 10.5 & 10.1 \\
 &  & 88.5 & 11.3 & 10.6 \\
 &  & 89.2 & 12.1 & 11.2 \\
 & after zoom & \multicolumn{3}{l}{the two curves do not overlap} \\
\midrule
Gemini 3.8 & unzoomed & 78.1 & 19.2 & 16.5 \\
 &  & 83.7 & 20.5 & 18.8 \\
 &  & 89.2 & 21.9 & 20.7 \\
 & after zoom & 87.3 & 5.3 & 3.8 \\
 &  & 88.6 & 5.5 & 4.5 \\
 &  & 90.0 & 5.7 & 5.1 \\
\bottomrule\end{tabular}\end{table}

{\footnotesize\setlength{\tabcolsep}{4pt}\begin{longtable}{@{}ll>{\raggedright\arraybackslash}p{4.2cm}rrrrr@{}}
\caption{Leave-one-source-out transfer, every cell (data of Table~\ref{tab:transfer}). Default applied: the construction chosen on the other five sources; ``averaged'' means the coordinates of the kept records are averaged over the calls that report them. Single and Default: loss of one call of the model and of the default applied. 95\% CI: paired image bootstrap of the change, default minus single (negative is better for $E$, positive for F1). \textbf{Bold}: the paired 95\% interval excludes 0 in the favourable direction; \underline{underlined}: it excludes 0 in the unfavourable direction. Recall (\%) and cost per image (\$, every call counted) are those of the default applied.}\ltlabel{tab:transfer_full}\\\toprule
Source & Model & Default applied & Single & Default & 95\% CI & Recall & Cost \\\midrule\endfirsthead
\multicolumn{8}{@{}l}{\emph{Table~\ref{tab:transfer_full}, continued}}\\\toprule Source & Model & Default applied & Single & Default & 95\% CI & Recall & Cost \\\midrule\endhead
\midrule\multicolumn{8}{r}{\emph{continued on the next page}}\\\endfoot
\bottomrule\endlastfoot
\multicolumn{8}{@{}l}{\emph{Same-model constructions only; set error $E$ per image (lower is better)}} \\
\quad DocLayNet & GPT-6.1 & unanimity of 3 calls, averaged & 8.57 & 9.15 & [$-$0.35, 1.65] & 67.7 & 0.0323 \\
\quad DocLayNet & Sonnet 5.5 & unanimity of 3 calls, averaged & 13.77 & 12.33 & {\bfseries\boldmath [$-$2.25, $-$0.70]} & 57.1 & 0.0384 \\
\quad DocLayNet & Gemini 3.8 & unanimity of 3 calls, averaged & 11.83 & 12.18 & [$-$0.63, 1.42] & 23.0 & 0.0132 \\
\quad DOTA & GPT-6.1 & unanimity of 3 calls, averaged & 45.00 & 34.98 & {\bfseries\boldmath [$-$16.73, $-$4.97]} & 46.4 & 0.0785 \\
\quad DOTA & Sonnet 5.5 & unanimity of 3 calls, averaged & 37.87 & 31.95 & {\bfseries\boldmath [$-$8.70, $-$3.77]} & 32.1 & 0.0472 \\
\quad DOTA & Gemini 3.8 & unanimity of 3 calls, averaged & 40.98 & 30.80 & {\bfseries\boldmath [$-$16.38, $-$5.38]} & 25.8 & 0.0178 \\
\quad FSC-147 & GPT-6.1 & unanimity of 3 calls, averaged & 30.12 & 30.90 & [$-$0.10, 1.78] & 63.7 & 0.0334 \\
\quad FSC-147 & Sonnet 5.5 & unanimity of 3 calls, averaged & 48.32 & 51.73 & \underline{[2.07, 4.83]} & 29.2 & 0.0187 \\
\quad FSC-147 & Gemini 3.8 & unanimity of 3 calls, averaged & 52.05 & 54.63 & \underline{[0.42, 5.42]} & 18.9 & 0.0097 \\
\quad SKU-110K & GPT-6.1 & unanimity of 3 calls, averaged & 198.35 & 148.32 & {\bfseries\boldmath [$-$58.67, $-$41.80]} & 52.9 & 0.2121 \\
\quad SKU-110K & Sonnet 5.5 & unanimity of 3 calls, averaged & 217.30 & 148.67 & {\bfseries\boldmath [$-$80.43, $-$57.82]} & 8.4 & 0.1258 \\
\quad SKU-110K & Gemini 3.8 & unanimity of 3 calls, averaged & 150.97 & 136.48 & {\bfseries\boldmath [$-$23.90, $-$6.18]} & 19.2 & 0.0584 \\
\quad GroundCUA & GPT-6.1 & unanimity of 3 calls, averaged & 92.58 & 89.25 & [$-$7.40, 0.50] & 49.2 & 0.0920 \\
\quad GroundCUA & Sonnet 5.5 & unanimity of 3 calls, averaged & 44.15 & 43.67 & [$-$3.13, 2.43] & 68.0 & 0.0999 \\
\quad GroundCUA & Gemini 3.8 & unanimity of 3 calls, averaged & 121.75 & 95.58 & {\bfseries\boldmath [$-$34.90, $-$17.95]} & 14.4 & 0.0438 \\
\quad ScreenParse & GPT-6.1 & unanimity of 3 calls, averaged & 16.23 & 15.78 & {\bfseries\boldmath [$-$0.83, $-$0.10]} & 59.7 & 0.0292 \\
\quad ScreenParse & Sonnet 5.5 & unanimity of 3 calls, averaged & 13.27 & 12.97 & [$-$0.75, 0.17] & 59.7 & 0.0318 \\
\quad ScreenParse & Gemini 3.8 & unanimity of 3 calls, averaged & 16.02 & 15.53 & [$-$1.28, 0.20] & 47.7 & 0.0109 \\
\midrule
\multicolumn{8}{@{}l}{\emph{Same-model constructions only; F1 in \% (higher is better)}} \\
\quad DocLayNet & GPT-6.1 & union of 3 calls & 71.0 & 68.6 & \underline{[$-$4.9, $-$0.3]} & 70.8 & 0.0323 \\
\quad DocLayNet & Sonnet 5.5 & union of 3 calls & 57.3 & 56.9 & [$-$2.1, 1.4] & 62.9 & 0.0384 \\
\quad DocLayNet & Gemini 3.8 & union of 3 calls & 38.4 & 45.9 & {\bfseries\boldmath [2.8, 12.9]} & 48.4 & 0.0132 \\
\quad DOTA & GPT-6.1 & union of 3 calls & 39.5 & 38.9 & [$-$2.4, 1.2] & 50.9 & 0.0785 \\
\quad DOTA & Sonnet 5.5 & union of 3 calls & 33.4 & 32.9 & [$-$2.2, 1.4] & 37.6 & 0.0472 \\
\quad DOTA & Gemini 3.8 & union of 3 calls & 31.2 & 31.9 & [$-$2.5, 4.4] & 36.1 & 0.0178 \\
\quad FSC-147 & GPT-6.1 & majority of 3 calls, averaged & 70.7 & 69.5 & [$-$4.4, 0.8] & 67.0 & 0.0334 \\
\quad FSC-147 & Sonnet 5.5 & majority of 3 calls, averaged & 45.2 & 42.5 & \underline{[$-$4.6, $-$0.8]} & 35.5 & 0.0187 \\
\quad FSC-147 & Gemini 3.8 & majority of 3 calls, averaged & 27.3 & 27.7 & [$-$3.4, 4.2] & 24.5 & 0.0097 \\
\quad SKU-110K & GPT-6.1 & union of 3 calls & 43.7 & 40.2 & \underline{[$-$4.7, $-$2.4]} & 54.1 & 0.2121 \\
\quad SKU-110K & Sonnet 5.5 & union of 3 calls & 12.8 & 13.0 & [$-$0.7, 1.1] & 15.7 & 0.1258 \\
\quad SKU-110K & Gemini 3.8 & union of 3 calls & 35.4 & 40.2 & {\bfseries\boldmath [2.2, 8.0]} & 39.0 & 0.0584 \\
\quad GroundCUA & GPT-6.1 & union of 3 calls & 48.3 & 46.2 & \underline{[$-$3.2, $-$1.0]} & 52.7 & 0.0920 \\
\quad GroundCUA & Sonnet 5.5 & union of 3 calls & 71.7 & 73.0 & [$-$0.2, 2.9] & 74.5 & 0.0999 \\
\quad GroundCUA & Gemini 3.8 & union of 3 calls & 25.6 & 28.6 & [$-$0.4, 6.7] & 37.4 & 0.0438 \\
\quad ScreenParse & GPT-6.1 & union of 3 calls & 56.4 & 56.6 & [$-$0.4, 0.9] & 61.8 & 0.0292 \\
\quad ScreenParse & Sonnet 5.5 & union of 3 calls & 61.5 & 61.5 & [$-$0.9, 1.0] & 62.9 & 0.0318 \\
\quad ScreenParse & Gemini 3.8 & union of 3 calls & 53.6 & 54.7 & [$-$0.7, 3.1] & 58.8 & 0.0109 \\
\midrule
\multicolumn{8}{@{}l}{\emph{Any construction, including cross-model ones; set error $E$ per image (lower is better)}} \\
\quad DocLayNet & GPT-6.1 & intersection with Sonnet 5.5, averaged & 8.57 & 8.87 & [$-$1.23, 1.77] & 65.1 & 0.0236 \\
\quad DocLayNet & Sonnet 5.5 & intersection with GPT-6.1, averaged & 13.77 & 8.87 & {\bfseries\boldmath [$-$6.63, $-$3.37]} & 65.1 & 0.0236 \\
\quad DocLayNet & Gemini 3.8 & intersection with GPT-6.1, averaged & 11.83 & 10.35 & {\bfseries\boldmath [$-$2.67, $-$0.35]} & 35.7 & 0.0151 \\
\quad DOTA & GPT-6.1 & majority of 3 models, averaged & 45.00 & 31.43 & {\bfseries\boldmath [$-$21.90, $-$7.27]} & 41.0 & 0.0481 \\
\quad DOTA & Sonnet 5.5 & majority of 3 models, averaged & 37.87 & 31.47 & {\bfseries\boldmath [$-$10.40, $-$3.15]} & 40.9 & 0.0481 \\
\quad DOTA & Gemini 3.8 & majority of 3 models, averaged & 40.98 & 31.85 & {\bfseries\boldmath [$-$13.83, $-$5.02]} & 40.1 & 0.0481 \\
\quad FSC-147 & GPT-6.1 & intersection with Sonnet 5.5, averaged & 30.12 & 47.62 & \underline{[11.28, 24.48]} & 35.6 & 0.0175 \\
\quad FSC-147 & Sonnet 5.5 & intersection with GPT-6.1, averaged & 48.32 & 47.62 & [$-$2.28, 0.75] & 35.6 & 0.0175 \\
\quad FSC-147 & Gemini 3.8 & intersection with GPT-6.1, averaged & 52.05 & 51.77 & [$-$0.80, 0.08] & 23.4 & 0.0142 \\
\quad SKU-110K & GPT-6.1 & majority of 3 models, averaged & 198.35 & 138.55 & {\bfseries\boldmath [$-$72.30, $-$47.70]} & 38.2 & 0.1353 \\
\quad SKU-110K & Sonnet 5.5 & majority of 3 models, averaged & 217.30 & 139.43 & {\bfseries\boldmath [$-$92.02, $-$64.67]} & 37.9 & 0.1353 \\
\quad SKU-110K & Gemini 3.8 & majority of 3 models, averaged & 150.97 & 139.13 & {\bfseries\boldmath [$-$17.20, $-$6.78]} & 37.9 & 0.1353 \\
\quad GroundCUA & GPT-6.1 & majority of 3 models, averaged & 92.58 & 68.75 & {\bfseries\boldmath [$-$29.70, $-$18.27]} & 55.7 & 0.0787 \\
\quad GroundCUA & Sonnet 5.5 & majority of 3 models, averaged & 44.15 & 67.02 & \underline{[17.92, 27.90]} & 56.9 & 0.0787 \\
\quad GroundCUA & Gemini 3.8 & majority of 3 models, averaged & 121.75 & 71.03 & {\bfseries\boldmath [$-$59.05, $-$42.70]} & 54.3 & 0.0787 \\
\quad ScreenParse & GPT-6.1 & majority of 3 models, averaged & 16.23 & 14.02 & {\bfseries\boldmath [$-$3.18, $-$1.28]} & 61.3 & 0.0241 \\
\quad ScreenParse & Sonnet 5.5 & majority of 3 models, averaged & 13.27 & 13.82 & [$-$0.58, 1.70] & 61.7 & 0.0241 \\
\quad ScreenParse & Gemini 3.8 & majority of 3 models, averaged & 16.02 & 14.02 & {\bfseries\boldmath [$-$2.85, $-$1.28]} & 61.3 & 0.0241 \\
\midrule
\multicolumn{8}{@{}l}{\emph{Any construction, including cross-model ones; F1 in \% (higher is better)}} \\
\quad DocLayNet & GPT-6.1 & union with Sonnet 5.5 & 71.0 & 66.0 & [$-$10.1, 0.1] & 72.8 & 0.0236 \\
\quad DocLayNet & Sonnet 5.5 & union with GPT-6.1 & 57.3 & 66.0 & {\bfseries\boldmath [5.6, 12.0]} & 72.8 & 0.0236 \\
\quad DocLayNet & Gemini 3.8 & union with GPT-6.1 & 38.4 & 67.4 & {\bfseries\boldmath [21.5, 36.8]} & 73.6 & 0.0151 \\
\quad DOTA & GPT-6.1 & union with Sonnet 5.5 & 39.5 & 36.9 & [$-$5.2, 0.3] & 49.1 & 0.0419 \\
\quad DOTA & Sonnet 5.5 & union with GPT-6.1 & 33.4 & 36.9 & [$-$2.5, 9.5] & 49.1 & 0.0419 \\
\quad DOTA & Gemini 3.8 & union with GPT-6.1 & 31.2 & 36.6 & {\bfseries\boldmath [0.1, 10.7]} & 48.8 & 0.0324 \\
\quad FSC-147 & GPT-6.1 & majority of 3 models, averaged & 70.7 & 45.8 & \underline{[$-$32.8, $-$17.3]} & 39.0 & 0.0207 \\
\quad FSC-147 & Sonnet 5.5 & majority of 3 models, averaged & 45.2 & 45.7 & [$-$2.3, 3.4] & 38.8 & 0.0207 \\
\quad FSC-147 & Gemini 3.8 & majority of 3 models, averaged & 27.3 & 45.8 & {\bfseries\boldmath [12.1, 25.2]} & 39.0 & 0.0207 \\
\quad SKU-110K & GPT-6.1 & union with Sonnet 5.5 & 43.7 & 33.9 & \underline{[$-$11.6, $-$7.9]} & 46.1 & 0.1155 \\
\quad SKU-110K & Sonnet 5.5 & union with GPT-6.1 & 12.8 & 33.9 & {\bfseries\boldmath [18.1, 24.4]} & 46.1 & 0.1155 \\
\quad SKU-110K & Gemini 3.8 & union with GPT-6.1 & 35.4 & 42.9 & {\bfseries\boldmath [3.5, 12.1]} & 51.8 & 0.0898 \\
\quad GroundCUA & GPT-6.1 & union with Sonnet 5.5 & 48.3 & 59.7 & {\bfseries\boldmath [9.1, 13.7]} & 70.1 & 0.0640 \\
\quad GroundCUA & Sonnet 5.5 & union with GPT-6.1 & 71.7 & 59.7 & \underline{[$-$14.3, $-$9.8]} & 70.1 & 0.0640 \\
\quad GroundCUA & Gemini 3.8 & union with GPT-6.1 & 25.6 & 41.2 & {\bfseries\boldmath [11.2, 20.5]} & 50.3 & 0.0454 \\
\quad ScreenParse & GPT-6.1 & majority of 3 models, averaged & 56.4 & 59.6 & {\bfseries\boldmath [1.1, 5.5]} & 61.3 & 0.0241 \\
\quad ScreenParse & Sonnet 5.5 & majority of 3 models, averaged & 61.5 & 60.1 & [$-$4.1, 1.3] & 61.7 & 0.0241 \\
\quad ScreenParse & Gemini 3.8 & majority of 3 models, averaged & 53.6 & 59.6 & {\bfseries\boldmath [4.0, 8.4]} & 61.3 & 0.0241 \\
\end{longtable}\ltlabelafter{tab:transfer_full}}

\begin{table}[H]
\centering\footnotesize\setlength{\tabcolsep}{4pt}
\caption{The four fixed single-call configurations on the four detection sources: F1 (\%, primary scoring of Table~\ref{tab:contract}), the mean over sources of the per-source mean F1 of the first of the three stored calls, and cost per image (\$, list prices). The shipped F1 configuration of each model is the highest F1 of its column (bold); for Gemini 3.8 the two default-setting configurations differ by less than 0.2 points.}\label{tab:config_menu}
\begin{tabular}{@{}lrrrrrr@{}}\toprule
 & \multicolumn{2}{c}{GPT-6.1} & \multicolumn{2}{c}{Sonnet 5.5} & \multicolumn{2}{c}{Gemini 3.8} \\\cmidrule(lr){2-3}\cmidrule(lr){4-5}\cmidrule(l){6-7}
Configuration & F1 & Cost & F1 & Cost & F1 & Cost \\\midrule
list, lowest & 56.2 & 0.0295 & 37.1 & 0.0201 & 33.1 & 0.0084 \\
regions, lowest & \textbf{58.6} & 0.0327 & 31.6 & 0.0291 & 39.4 & 0.0101 \\
list, default & 54.0 & 0.0436 & 44.7 & 0.0237 & \textbf{45.0} & 0.0192 \\
regions, default & 56.7 & 0.0475 & \textbf{48.1} & 0.0477 & 44.8 & 0.0200 \\
\bottomrule\end{tabular}
\end{table}

\section{Experiment Registry, Costs and Settings}\label{app:registry}

\paragraph{Registry.} Table~\ref{tab:registry} lists every experiment with model calls. Counts include smoke tests, pilots, repairs and retries; costs are list prices for every attempted call. The open model was served through a routing provider and is priced at 0 here (0.82M input and 1.75M output tokens in the open-model experiment). The registry covers the models evaluated here; calls to an earlier generation of models, used only for background, are not part of it.

\begin{longtable}{@{}>{\raggedright\arraybackslash}p{3.0cm}>{\raggedright\arraybackslash}p{6.35cm}>{\raggedright\arraybackslash}p{2.3cm}rr@{}}
\caption{All experiments with model calls (67,364 calls, about \$758).}\ltlabel{tab:registry}\\\toprule
Experiment & Data and arms & Models & Calls & Cost \\\midrule\endfirsthead
\toprule Experiment & Data and arms & Models & Calls & Cost \\\midrule\endhead
Pilot v1 & 12 procedural scenes, 768 px; responsibility-split arms; cap 8,192 & G5.6, G6.1, S5.5, O5.5, Q & 3,540 & \$29.30 \\
Pilot v2 & 24 scenes, 1024 px, 16 to 384 targets; dose response & same & 4,080 & \$47.50 \\
Subset execution & 12 scenes, 96 or 192 targets; representation and drift arms & G5.6, G6.1, O5.5, S5.5 & 1,924 & \$53.84 \\
Scan responsibility and grids & 48 scenes; scan responsibility, grid on/off, granularity; repairs, smoke, supplement & above + Gem & 2,752 & \$113.80 \\
Detection & SKU-110K, DOTA, DocLayNet, FSC-147, 60 images each; list / regions $\times$ lowest / default; 16k control & G6.1, S5.5, Gem & 6,523 & \$171.13 \\
GUI & ScreenSpot-Pro 200, ScreenSpot-v2 60, GroundCUA 60, ScreenParse 60; click and parse arms; pilots 1,050 & same & 7,232 & \$67.77 \\
Same-view review & SKU, DOTA, FSC, 20 images each; draft review (same view, crop) & same & 663 & \$7.40 \\
Readout & SCP 100, SCP-dense 50, PixMo 100; 13 prompt arms & same & 7,900 & \$88.48 \\
Tiling & FSC-147 100 (three size strata); count-first, list-only, $2\times2$ tiles & same & 1,800 & \$8.02 \\
Review & GroundCUA 30 + SKU 30, four crops each; blind vs.\ draft (2 deleted, 2 injected, 1 moved) & same & 1,440 & \$15.41 \\
Comparator & 80 disagreeing clicks; SKU and DOTA crops, 150 per model; both orders & same & 1,960 & \$4.76 \\
Resolution & ScreenSpot-Pro 100; down / native / up & same & 600 & \$1.52 \\
End-to-end & SCP-dense 50, FSC 100, clicks 200; naive / common / proposed & same & 1,800 & \$22.65 \\
Reasoning & SCP-dense 30, FSC 30, GroundCUA 20; default setting & same & 1,110 & \$26.37 \\
Open model & SCP 100, SCP-dense 50; readout arms & Q & 650 & not priced \\
Grid & SCP 100, SCP-dense 40; grid, ticks, cells, scan, type names; 100 default calls & G6.1, S5.5, Gem & 1,340 & \$15.40 \\
Pilots & small runs before the readout experiments & same & 647 & \$5.54 \\
Readout, full protocol & SCP-dense 50; full list-first protocol, lowest setting, 16k cap & same & 150 & \$3.87 \\
\quad Gemini, default 16k & same images; default setting, 16k cap & Gem & 50 & \$1.71 \\
\quad Gemini, default 32k & same images; default setting, 32k cap & Gem & 50 & \$1.83 \\
Multi-class readout & DOTA 30 windows, DocLayNet 50 pages; count-first and list-first, lowest setting; Gemini list-first at default, 16k cap; 14 smoke calls & G6.1, S5.5, Gem & 574 & \$6.25 \\
Zoom on second click & the 260 click items of GUI; two-stage zoom around each model's second click, the same request as the first-click zoom & G6.1, S5.5, Gem & 772 & \$1.60 \\
Click scores & the 260 click items of GUI; each model's zoomed point scored by each model (hit score) & G6.1, S5.5, Gem & 2,325 & \$3.91 \\
Click choice & the 260 click items; default-setting zooms of GPT-6.1 and Sonnet 5.5, re-grounding and second-opinion arms, scores of default-setting points, pairwise comparisons of cluster representatives, crop-first baselines & same & 3,639 & \$13.56 \\
Clicks, 600 items & 600 further ScreenSpot-Pro items; all arms of Table~\ref{tab:choose}; 13,120 requests and 6 retries & same & 13,126 & \$44.93 \\
Fourth click source & the same 600 items; scorer's scores of the crop-first point and the comparisons the four-source tier needs that were not already asked & Gem & 717 & \$1.44 \\\midrule
Total & & & 67,364 & \$757.99 \\\bottomrule
\end{longtable}\ltlabelafter{tab:registry}
\noindent G5.6, G6.1: GPT-5.6, GPT-6.1. S5.5: Sonnet 5.5. O5.5: Opus 5.5. Gem: Gemini 3.8 Flash. Q: Qwen3-VL-235B (instruct). The open-model calls have no logged per-token price, so the total cost excludes them.

\paragraph{Settings audit.}\label{app:settings} Table~\ref{tab:settings} records what each setting label meant in practice. The lowest setting used an output cap of 16,384 tokens and the default setting 32,768, except where a 16k default was run to separate the two (Table~\ref{tab:setting}, a 90-call control on the detection sources, and a control on controlled scenes).

\begin{table}[h]
\centering\small
\caption{Reasoning settings as sent and as observed (share of calls with reasoning tokens).}\label{tab:settings}
\begin{tabular}{@{}l>{\raggedright\arraybackslash}p{3.8cm}>{\raggedright\arraybackslash}p{8.0cm}@{}}\toprule
Model & Lowest setting as sent & Observed \\\midrule
GPT-6.1 & \texttt{reasoning\_effort=low} & reasoning in most calls (SCP 100\%, PixMo 61\%, clicks 39\%); default: 100\% \\
GPT-5.6 & effort none & no reasoning \\
Sonnet 5.5 & \texttt{thinking: between\_tools} & lowest: 0\%; default (adaptive thinking): 0\% to 23\% on the dense, FSC and screen runs, 4 of 48 calls on controlled scenes, 0\% to 90\% on the detection sources by cell; pilots: not logged \\
Opus 5.5 & effort low (adaptive thinking cannot be disabled) & under 1\% of logged calls at effort low; pilots: not logged \\
Gemini 3.8 & \texttt{thinking\_budget=0} & little on list and box tasks (FSC list 6\%, regions 12\%); still reasons on clicks and points (GUI clicks about 99\%, PixMo 23\%, FSC points 32\%); default: 51\% to 100\% \\
Qwen3-VL & instruct model, no reasoning parameter & no reasoning \\\bottomrule
\end{tabular}
\end{table}

\section{Protocol Details}\label{app:protocol}

\paragraph{List-first prompt.} The full-protocol readout runs used the following system prompt for GPT-6.1 and Gemini 3.8 (native 0 to 1000 frame); the user message was ``List every geometric shape in this image with its center, then counts by type. Respond with JSON only.''

{\small
\begin{verbatim}
You are a precise visual counting and localization assistant.

Given an image containing geometric shapes, identify every shape.
Report coordinates on a 0-1000 grid: x = 0 is the left edge and x = 1000 the
right edge of the image, y = 0 is the top edge and y = 1000 the bottom edge.
Use integers.
List the shapes in reading order: scan the image row by row from top to bottom,
and within a row from left to right.
The "type" value must be exactly one of: circle, triangle, square, star,
pentagon. A square rotated by any angle is still a "square"; do not use any
other type name.

Respond with ONLY a JSON object (no markdown, no commentary) in this format:
{"shapes": [{"type": "circle", "x": 402, "y": 311}, ...], "counts_by_type":
{"circle": 5, "triangle": 3, "square": 0, "star": 2, "pentagon": 4}}

Shape types to recognize: circle, triangle, square, star, pentagon.
Locate every shape carefully. Count systematically; do not miss any and do not
invent shapes that are not there.
Write the full shapes list first; after it, give counts_by_type for the shapes
you listed.
\end{verbatim}
}
\noindent For Sonnet 5.5 the coordinate paragraph reads ``The image is 1024x1024 pixels. Report coordinates as integer pixel positions in this image, where (0,0) is the top-left corner.'' The count-first prompts of the readout experiment ask for \texttt{counts\_by\_type} before the list; all prompts are released with the responses.

\paragraph{Decoder.} The decoder strips a code fence and parses the whole response as JSON (L2). If that fails, it recovers each balanced \texttt{\{...\}} object that has the record fields (declared salvage) and reads \texttt{counts\_by\_type} if present, with its class names mapped by the same alias map (the pure-readout comparisons of Table~\ref{tab:readout} compare stated and list counts with neither mapped). Coordinates are mapped from the model's frame to image pixels. Classes are lower-cased and mapped by the declared alias map; for SCP, \{diamond, rhombus, rotated square\} $\mapsto$ square. Records whose class is outside the vocabulary after mapping stay in the audit collection $Z^{\mathrm{all}}$, are excluded from the delivered collection $Z$ that set scores evaluate, and are counted by the first degeneracy flag; Table~\ref{tab:setting} adds a sensitivity that keeps them and scores them as unmatched. On the detection sources, records with a label outside the vocabulary or a box outside the image are likewise kept in the audit collection, excluded from delivery and counted as invalid (Table~\ref{tab:contract}).

\paragraph{Degeneracy flags.} A list is flagged if any of the following holds: more than 10\% of records lack a valid class; the list length is outside $[0.5,1.5]$ times the stated total; more than 10\% of records repeat a rounded coordinate pair already in the list; or the finish reason is ``length''. The same thresholds are used for every model, including the open model.

\paragraph{Scoring and clustering.} Scoring uses the compatibility relation of each task in Table~\ref{tab:contract}: for example, points of equal class within 30 pixels on SCP and 20 pixels on SCP-dense, detection boxes of equal class at IoU $\ge 0.75$, screen elements at IoU $\ge 0.5$ without class, and clicks inside the target box. Maximum matchings are computed with a linear assignment on the 0/1 compatibility matrix \citep{kuhn1955hungarian}. Clusters across calls use the association rule of Table~\ref{tab:contract} (boxes: equal class and IoU $\ge0.5$, also on the detection sources where scoring uses 0.75), with a maximum-cardinality, then maximum-overlap assignment against cluster representatives, and the averaged record is the coordinate mean of members.

\paragraph{Constructions.} Same-model: \texttt{inter2} (two calls, intersection), \texttt{union2}, \texttt{unan3} and \texttt{maj3} (three calls, unanimity or majority), \texttt{union3}; each with averaged (\texttt{\_fus}) or base-record (\texttt{\_del}) coordinates. Cross-model: \texttt{x2\_inter} and \texttt{x2\_union} with a named partner, \texttt{x3\_unan} and \texttt{x3\_maj} over the three models (one call each), \texttt{x3\_union}. The leave-one-source-out default for an evaluated source is the construction with the best mean relative change over the other five sources, computed separately for each loss and pool.

\paragraph{Alarms.} For clicks, the same-model alarm compares two unzoomed calls of one model; the cross-model alarm compares one unzoomed call of the user's model with one of the partner; after zoom, the same-model alarm compares the model's zoomed first click with its zoomed second click, and the cross-model alarm compares the zoomed first clicks of the two models. An item is set aside in the risk-coverage analysis when the two points are farther apart than $\delta$ times the longer image side. The partner is set by a fixed rule, the cheapest other model per click call (GPT-6.1 and Sonnet 5.5: Gemini 3.8; Gemini 3.8: Sonnet 5.5).

\paragraph{Clicks from several models.} Candidates are the two-stage zoom points of Step 4, mapped to the original screenshot. A candidate's identifier is the first 16 hexadecimal digits of the SHA-256 of the item, model, setting, parent call and coordinates. The scorer's score request shows a native crop of one fifth of the screenshot's width and height around the candidate, with a marker at the point, the instruction and the platform, and asks for a verdict and the probability that a click at the marker hits the intended element; it never names the model that produced the candidate. A missing or unreadable score counts as $-1$ for ranking. The comparison request shows the downscaled screenshot with both representatives marked by their identifiers and a native crop around each, asks which candidate satisfies the instruction (element, text, state, relation and whether the exact point lies inside the element), and allows ``undecided''; it forbids proposing a new point. Each pair is sent in both orders, and the code asserts that the points drawn equal the representatives' coordinates. Ranking keys, ties and failures follow Algorithm~\ref{alg:clicks}: a failed first click leaves that model without a candidate, a failed zoom keeps the first click, a failed comparison gives no win.

\paragraph{The 600 click items and baselines.} The 600 items were sampled from the ScreenSpot-Pro annotations not among the 200 ScreenSpot-Pro items of the GUI experiment, in proportion to each application file and, within it, to each pair of group and element type, with a fixed seed; they share no screenshot with those items. A first batch of 100 seeded items, part of the 600, was run through every arm to check costs and transport only; its calls are among the 13,126. The three-call baseline runs three independent default-setting zooms of GPT-6.1, clusters the three points with the same $\delta$, and returns the earliest point of the largest cluster (ties: the cluster with the earliest point). The crop-first baseline uses GPT-6.1 at its default setting: the first call must return a region of the downscaled screenshot, the second sees that region at native resolution with 8\% padding and may return a smaller region or a click, and a third call, if needed, must click; an unreadable reply falls back to the centre of the last region. It follows the coarse-to-fine cropping of GUI-Lens \citep{fu2026guilens} without its OCR and detector anchors or verification step, with our own prompts. Paired intervals resample screenshots (4,000 draws, seed 20261014); McNemar tests are exact.

\paragraph{Configuration transfer on the detection sources.} The menu contains the four single-call configurations run on every image of the four detection sources (first repetition): list or regions interface, each at the lowest and the default setting. In the regions interface the model returns one row per region of a fixed coordinate registry, with a status of empty, found or unknown, and reports each object in the region that contains its centre; in the list interface it returns one list for the whole image in registry scan order. For each model and evaluated source, F1 picks the configuration with the highest mean of per-source mean F1 on the other three sources, and set error the one with the lowest mean relative change against the list interface at the lowest setting; ties keep the earlier configuration in the order list-lowest, regions-lowest, list-default, regions-default. Intervals are paired bootstrap intervals over the images of the evaluated source.

\paragraph{Detection prompts.} Prompts below are verbatim; line breaks inside long lines are added for display. The detection sources used one system prompt per image, assembled from a task sentence, a coordinate contract, an output contract and a schema example generated from the region registry; the user message carried the registry as JSON. For the list interface on a DocLayNet page the system prompt was:

{\footnotesize
\begin{verbatim}
You are performing visual set reporting. This is a document page. Find every layout
element and label it with exactly one of: Caption, Footnote, Formula, List-item,
Page-footer, Page-header, Picture, Section-header, Table, Text, Title. Each text
block, list item, section header, page header, page footer, footnote, caption,
table, picture, formula and title is one element. A coordinate registry of regions
is given as public metadata: a region is an area to inspect, not a pre-detected
object, and it may contain zero, one, or several matching instances. Scan the image
systematically, region by region in registry order, and check every region before
you finish; do not stop early. Do not invent objects to reach an expected number.
Keep distinct nearby instances distinct.

Each object has x1, y1, x2, y2: the left, top, right and bottom edges of its
axis-aligned bounding box, normalized to the FULL IMAGE (native = image size times
value), with 0 <= x1 < x2 <= 1 and 0 <= y1 < y2 <= 1. Give four decimal places. Each
object also has label, one of the allowed labels exactly as written.

Output contract: return exactly one JSON object with status and objects. objects is
a single list of all matching objects in the whole image, listed in registry scan
order. status is found when you have completed your inspection of every region,
empty when the image contains no matching object, and unknown when you could not
complete it (include all recovered objects). No region rows, total count, prose,
code fence or explanation.

Example of the schema only, not an answer for the current image:
{"status":"found","objects":[{"x1":0.3078,"y1":0.1259,"x2":0.3478,"y2":0.1659,"label
":"Caption"},{"x1":0.4173,"y1":0.0366,"x2":0.4573,"y2":0.0766,"label":"Caption"},{"x
1":0.0829,"y1":0.3129,"x2":0.1229,"y2":0.3529,"label":"Caption"}]}
\end{verbatim}
}

\noindent The regions interface replaces the output contract and the schema example by:

{\footnotesize
\begin{verbatim}
Output contract: return exactly one JSON object per registered region, one object
per line (NDJSON), in registry order, covering EVERY region in the registry. Each
object has region_id, status and objects. status is empty, found, or unknown. Use
empty only when you judge that no matching object center is in the region. Use found
when you have completed your best inspection and report all matching objects whose
center is in that region. Use unknown when you cannot complete the judgment; partial
recovered objects may be included. Report an object only in the region containing
its center. No surrounding list, prose, code fence, total count or explanation.

Example of the schema only, not an answer for the current image:
{"region_id":"R001","status":"empty","objects":[]}
{"region_id":"R002","status":"found","objects":[{"x1":0.3078,"y1":0.1259,"x2":0.3478
,"y2":0.1659,"label":"Caption"},{"x1":0.4173,"y1":0.0366,"x2":0.4573,"y2":0.0766,"la
bel":"Caption"}]}
{"region_id":"R003","status":"unknown","objects":[]}
{"region_id":"R005","status":"found","objects":[{"x1":0.0829,"y1":0.3129,"x2":0.1229
,"y2":0.3529,"label":"Caption"}]}
\end{verbatim}
}

\noindent The user message, for both interfaces, was (registry abridged):

{\footnotesize
\begin{verbatim}
Inspect the image using the region registry in the following payload. Follow the
output contract exactly.

{"image_width": 1025, "image_height": 1025, "coordinate_contract":
"full_image_normalized_half_open_v1", "ordered_registry": [{"region_id": "R001",
"x0": 0, "y0": 0, "width": 256, "height": 256}, ..., {"region_id": "R016", ...}],
"allowed_labels": [...]}
\end{verbatim}
}

\noindent The task sentence for the other sources: SKU-110K, ``Find every individual product on the shelves in the image: each separately visible package or item, including partly visible ones.''; DOTA, ``This is an aerial image. Find every object of the following categories: <categories>. Give each object its category exactly as written.''; FSC-147, ``The first image is a reference crop showing one instance of the target object. Find every instance of this object in the second image (the scene), including the instance the reference was cropped from.''. Point tasks (FSC-147) use the coordinate contract ``Each object has x and y, the coordinates of its center normalized to the FULL IMAGE: native\_x = image\_width*x, native\_y = image\_height*y, both at least 0 and strictly below 1. Give four decimal places.'', and the FSC-147 user message is preceded by ``The first image is the reference crop; the second image is the scene.''

\paragraph{Click prompts.} The first click of two-stage zoom used the system prompt

{\footnotesize
\begin{verbatim}
You are a computer-use agent. Given a screenshot and an instruction, return the
point to click to carry out the instruction. Coordinates are integers on a 0 to 1000
grid of the FULL IMAGE: 0 is the left (or top) edge and 1000 the right (or bottom)
edge, whatever the pixel size of the image. Never give pixel coordinates.

Output contract: one JSON object {"x": ..., "y": ...}; if the target is not visible,
{"status": "not_found"}. No prose, no code fence.
\end{verbatim}
}

\noindent with the user message \texttt{Platform: macos. Instruction: "open options of terminal in android studio". Return the click point.} (platform and instruction of the item). For Sonnet 5.5 the coordinate sentences read ``Coordinates are pixel coordinates of the FULL IMAGE exactly as you see it: x from 0 at its left edge to its width at the right edge, y from 0 at its top edge to its height at the bottom edge. Pixel sizes are given in the request.'', and the user message begins with the image size (for example \texttt{Pixel sizes: image 1429x804.}); the crop prompt changes in the same way. The click in the crop used

{\footnotesize
\begin{verbatim}
You are a computer-use agent. The image is a zoomed crop of a larger screenshot, at
native resolution. Given an instruction, return the point to click inside THIS crop.
Coordinates are integers on a 0 to 1000 grid of the CROP: 0 is the left (or top)
edge and 1000 the right (or bottom) edge, whatever the pixel size of the image.
Never give pixel coordinates. If the target element is not inside this crop, say so.

Output contract: one JSON object {"x": ..., "y": ...} or {"status": "not_in_crop"}.
No prose, no code fence.
\end{verbatim}
}

\noindent with \texttt{Platform: macos. Instruction: "open options of terminal in android studio". Return the click point in this crop.} The scorer's request (Algorithm~\ref{alg:clicks}, lines 5 and 6) used

{\footnotesize
\begin{verbatim}
You are checking a click proposed by a computer-use agent. The image is a zoomed
crop of a screenshot at native resolution; the red circle with a cross marks the
proposed click. Judge whether clicking at the center of the marker would hit the UI
element the instruction refers to. Do not propose another point.

Output contract: one JSON object {"decision": "correct" or "uncertain" or "wrong",
"p_hit": a number from 0 to 1}, where p_hit is your probability that clicking at the
center of the marker hits the intended element. Give your best estimate even when
unsure. No prose, no code fence.
\end{verbatim}
}

\noindent with \texttt{Platform: macos. Instruction: "open options of terminal in android studio". Is the marked click correct?} The comparison (Algorithm~\ref{alg:clicks}, lines 10 to 12) used

{\footnotesize
\begin{verbatim}
You are comparing two fixed predictions for the original visual task: a click on a
screenshot of a computer application.

Each candidate has an immutable ID and is marked on the images with its ID label.
View "overview" is the full screenshot, downscaled, with both candidates marked.
View "crop_<ID>" is a native-resolution crop around that candidate, with the
candidate marked. Candidate source or model names are not provided.

Compare the candidates using the actual task requirements:
- the requested element or control;
- the exact text, state, or attribute when specified;
- the requested relation and its anchor or reference frame (panel, toolbar,
neighbouring label);
- whether the exact candidate point is inside the intended target element.

Inspect the information that distinguishes the candidates. Do not merely decide
whether both regions contain a control. Do not rewrite any coordinates or invent a
new candidate. Cite only supplied view IDs.

Output contract: one JSON object {"preferred_id": one of the supplied candidate IDs
or "undecided", "discriminating_requirement": "identity" or "text" or "state" or
"relation" or "geometry" or "none", "evidence_view_ids": a list of supplied view
IDs}. No prose, no code fence.
\end{verbatim}
}

\noindent with the user message

{\footnotesize
\begin{verbatim}
Original task (platform macos): "open options of terminal in android studio"
Candidate IDs: C85E5 (red), C10D4 (blue).
Views in order: overview, crop_C85E5, crop_C10D4.
Which candidate satisfies the original task? Follow the output contract exactly.
\end{verbatim}
}

\paragraph{Reproducing the tables.} The code and all requests and responses are at \url{https://github.com/shijunzhe/vlm-set-reporting}, together with the labels whose licences allow redistribution (DocLayNet, ScreenSpot-Pro, ScreenSpot-v2, GroundCUA, ScreenParse, PixMo-Count). The labels of DOTA v1.0, SKU-110K and FSC-147 are not redistributed: the repository stores a placeholder for each value, naming the official file and the window, scale and rounding applied, and \path{thirdparty/tp.py} rebuilds the files from the official downloads, writing a file only when its SHA-256 digest matches the file we scored against. \verb|bash reproduce.sh --sources <dir>| performs this rebuild, sets the paths, runs \path{stageB/analysis/run_all.sh}, which recomputes the data of every table and figure in Section~\ref{sec:results}, and writes the LaTeX tables and figures; without \path{--sources} it recomputes the analyses that do not score against those three label sets and keeps the shipped outputs of the others. No step calls a model. The synthetic dense images are regenerated, byte for byte, with
{\small
\begin{verbatim}
python3 -I stageB/code/scp_generate.py --num_images 50 --min_shapes 120 \
        --max_shapes 200 --min_radius 9 --max_radius 16 --seed 2026
\end{verbatim}
}

\section{Headline Claims and Their Evidence}\label{app:claims}

Table~\ref{tab:claims} traces the main numbers used in the abstract and introduction, and a few supporting ones, to the operation, comparator, metric and table that support it. Call and cost totals (67,364 calls, \$757.99) are itemized in Table~\ref{tab:registry}. 

{\small\setlength{\tabcolsep}{3pt}
\begin{longtable}{@{}>{\raggedright\arraybackslash}p{0.5cm}>{\raggedright\arraybackslash}p{4.1cm}>{\raggedright\arraybackslash}p{2.5cm}>{\raggedright\arraybackslash}p{2.5cm}>{\raggedright\arraybackslash}p{3.4cm}>{\raggedright\arraybackslash}p{2.0cm}@{}}
\caption{Headline claims.}\ltlabel{tab:claims}\\\toprule
\# & Claim & Operation & Comparator & Metric & Evidence \\\midrule\endfirsthead
\toprule \# & Claim & Operation & Comparator & Metric & Evidence \\\midrule\endhead
H1 & On SCP-dense image 28, GPT-6.1 lists 173 records that match all 173 shapes and states counts that sum to 188. Exact lists: GPT-6.1 20 of 50 scenes, with wrong stated counts in 19; Sonnet 5.5 39 of 50, wrong in all 39. List length misses the true total by 0.04 and 0.02; stated totals by 7.78 and 10.48 & none (same response) & stated counts, same response & exact list (every record matched, every shape covered); total-count error; SCP-dense 50, lowest setting & Fig.~\ref{fig:teaser}, Table~\ref{tab:readout} \\
H2 & Reading the list cuts per-type count error by 71\% to 84\% & list count & stated per-type counts, same response & per-type error; GPT $-75.6\%$ / $-84.2\%$, Sonnet $-77.7\%$ / $-70.7\%$ (two protocols) & Table~\ref{tab:readout} \\
H3 & Complete reporting and decoding protocol $-85.2\%$ (GPT-6.1), $-98.8\%$ (Sonnet 5.5); Gemini 3.8 (lowest) $+15.6\%$ $[-5.3,39.4]$, not significant & list-first request, scan, allowed names and alias map, readout of Step 6 & count-first prompt, stated count & per-type error & Table~\ref{tab:readout} \\
H4 & Alias map takes Sonnet 5.5 from 5.42 to 0.42 & alias map on the same list & raw list count of the same list & per-type error & Table~\ref{tab:readout} \\
H5 & Degeneracy check flags 45 of 50 Gemini 3.8 (lowest) lists; 40 fall back to the stated count, 5 keep the list count & four L3 flags & none & flag rate, readout branch & Table~\ref{tab:setting} \\
H6 & Default setting at the same 16k cap: flagged lists 90\% to 12\%, readout error $-54\%$ $[-72,-30]$; 32k cap: further $-37\%$ $[-62,8]$, not significant & setting, then cap & lowest 16k, then default 16k & count readout error & Table~\ref{tab:setting} \\
H7 & Same-model second click catches 17\% of GPT-6.1 errors; cheaper second model 61\%, acting on 86\% of items against 96\%; at 93\% acceptance coverage 39\% against 26\%; miss rate lower by 0.5 to 2.7 points at matched acceptance coverage (point estimates), and before zoom only the cross-model alarm reaches a miss rate near 4\% when acting (GPT-6.1, 4.0\%); after zoom 40\% to 50\% against 68\% to 78\% & agreement alarm, $\delta=1.5\%$ & same-model second call & errors caught; miss rate at equal acceptance coverage (point estimates) & Tables~\ref{tab:alarm} and \ref{tab:alarm_matched}, Fig.~\ref{fig:alarm} \\
H8 & Perfect selection 45\%; label-fitted selectors 16\% & constrained oracle; cross-fitted logistic gate & single call & reduction of $E$, median of 12 cells & Table~\ref{tab:select} \\
H9 & Unanimity ($K=3$) lowers set error significantly in 9 of 15 over-reporting cells, never raises it significantly & leave-one-source-out default & single call & $E$ & Table~\ref{tab:transfer} \\
H10 & The evaluated same-model transfer rule did not consistently improve F1 & leave-one-source-out default & single call & F1 (2 better, 4 worse, 12 n.s.) & Table~\ref{tab:transfer} \\
H11 & Cross-model defaults are better in 20 of 24 cells where the user's model is not strongest; all significantly worse cells are where it is strongest & leave-one-source-out cross-model default & single call & $E$ (12/2/4), F1 (10/3/5) & Table~\ref{tab:transfer} \\
H12 & Coordinate averaging is a separate term of either sign & selection and geometry decomposition & deletion-only intersection & $E$; exact identity on every image & Table~\ref{tab:intersect} \\
H13 & Systematic coordinate offset at most 0.3 px against a 30 px tolerance (about 1\%); overlays do not correct coordinates of current models (localization changes $\le0.23$ px); for Gemini 3.8 at its lowest setting a grid lowers set error by 3.5 through fewer omitted class fields, a scan by 11.5 & grid, ticks, cell labels & no overlay & localization error; set error & Section~\ref{sec:res:view} \\
H14 & Two-stage zoom: the miss rate $-65\%$ (Gemini 3.8), $+23$ points; Sonnet 5.5 $-41\%$ on 200 items, $-45\%$ on the 196 items every arm of Table~\ref{tab:common} answered & zoom & single click & hit rate, ScreenSpot-Pro 200 & Section~\ref{sec:res:view} \\
H15 & Self-consistency over three calls gains at most 23\% on dense counts; list-first readout 85\% and 99\% at about one call's cost & median of 3 stated counts; readout of Step 6 & one count-first call & per-type error, cost per image & Table~\ref{tab:common} \\
H16 & Two-stage zoom cuts the miss rate by up to 65\% for less than a three-click vote; vote and gated zoom gain at most 13\% & zoom on every item & one click; 3-click vote; gated zoom & miss rate, cost per item & Table~\ref{tab:common} \\
H17 & Unanimity beats box fusion (majority clusters, coordinates averaged with equal weights) in 11 of the 15 over-reporting cells, not significantly worse in any & leave-one-source-out default & majority with averaged coordinates & $E$ & Table~\ref{tab:common} \\
H18 & One entry point reproduces the component results from raw responses (counts 1.60, 0.42, 24.52; clicks 90.0\%, 87.0\%, 87.5\%; set error in 18 cells within 0.07, recovery off) and runs the synthetic boundary cases & Algorithm~\ref{alg:pipeline} with a replay client & component analyses & as in each table & replay, Section~\ref{sec:method} \\
H19 & Confirming the whole-image report with four crop reports lowers set error in 4 of 6 cells (up to $-45.5\%$); in the other 2, $p_1>1/2$ & cross-view confirmation, shipped rule & one whole-image call; second whole-image call; three-call unanimity & $E$ at IoU $\ge0.5$, 30 images per cell & Section~\ref{sec:res:view} \\
H20 & $p_1$ on 5 labelled images predicts whether deletion helps on the other images in 83\% of draws, the family default without labels in 85\%, and the Step 7 rule from the family default in 87\%, cutting the loss from a wrong decision from 0.43 to 0.17 errors per image & deletion of unconfirmed records & family default; plain threshold & sign of the selection gain on the other images, 54 pairs $\times$ 200 draws & Table~\ref{tab:intersect}, Section~\ref{sec:res:set} \\
H21 & On aerial and document images with several classes the readout gap is small and the protocol changes count error by $-20\%$ to $+0.4\%$ (GPT-6.1, Sonnet 5.5, not significant) and $-25.8\%$, $-28.2\%$ (Gemini 3.8 default; DOTA interval reaches zero) & list-first protocol & count-first stated count & per-type count error, 80 images & Table~\ref{tab:natural} \\
H22 & Three-model tier misses 9.33\% against 13.67\% ($-4.33$ points $[-6.33,-2.33]$, 34/8, $p=7\times10^{-5}$) on 600 ScreenSpot-Pro items; against three GPT-6.1 zooms 12.67\% (Holm $p=0.002$) at about the same cost (\$0.0273 against \$0.0255 per item) & Algorithm~\ref{alg:clicks} & GPT-6.1 default two-stage zoom; three such zooms with a cluster vote & miss rate, all 600 items in the denominator & Table~\ref{tab:choose} \\
H23 & Against a crop-first baseline in GUI-Lens style, $-2.17$ points $[-4.17,0.00]$, Holm $p=0.060$, not significant; the comparison step adds $-0.83$ $[-2.00,0.17]$, not significant & Algorithm~\ref{alg:clicks}; comparison step & crop-first baseline (ours); scorer's score alone & miss rate & Table~\ref{tab:choose} \\
H24 & Misses split into 43 with no correct candidate, 3 lost representatives, 10 wrong choices; a repeated GPT-6.1 call hits 18\% of the items GPT-6.1 missed, Sonnet 5.5 29\%, Gemini 3.8 37\%; three GPT-6.1 zooms leave 62 items with no correct candidate, the three models 43 & three-term split; rescue rate $\phi$ & one and three GPT-6.1 zooms & items with no correct candidate & Section~\ref{sec:res:choose} \\
H25 & A single-call configuration chosen on three detection sources raises F1 on the fourth in 5 of 12 cells and lowers it in 1; macro F1 42.1 to 49.1 & leave-one-source-out configuration & list interface, lowest setting & F1, primary scoring & Table~\ref{tab:config} \\
H26 & Adding the crop-first point as a fourth source lowers items with no correct candidate from 43 to 36 but raises lost representatives from 3 to 7 and wrong choices from 10 to 11; 54 against 56 misses, $-0.33$ $[-1.17,0.34]$, not significant & four-source tier, same rules & three-source tier & miss rate and its three terms, same 600 items & Section~\ref{sec:res:choose} \\
H27 & A plain cluster vote over the same three models' points misses 74 items (12.33\%), not significantly different from three GPT-6.1 zooms ($-0.33$ $[-1.83,1.17]$); the tier's miss rate is lower than the vote's by 3.00 points ($-3.00$ $[-4.67,-1.50]$), 22/4, $p=5\times10^{-4}$; given GPT-6.1's 82 misses, the partners' rescue rates on the 260 GUI items give 46 items with no correct candidate among the 600, against 43 observed & scorer's score and comparison (Algorithm~\ref{alg:clicks}) & plain three-model cluster vote, same candidates & miss rate, all 600 items in the denominator & Table~\ref{tab:choose}, Section~\ref{sec:res:choose} \\\bottomrule
\end{longtable}\ltlabelafter{tab:claims}}

\end{document}